\documentclass[sn-mathphys,Numbered]{sn-jnl}%

\usepackage{graphicx}
\usepackage{multirow}
\usepackage{amsmath,amssymb,amsfonts}
\usepackage{amsthm}%
\usepackage{mathrsfs}%
\usepackage[title]{appendix}%
\usepackage{xcolor}%
\usepackage{textcomp}%
\usepackage{manyfoot}%
\usepackage{algorithm}%
\usepackage{adjustbox}
\usepackage{algpseudocode}%
\usepackage{colortbl}
\usepackage[font=scriptsize]{caption}
\usepackage{listings}%
\usepackage{cleveref}
\definecolor{bgcolor}{rgb}{0.8,1,1}
\definecolor{bgcolor2}{rgb}{0.8,1,0.8}
\usepackage{pifont}
\usepackage{xcolor}
\usepackage{algorithm,algorithmicx,algpseudocode}
\usepackage{tabularx}
\usepackage{multirow}
\usepackage{colortbl}
\usepackage{makecell}
\usepackage{lscape}
\usepackage{mathtools}

\newtheorem{theorem}{Theorem}
\newtheorem{lemma}{Lemma}%
\newtheorem{assumption}{Assumption}

\newtheorem{definition}{Definition}%

\def\R{\mathbb{R}}

\def\R{\mathbb R}

\def\EE{\mathbb E}

\def\y{\mathbf{y}}

\def\x{\mathbf{x}}

\def\<#1,#2>{~\left\langle #1,  #2\right\rangle}
\newcommand{\norm}[1]{\left\| #1 \right\|}
\newcommand{\sqn}[1]{\norm{#1}^2}
\newcommand{\EndProof}{\begin{flushright}$\square$\end{flushright}}

\newcommand{\mW}{{\bf W}}

\newcommand{\sm}[2]{\begin{smallmatrix}\item1\\\item2 \end{smallmatrix}}

\def\R{\mathbb{R}}

\newcommand\ddfrac[2]{\frac{\displaystyle \item1}{\displaystyle \item2}}

\newcommand{\mc}[1]{\mathbb{\item1}}

\begin{document}

\title[Decentralized Personalized Federated Learning for Min-Max Problems]{Decentralized Personalized Federated Learning for Min-Max Problems}

\author*[1]{\fnm{Ekaterina} \sur{Borodich}}\email{borodich.ed@phystech.edu}

\author[1,2]{\fnm{Aleksandr} \sur{Beznosikov}}

\author[1,3]{\fnm{Abdurakhmon} \sur{Sadiev}}

\author[4]{\fnm{Vadim} \sur{Sushko}}

\author[1]{\fnm{Nikolay} \sur{Savelyev}}

\author[2]{\fnm{Martin} \sur{Takac}}

\author[1,3]{\fnm{Alexander} \sur{Gasnikov}}

\affil[1]{\orgname{Moscow Institute of Physics and Technology}, \orgaddress{\city{Moscow}, \country{Russia}}}

\affil[2]{\orgname{Mohamed bin Zayed University of Artificial Intelligence}, \orgaddress{\city{Abu Dhabi}, \country{United Arab Emirates}}}

\affil[3]{\orgname{Ivannikov Institute for System Programming RAS}, \orgaddress{\city{Moscow}, \country{Russia}}}

\affil[4]{\orgname{Bosch Center for Artificial Intelligence}, \orgaddress{\city{Renningen}, \country{Germany}}}


\maketitle

\begin{abstract}

Personalized Federated Learning (PFL) has witnessed remarkable advancements, enabling the development of innovative machine learning applications that preserve the privacy of training data.  
However, existing theoretical research in this field has primarily focused on distributed optimization for minimization problems. 
This paper is the first to study PFL for saddle point problems encompassing a broader range of optimization problems,   
that require more than just solving minimization problems.
In this work, we consider a recently proposed PFL setting with the mixing objective function, an approach combining the learning of a global model together with locally distributed learners. 
Unlike most previous work, which considered only the centralized setting, we work in a more general and decentralized setup that allows us to design and analyze more practical and federated ways to connect devices to the network. We proposed new algorithms to address this problem and provide a theoretical analysis of the smooth (strongly) convex--(strongly) concave saddle point problems in stochastic and deterministic cases. Numerical experiments for bilinear problems and neural networks with adversarial noise demonstrate the effectiveness of the proposed methods.
\end{abstract}

\section{Introduction}
Distributed optimization methods have already become integral to solving problems, including many applications in machine learning. 
For example, distributing training data evenly across multiple devices can greatly speed up the learning process. Recently, a new research direction has appeared concerning distributed optimization -- Federated Learning (FL) \cite{mcmahan2017communication,konevcny2016federated}. 
Unlike classical distributed learning methods, the FL approach assumes that data is not stored within a centralized computing cluster but is stored on clients' devices, such as laptops, phones, and tablets. This formulation of the training problem gives rise to many additional challenges, including the privacy of client's data and the high heterogeneity of data stored on local devices, to name a few.
The goal of the most standard setting of distributed or federated learning is to find the global models weights based on all local data. 

{\bf Personalized FL.} 
In this work, we allow each client to build their own personalized model and utilize a decentralized communication protocol that will enable harvesting information from other local models (trained on local clients' data).
Predicting the next word written on a mobile keyboard \cite{hard2018federated} is a typical example when the performance of a local (personalized) model is significantly ahead of the classical FL approach that trains only the global model.  
Improving the local models using this additional knowledge may need a more careful balance, considering a possible discrepancy between data splits that the local models were trained on. 
Attempts to find the balance between personalization and globalization have resulted in a series of works united by a general name -- Personalized Federated Learning (PFL). 
We refer the reader to the following survey papers
\cite{kulkarni2020survey,kairouz2019advances} for more details and explanations of different techniques.  

{\bf Saddle Point Problems.} All previous results around personalized setting focus on the minimization problem, we consider Saddle Point Problems (SPPs). 
SPPs cover a wider range of problems than minimization ones and has numerous important practical applications \cite{Du1995}.
These include well-known and famous examples from game theory or optimal control \cite{facchinei2007finite}. In recent years, saddle point problems have become popular in several other respects.
One can note a branch of recent work devoted to solving non-smooth problems by  reformulating them as saddle point problems \cite{nesterov2005smooth,nemirovski2004prox}, as well as applying such approaches to image processing
\cite{chambolle2011first,esser2010general}. Recently, significant attention was devoted to saddle problems in machine learning. For example, Generative Adversarial Networks (GANs) are written as a min-max problem 
\cite{goodfellow2014generative}. In addition, there are many popular examples: robust models with adversarial noise \cite{madry2017towards}, 
Lagrangian multipliers \cite{boyd2011distributed,srivastava_distributed,Nunec_distributed}, supervised learning (with non-separable loss \cite{Thorsten}, with non-separable regularizer\cite{bach2011optimization}), unsupervised learning \cite{NIPS2004_64036755} and reinforcement learning \cite{Omidshafiei2017:rl,Jin2020:mdp}. 

Furthermore, there are a lot of personalized federated learning problems utilize saddle point formulation. In particular,  Personalized Search Generative Adversarial Networks (PSGANs) \cite{PSGAN}. As mentioned in examples above, saddle point problems often arise as an auxiliary tool for the minimization problem. It turns out that if we have a personalized minimization problem, and then for some reason (for example, to simplify the process of the solution or to make the learning more stable and robust) rewrite it in the form of a saddle point problem, then we begin to have a personalized saddle point problem. We refer the reader to Section~\ref{sec:a_mot} for more details.

\subsection{Problem formulation}
In this paper, we focus on Decentralized Personalized Federated Saddle Point Problem (PF SPP) with a mixing objective:
\begin{align}
    \label{PF}
   & \min_{x_1, \ldots, x_M} \max_{y_1,\ldots,y_M } \left\{\sum_{m=1}^M f_m(x_m,y_m) + \frac{\lambda}{2}\left\| \sqrt{W} X\right\|^2 - \frac{\lambda}{2} \left\| \sqrt{W} Y\right\|^2\right\},
 \end{align}
where $x_1, \ldots, x_M$ and $y_1,\ldots,y_M$ are interpreted as local models on nodes which are grouped into matrices $X := [x_1, \ldots, x_M]^T$ and $Y := [y_1, \ldots, y_M]^T$. $W$ is the gossip matrix reflecting the properties of the communication graph between the nodes. $\lambda > 0$ is the key  regularization parameter, which corresponds to the personalization degree of the models. 

Note that in the proposed formulation \eqref{PF} we consider both the centralized and decentralized cases. In the decentralized setting, all nodes are connected within a network, and each node can communicate/exchange information only with their neighbors in the network.  While the centralized architecture consists of master-server that connected with all devices which communicate to the central server.  
But in theory, the centralized case is similar to decentralized with  a complete computational graph. If we set $W$ to the Laplacian of a complete graph, it is easy to verify that we obtain the following centralized PF SPP:
\begin{align}
    \label{PF_cent}
   & \min_{x_1, \ldots, x_M} \max_{y_1,\ldots,y_M } \left\{\frac{1}{M} \sum_{m=1}^M f_m(x_m,y_m) + \frac{\lambda}{2M}  \sum_{m=1}^M \| x_m - \bar x\|^2 - \frac{\lambda}{2M} \sum_{m=1}^M \| y_m - \bar y\|^2\right\},
\end{align}
where $\bar x = \frac{1}{M}\sum_{m=1}^M x_m$ and $\bar y = \frac{1}{M}\sum_{m=1}^M y_m$ are average global models. 

Unlike \eqref{PF_cent}, the formulation \eqref{PF} penalizes not the difference with the global average, but the sameness with other connected local nodes. Thereby the decentralized case can be artificially created in centralized architecture, e.g., if we want to create the network and $W$ matrix to connect only some clients based on their location, age and other meta data. The regularization parameter $\lambda$ is responsible for importance degree of this difference. For example, with $\lambda = 0$ the problem \eqref{PF} will decompose into $M$ separable problems and each $m\in M$ will independently train just a local model. As $\lambda$ increases, local models begin to use the information from their neighbours due to increase the "importance" degree of regularization terms. The idea of using this type of penalty is not new and has been used in the literature in several contexts, in particular for classical decentralized minimization \cite{li2020decentralized,gorbunov2019optimal} with large $\lambda$ and for multitask PFL \cite{smith2017federated,wang2018distributed} with small $\lambda$.

\subsection{Summary of Contributions}

\begin{table*}[h]
\centering
\small
\resizebox{0.6\textwidth}{!}{
\begin{threeparttable}
\begin{tabular}{|c|c|c|c|}
\cline{3-4}
\multicolumn{2}{c|}{}
& Communications & Local computations 
\\ 
\hline
%
\multirow{4}{*}{\rotatebox[origin=c]{90}{Proximal~~~}} & \multirow{2}{*}{\rotatebox[origin=c]{90}{Upper}}
&\cellcolor{bgcolor2}$\min\left[\tfrac{L\sqrt{\chi}}{\mu}; \sqrt{\tfrac{\lambda \lambda_{\max}(W)}{\mu}}\right] \log \tfrac{1}{\varepsilon}$
& \cellcolor{bgcolor2}$\min\left[\tfrac{L}{\mu}; \sqrt{\tfrac{\lambda \lambda_{\max}(W)}{\mu}}\right] \log \tfrac{1}{\varepsilon}$
\\&
& \cellcolor{bgcolor2}{\scriptsize \textbf{(Algorithm \ref{alg:sliding_opt_comm} and Algorithm \ref{alg:sliding_big_lambda})}}
& \cellcolor{bgcolor2}{\scriptsize \textbf{(Algorithm \ref{alg:sliding_opt_comm} and Algorithm \ref{alg:sliding_big_lambda})}}
\\\cline{2-4}
%
& \multirow{2}{*}{\rotatebox[origin=c]{90}{Lower}}
&$\min\left[\tfrac{L\sqrt{\chi}}{\mu}; \sqrt{\tfrac{\lambda \lambda_{\max}(W)}{\mu}}\right] \log \tfrac{1}{\varepsilon}$
&$\min\left[\tfrac{L}{\mu}; \sqrt{\tfrac{\lambda \lambda_{\max}(W)}{\mu}}\right] \log \tfrac{1}{\varepsilon}$
\\&
& {\scriptsize \textbf{(Section \ref{sec:lower})}}
& {\scriptsize \textbf{(Section \ref{sec:lower})}}
\\\hline
\multirow{4}{*}{\rotatebox[origin=c]{90}{Gradient \quad}} & \multirow{2}{*}{\rotatebox[origin=c]{90}{Upper}}
& \cellcolor{bgcolor2}$\min\left[\tfrac{L\sqrt{\chi}}{\mu}; \sqrt{\tfrac{\lambda \lambda_{\max}(W)}{\mu}}\right] \log \tfrac{1}{\varepsilon}$
& \cellcolor{bgcolor2}$\tfrac{L}{\mu} \log \tfrac{1}{\varepsilon}$
\\&
& \cellcolor{bgcolor2}{\scriptsize \textbf{(Algorithm \ref{alg:sliding_opt_comm} and Algorithm \ref{alg:sliding_big_lambda})}}
& \cellcolor{bgcolor2} {\scriptsize \textbf{(Algorithm \ref{alg:sliding_opt_comm} and Algorithm \ref{alg:sliding_big_lambda})}}
\\\cline{2-4}
& \multirow{2}{*}{\rotatebox[origin=c]{90}{Lower}}
&$\min\left[\tfrac{L\sqrt{\chi}}{\mu}; \sqrt{\tfrac{\lambda \lambda_{\max}(W)}{\mu}}\right] \log \tfrac{1}{\varepsilon}$
&$\tfrac{L}{\mu} \log \tfrac{1}{\varepsilon}$
\\&
& {\scriptsize \textbf{(Section \ref{sec:lower})}}
& {\scriptsize \textbf{(Section \ref{sec:lower})}}
\\\hline
\multirow{6}{*}{\rotatebox[origin=c]{90}{Stochastic \quad \quad}} & \multirow{4}{*}{\rotatebox[origin=c]{90}{Upper}}
&$\tfrac{\lambda \lambda_{\max}(W)}{\mu}\log \tfrac{1}{\varepsilon}$
& \cellcolor{bgcolor2}$\max \left[r, \sqrt{r}\tfrac{L}{\mu}\right] \log \tfrac{1}{\varepsilon}$
\\&
& \scriptsize \textbf{(Algorithm \ref{alg_sum})}
& \cellcolor{bgcolor2}{\scriptsize \textbf{(Algorithm \ref{alg_sum})}}
\\\cline{3-4}
&
& \cellcolor{bgcolor}$\sqrt{\tfrac{\lambda \lambda_{\max}(W)}{\mu}} \log \tfrac{1}{\varepsilon}$
& \cellcolor{bgcolor} $\max\left[r\sqrt{\tfrac{\lambda \lambda_{\max}(W)}{\mu}}, \tfrac{\sqrt{r} L}{ \mu} \right]\log \tfrac{1}{\varepsilon}$
\rule[0pt]{0pt}{14pt} 
\\&
& \cellcolor{bgcolor}{\scriptsize \textbf{(Algorithm \ref{alg:sliding_opt_comm})}}
& \cellcolor{bgcolor}{\scriptsize \textbf{(Algorithm \ref{alg:sliding_opt_comm})}}
%
\\ 
\cline{2-4}
%
& \multirow{2}{*}{\rotatebox[origin=c]{90}{Lower}}
&$\min\left[\tfrac{L\sqrt{\chi}}{\mu}; \sqrt{\tfrac{\lambda \lambda_{\max}(W)}{\mu}}\right] \log \tfrac{1}{\varepsilon}$
&$\max \left[r, \sqrt{r}\tfrac{L}{\mu}\right] \log \tfrac{1}{\varepsilon}$
\\&
& {\scriptsize \textbf{(Section \ref{sec:lower})}}
& {\scriptsize \textbf{(Section \ref{sec:lower})}}
\\\hline
\end{tabular}
\end{threeparttable}
}
\caption{Summary of complexity results (upper and lower bounds) for finding an $\varepsilon$-solution in the deterministic proximal (any local computations are cheap), deterministic gradient and stochastic (finite-sum) setups. 
Convergence is measured by the distance to the solution. 
{\em Notation:} $\mu$ is a strong-convexity constant of $f$, $L$ is a smoothness constant of $f$,  
$\lambda_{\max}(W)$ is the maximum eigenvalue of $W$, $\lambda^+_{\min}(W)$ is the minimum positive eigenvalue of $W$, $\chi = \lambda_{\max}(W) / \lambda^+_{\min}(W)$ , $r$ is the size of the local dataset. 
\\
We highlighted optimal bounds (match lower bounds) in \colorbox{bgcolor2}{green} and optimal bounds for some cases  in \colorbox{bgcolor}{blue}.}\label{tab:main}
\end{table*}

To the best of our knowledge, this paper is the first to consider decentralized personalized federated saddle point problems, propose optimal algorithms and derives the computational and communication lower bounds for this setting. In the literature, there are works on general (non-personalized) SPPs. We make a detailed comparison with them in Appendix \ref{sec:comp}. Due to the fact that we consider a personalized setting, we can have a significant gain in communications. For example, when $\lambda = 0$ or small enough in \eqref{PF} the importance of local models increases and we may communicate less frequently.
We now outline the \textit{main} contribution of our work as follows (please refer also Table~\ref{tab:main} for an overview of the results):
\begin{itemize}
\item We present a new SPP formulation of the PFL problem \eqref{PF} as the decentralized min-max mixing model. This extends the classical PFL problem to a broader class of problems beyond the classical minimization problem. It furthermore covers various communication topologies and hence goes beyond the centralized setting.
\item We propose a lower bounds both on the communication and the number of local oracle calls for a general algorithms class (that satisfy Assumption~\ref{ass_lower_bounds}). The bounds naturally depend on the communication matrix $W$ (as in the minimization problem), but our results apply to SPP (see "Lower" rows in Table~\ref{tab:main}
for various settings of the SPP PFL formulations).
\item
We develop multiple novel algorithms to solve decentralized personalized federated saddle-point problems. These methods (\Cref{alg:sliding_opt_comm} and \Cref{alg:sliding_big_lambda}) are based on recent sliding technique \cite{lan2016gradient,rogozin2021decentralized,hanzely2020lower} adapted to SPPs in a decentralized PFL.  In addition, we present \Cref{alg_sum} which used the randomized local method from \cite{alacaoglu2021stochastic}. This algorithm is used to compare \Cref{alg:sliding_opt_comm} with Local randomized methods (like \Cref{alg_sum}) in practice. 
\item  We provide the theoretical convergence analysis for all proposed algorithms. According to Table \ref{tab:main}, 
we have optimal algorithms that achieve the lower bounds in all cases. 
\item We adapt the proposed algorithm for training neural networks. 
We compare our algorithms: type of sliding (\Cref{alg:sliding_opt_comm}) and type of local method (\Cref{alg_sum}). To the best of our knowledge, this is the first work that compares these approaches in the scope of neural networks, as previous studies were limited to simpler methods, such as regression problems \cite{hanzely2020federated,hanzely2020lower}. Our experiments confirm the robustness of our methods on the problem of training a classifier with adversarial noise.
\end{itemize}

\section{Notation and Assumptions}\label{sec:formul}

For vectors, we use the Euclidean norm $\| \cdot\| = \| \cdot \|_2$ everywhere, and for matrices the Frobenius norm $\| \cdot \|_F$.  We denote by $I$ the identity matrix and by $E$ an all-ones matrix.
For given convex-concave function $g(x,y)$ and any $\hat{x} \in \mathbb{R}^{d_x}, \hat{y} \in \mathbb{R}^{d_y}$ we define proximal operator as follows: $\text{prox}_g (x,y) = \arg \min_{x \in \mathbb{R}^{d_x}} \arg\max_{y \in \mathbb{R}^{d_y}} \{ g(x,y) + \frac{1}{2}\|\hat{x} - x\|^2 - \frac{1}{2}\|\hat{y} - y\|^2\}$. We say that $\{(x_m, y_m)\}_{m = 1}^M$ is $\varepsilon$-solution to \eqref{PF} if $\sum_{m=1}^M \|x_m - x_m^*\|^2 + \|y_m - y_m^*\|^2 \leq \varepsilon$.

Also, we introduce standard assumptions on $f_m$.
\begin{assumption} \label{ass:smooth} Each $f_m$ is $L$-smooth on $\R^{d_x} \times \R^{d_y}$, i.e. for all $u_1, u_2 \in \R^{d_x}$, $v_1, v_2 \in \R^{d_y}$ it holds that
$$
    \left\|\nabla f_m (u_1, v_1) - \nabla f_m (u_2, v_2) \right\| 
     \leq L \left\|(u_1, v_1) - (u_2, v_2) \right\|.
$$
For case, when $f_m(x) = \frac{1}{r}\sum_{i=1}^r f_{m, i}(x)$ we assume that $f_{m, i}(x)$ is $L$-smooth.
\end{assumption}

\begin{assumption}\label{ass:sc}
Each $f_m$ is $\mu$-strongly convex - strongly concave ($\mu > 0$) on $\R^{d_x} \times \R^{d_y}$, i.e. for all $x_1, x_2 \in \R^{d_x}$, $y_1, y_2 \in \R^{d_y}$ it holds that
$$
    \left\langle \begin{matrix}
        \nabla_x f_m(x_1, y_1) - \nabla_x f_m (x_2, y_2) \\
        - \nabla_y f_m (x_1, y_1) + \nabla_y f_m (x_2, y_2)
    \end{matrix}; \begin{matrix}
    x_1 - x_2 \\
    y_1 - y_2
    \end{matrix}\right\rangle
    \geq \mu \left( \|x_1 - x_2 \|^2 + \|y_1 - y_2 \|^2\right).
$$
\end{assumption}


{\bf Communication network.}
The communication network between devices can be represented as a fixed, connected, undirected graph $\mathcal{G} := (\mathcal{V}, \mathcal{E})$, where the set of vertices $\mathcal{V}$ represents  the set of computing nodes, and the edges $\mathcal{E} \subseteq \mathcal{V} \times \mathcal{V}$ indicate the presence or absence of connection between the corresponding nodes. Recall that we consider a decentralized case, where information exchange (by gossip protocol) is possible only between neighbors. In such a case, communication can be represented as a matrix multiplication with a matrix $W$ introduced in \eqref{PF}. It remains to introduce a formal definition of gossip matrix \cite{nedic2009distributed, boyd2006randomized}.
\begin{definition}
\label{def_gossip}
We call a matrix $W$ a gossip matrix if it satisfies the following conditions: 
\\
1) $W$ is an $M \times M$ symmetric and positive semi-definite, 
\\
2) $\text{ker}(W) = \text{span}\{(1,\ldots,1)\}$, 
\\
3) $W$ is defined on the edges of the network: $w_{ij} \neq 0$ only if $i=j$ or $(i,j) \in \mathcal{E}$.
\end{definition}
For this matrix we define $\lambda_{\max}(W)$ for a maximum eigenvalue of $W$, $\lambda^+_{\min}(W)$ for a minimum positive eigenvalue of $W$ and $\chi = \lambda_{\max}(W) / \lambda^+_{\min}(W)$.

{\bf Local oracles.}
For local computations we introduce three different oracles: proximal, gradient and summand gradient. More formally, for each device $m$ and local points $x_m, y_m$ we can compute one of these oracles.  
For the \textit{proximal oracle} $ \text{Loc}(x^k_m, y^k_m, m) = \{ \text{prox}_{f_m}(x_m,y_m)\}_{m=1}^M$ we solve any local (depends on $f_m$ and $x_m, y_m$) min-max subproblems in $\mathcal{O}(1)$ local computation (for convex-concave problems a solution always exists \cite{nikaido2016convex}).
For the \textit{gradient oracle} $ \text{Loc}(x^k_m, y^k_m, m) = \{(\nabla_{x} f_m(x_m, y_m), - \nabla_{y} f_m(x_m, y_m))^{T}\}_{m=1}^M$ gives access to local gradients. For \textit{summand gradient} $\{(\nabla_x f_{m,j_m}(x_m, y_m), - \nabla_y f_{m, j_m}(x_m, y_m))^{T}\}_{m=1}^M$ refers to the stochastic case, when local functions have finite-sum structure: $f_m = \frac{1}{r}\sum_{j=1}^r f_{m,j}$, and in each call of oracle we can compute gradients of only summand (selected randomly or deterministically).

\section{Lower bounds}\label{sec:lower}

Before presenting the lower complexity bound for solving the problem \eqref{PF}, we define the algorithm class for which the lower bound will be proved. 
\begin{assumption}\label{ass_lower_bounds}
Let us consider the algorithm $\mathcal{A}$ for \eqref{PF} with the communication graph $\mathcal{G}$. For each device in the network $\mathcal{G}$ we define sequence of local memory $\{\mathcal{M}_{m, k}\}^{\infty}_{k = 1}$ for $1 \leq m \leq M$ with $\mathcal{M}_{m, 0} =\text{span}\left\{(x^0_m, y^0_m\}\right)$ and the update rule for these memories: 
\begin{eqnarray*}
\mathcal{M}_{m, k+1} =\begin{cases}
 \text{span}\left\{\mathcal{M}_{m, k}, \text{Loc}(x^k_m, y^k_m, m)\right\}& ~\text{if local oracle  at the iteration }k\\
 \text{span}\left\{\bigcup_{j: (m,j) \in\mathcal{E} }\mathcal{M}_{j, k}\right\}& ~\text{if communication oracle at the iteration } k,
\end{cases}
\end{eqnarray*}
where $(x^k_m, y^k_m) \in \mathcal{M}_{m, k}$. After $N$ iterations, the output of each device $m$ is some point $(\hat x^N_m, \hat y^N_m) \in \mathcal{M}_{m, N}$.
\end{assumption}
Note that the communication oracle starts a communication round, i.e. all devices exchange information with their neighbours on the network. The local computation oracle is defined above (see \Cref{sec:formul}). 
In fact, local memory $\mathcal{M}_{m, k}$ limits the set of points the device $m$ can reach after $k$ iterations. Information exchange with neighbours as well as local computation can increase this set of points. The algorithm $\mathcal{A}$ can be either stochastic or deterministic, it depends on what is used as the local oracle.
Similar assumption are used in other works on different lower bounds \cite{nesterov2018lectures, scaman2017optimal, hendrikx2020optimal, hanzely2020lower}. 

\subsection{Lower complexity bounds on the communications}
The following theorem presents the lower bound on the communication complexity.
\begin{theorem}
\label{theorem_lower_bounds}
For any positive parameters $\chi \geq 3$, $\mu > 0$, $L \geq 2\mu$,  $\lambda > 0$, $\lambda^+_{\min} > 0$ ($\lambda \lambda_{\min}^+ \geq \mu$), $\lambda_{\max} \geq \lambda^+_{\min}$ and any integer $k > 0$ there exists a problem of the form \eqref{PF} satisfying Assumptions \ref{ass:smooth} - \ref{ass:sc} on graph $\mathcal{G}$ with a gossip matrix $W$ (such that $\lambda_{\max}(W) = \lambda_{\max}$, $\lambda^+_{\min}(W) = \lambda^+_{\min}$) and a starting point $(x^0, y^0)$, such that any algorithm $\mathcal{A}$ satisfying Assumption \ref{ass_lower_bounds} needs at least
\begin{align*}
    \Omega \left( \min \left\{ \sqrt{\frac{\lambda \lambda_{\max}(W)}{\mu}}, \frac{L}{\mu}\sqrt{\chi}\right\} \log \frac{1}{\varepsilon}\right)~~\text{communications to find $\varepsilon$-solution of \eqref{PF}.}
\end{align*}
\end{theorem}
Note that the lower bound not depend on which local oracles we use. This seems natural, because from a communication point of view it does not matter how certain local subproblems are solved. The same effect can be seen for decentralized (not personalized) minimization problems: \cite{scaman2017optimal} gives lower bounds on communications in the deterministic case and \cite{hendrikx2020optimal} in the stochastic case, both these results are the same.

\subsection{Lower complexity bounds on the local oracle calls}

Next, we present the lower complexity bounds on the number of the local oracle calls for their various types.

To get the lower bounds on the number of \textbf{gradient oracle} calls for fixed $M$ (number of nodes) we choose gossip matrix $W$ as a Laplace matrix of a fully connected graph. If we taking $f_1 = f_2 = \cdots = f_M$ and starting from $(x_m^0,y_m^0)=(x^0,y^0)$ for all nodes $ m \in \overline{(1, M)}$ then the problem \eqref{PF} reduces to a min-max single
local function $f_1$. 
From \cite{zhang2019lower}, we know that the worst-case need at least $\Omega\left(\frac{L}{\mu}\log{\frac{1}{\varepsilon}}\right)$ 
gradient calls to find $\varepsilon$-solution \cite{zhang2019lower}. Hence, we can obtain the lower bound for gradient oracle. Due to we start from the same starting point on each node and all the local functions are identical and the communication does not help in the convergence. 

 To present the lower bound for \textbf{proximal oracle}  we note that the gradient oracle can be considered as a particular case of the proximal ones. Since, it can also solve local subproblem but not necessarily in a single local iteration.  Due to this fact, there is a final lower bound for the proximal oracle can be smaller or the same than the lower bound for the gradient ones. The construction we performed in Theorem \ref{theorem_lower_bounds} requires  $\Omega\left(\sqrt{\frac{\lambda\lambda_{\max}(W)}{\mu}}\log{\tfrac{1}{\varepsilon}}\right)$ (when $\sqrt{\tfrac{\lambda\lambda_{\max}(W)}{\mu}} = \mathcal{O}\left(\frac{ L}{\mu}\sqrt{\chi}\right)$) communication rounds to reach $\varepsilon$-solution of \eqref{PF}, but on top of that, it also requires at least $\Omega\left(\sqrt{\tfrac{\lambda\lambda_{\max}(W)}{\mu}}\log{\tfrac{1}{\varepsilon}}\right)$
calls of local oracle. It means that the final lower bounds on the proximal oracle is $\Omega\left(\min\left\{\sqrt{\tfrac{\lambda\lambda_{\max}(W)}{\mu}}, \tfrac{L}{\mu}\right\}\log{\tfrac{1}{\varepsilon}}\right)$.

To present the lower bound for \textbf{summand gradient oracle} calls we assume that algorithm $\mathcal{A}$ is either deterministic or generated by given seed that is initialized identically for all clients. The same way as for gradient oracle let us set $(x_m^0, y_m^0) = (x_0, y_0)$ for all nodes, $W$ is the Laplace matrix of a fully connected graph and $f_1 = f_2 = \dots = f_M$. For this algorithm $\mathcal{A}$ all local iterates will be identical, i.e., $(x^k_1, y^k_1)= (x^k_2,y^k_2) = \dots = (x^k_M, y^k_M)$ for all $k \geq 0$. Consequently, the problem reduces to min-max of a single finite sum objective $f_1$, which needs at least $\Omega\left(\left(r+\tfrac{\sqrt{r}L}{\mu}\right)\log{\tfrac{1}{\varepsilon}}\right)$ summand gradient oracle calls \cite{han2021lower} to get $\varepsilon$-solution.

\section{Optimal Algorithms}

In this section, we present three new algorithms for solving the problem \eqref{PF}. We include their convergence properties and discuss their limitations.

The simplest method for solving \eqref{PF} is to consider the function $\left\{\sum_{m=1}^M f_m(x_m,y_m) + \tfrac{\lambda}{2}\| \sqrt{W} X\|^2 - \tfrac{\lambda}{2} \| \sqrt{W} Y\|^2\right\}$ as a whole, not to take into account its composite structure. As a basic method we may consider the classical method for smooth saddle point problems -- Extra Step Method \cite{Korpelevich1976TheEM} (or Mirror Prox \cite{juditsky2011solving}). Then the number of oracle calls for the saddle function and for the composites are the same. 
Note that in the problem \eqref{PF} the step along the gradient of the regularizer ($\tfrac{\lambda}{2}\| \sqrt{W} X\|^2 - \tfrac{\lambda}{2} \| \sqrt{W} Y\|^2$) requires communication with neighboring nodes (due to multiplication by the matrix $W$). Meanwhile, for gradient calculation of $\sum_{m=1}^M f_m(x_m, y_m)$ it is enough to calculate all the local gradients of $f_m$ and do not exchange information at all. 
Certainly, we want to reduce the number of communications (or calls the regularizer gradient) as much as possible. 
This is especially important when the problem \eqref{PF} is a fairly personalized ($\lambda \ll L$) and information from other nodes is not significant. To solve this problem and separate the oracle complexities for the saddle function and the composites, we base our method on \emph{sliding technique} \cite{lan2016gradient}. 
The optimal method for PF minimization from \cite{hanzely2020lower} is also a kind of sliding method. 

It is clear that the method from \cite{hanzely2020lower} cannot be used for saddle point problems. Sliding for saddles has its own specifics -- exactly for the same reasons why Extra Step Method is used for smooth saddles instead of the usual Descent-Ascent \cite{gidel2018variational} (at least because Descent-Ascent diverges for the most common bilinear problems).
\vspace{-0.2cm}
\subsection{Sliding for case $\sqrt{\tfrac{\lambda\lambda_{\max}(W)}{\mu} } = \mathcal{O}\left(\tfrac{L}{\mu}\sqrt{\chi}\right)$}

\begin{algorithm}[h]
	\caption{Accelerated Sliding for small $\lambda$}
	\label{alg:sliding_opt_comm}
	\hspace*{\algorithmicindent} {\bf Parameters:} $\alpha $, $\eta$\\
	\hspace*{\algorithmicindent} {\bf Initialization:} choose  $ x^0,y^0$, $x^0_m = x^0$, $y^0_m = y^0$  for all nodes $m \in (1, M)$
	\begin{algorithmic}[1]
		\For{$k=0,1,2,\ldots$}
		\Statex \textcolor{blue}{\textbf{Local updates}} for all clients: \State  $v^k_{x_m} = \alpha x_m^k + (1 - \alpha) u_{x_m}^k$, \ \ \ \ $v^k_{y_m} = \alpha y_m^k + (1 - \alpha) u_{y_m}^k$ \label{alg1:line2}
		\Statex 
		All clients \textcolor{red}{\textbf{communicate}} to locally compute 
        \State  $\bar v_{x_m}^{k+1} = \sum_{i=1}^M w_{m,i} v^k_{x_i}$, \ \ \ \ $\bar v_{y_m}^{k+1} = \sum_{i=1}^M w_{m,i} v^k_{y_i}$ \label{alg1:line:comm}
        \State and use \textcolor{blue}{\textbf{local}} method $\mathcal{M}$ to find a solution $(\hat x_m^{k+1}, \hat y_m^{k+1})$ of: \label{alg1_line:subproblem}
		\begin{equation}\label{alg1:subproblem}
			 \min_{x_m}  \max_{y_m} \left\{A_m^k(x_m, y_m) = \lambda\langle  \bar v_{x_m}^{k+1},x_m \rangle + \tfrac{1}{2\eta} \|x_m- x_m^k \|^2 + f_m(x_m,y_m) - \lambda\langle \bar v_{y_m}^{k+1},y_m \rangle
			  - \tfrac{1}{2\eta} \|y_m - y_m^k \|^2\right\},
		\end{equation} 
        \Statex such that $\| \nabla A_m^k (\hat x_m^{k+1}, \hat y_m^{k+1})\|^2  \leq \frac{1}{6\eta^2}\left(\|\hat x_m^{k+1} - x^k\|^2 + \|\hat y_m^{k+1} - y^k\|^2\right)$.
        \Statex \textcolor{blue}{\textbf{Local updates}} for all clients: \State \label{alg1_line:update}$x_m^{k+1} = x_m^k - \eta(\lambda \bar v_m^{k+1} + \nabla_{x} f_m(\hat x_m^{k+1}, \hat y_m^{k+1}))$ \State $y_m^{k+1} = y_m^k - \eta(\lambda \bar v_m^{k+1} - \nabla_{y} f_m(\hat x_m^{k+1}, \hat y_m^{k+1}))$        
		\State\label{alg1_line:acc} $u_{x_m}^{k+1} =  v^k_{x_m} + \alpha(\hat x_m^{k+1} - x_m^k), \ \ u_{y_m}^{k+1} =  v^k_{y_m} + \alpha(\hat y_m^{k+1} - y_m^k)$ 
		\EndFor
	\end{algorithmic}
\end{algorithm} 

For this case we present optimal algorithm (\Cref{alg:sliding_opt_comm}). This method is based on Fast Gradient Descent \cite{nesterov2003introductory} with a proximal operator calculation \eqref{sliding:eq:prox}. The proximal operator is an inexact solution of the auxiliary saddle point subproblem. 
Note that we need to communicate with other devices only in \cref{alg1:line:comm}. This step requires information from the node's neighbors. Furthermore, the subproblem \eqref{alg1:subproblem} is solved locally and separately on each machine (in parallel). Moreover, the stopping criteria for this subproblem $\| \nabla A_m^k (\hat x_m^{k+1}, \hat y_m^{k+1})\|^2  \leq \frac{1}{6\eta^2}\left(\|\hat x_m^{k+1} - x^k\|^2 + \|\hat y_m^{k+1} - y^k\|^2\right)$ is practical due to it does not depend on the solution of the subproblem. For example, Extra Step Method \cite{juditsky2011solving} or Randomized Extra Step Method \cite{alacaoglu2021stochastic} can be applied to solve locally problems.

\textbf{\Cref{alg:sliding_opt_comm} with Extra Step Method.} 
The following theorem states the convergence rate of \Cref{alg:sliding_opt_comm} with Extra Step Method \cite{juditsky2011solving} as a local algorithm for the subproblem \eqref{sliding:eq:prox}.

\begin{theorem}\label{th:sliding_opt_comm}
Let Algorithm \ref{alg:sliding_opt_comm} be applied for solving \eqref{PF} under Assumptions \ref{ass:smooth} and \ref{ass:sc} on local functions $f_m$, with $\sqrt{\frac{\lambda \lambda_{\max}(W)}{\mu}} \leq \frac{L}{\mu}$. Then, to find an $\varepsilon$-solution to the problem \eqref{PF},  \Cref{alg:sliding_opt_comm}  with Extra Step Method requires
\begin{equation*}
	\mathcal{O}\left(\max\left\{1, \sqrt{\frac{\lambda \lambda_{\max}(W)}{\mu}} \right\}\log\frac{1}{\varepsilon}\right) ~~ \text{communication rounds and}
\end{equation*} 
\begin{equation*}
    \mathcal{O}\left(\frac{L}{\mu}\log \frac{L}{\mu}\log\frac{1}{\varepsilon}\right) ~~ \text{local computations on each node}.
\end{equation*} 
\end{theorem}

As expected, the communication complexity of \Cref{alg:sliding_opt_comm} with Extra Step Method is $O\left(\sqrt{\tfrac{\lambda 
\lambda_{\max}(W)}{\mu}}\log{\tfrac{1}{\varepsilon}}\right)$, thus optimal when $\sqrt{\frac{\lambda\lambda_{\max}(W)}{\mu} } = \mathcal{O}\left(\frac{L}{\mu}\sqrt{\chi}\right)$. Also, the local gradient complexity is $\mathcal{O}\left(\frac{L}{\mu}\log \frac{L}{\mu}\log \frac{1}{\varepsilon}\right)$ which is (up to log and constant factors) identical to the lower bound on the local gradient calls. To get an estimate for the proximal oracle, it is enough to note that the subproblem \eqref{sliding:eq:prox} is local and matches the definition of the proximal oracle, then we can solve this problem in one oracle call.

\textbf{\Cref{alg:sliding_opt_comm} with Randomized Extra Step Method.}
 The following theorem states the convergence rate of \Cref{alg:sliding_big_lambda} with with Randomized Extra Step Method \cite{alacaoglu2021stochastic} as a local algorithm.
\begin{theorem} \label{theorem_sum_structure}
Let Algorithm \ref{alg:sliding_opt_comm} be applied for solving \eqref{PF} under Assumptions \ref{ass:smooth} and \ref{ass:sc} on local functions $f_m(x_m, y_m) = \frac{1}{r}\sum_{i=1}^r f_{m, i} (x_m, y_m)$. Then, to find an $\varepsilon$-solution to \eqref{PF} \Cref{alg:sliding_opt_comm} with Randomized Extra Step Method requires
\begin{equation*}
    \mathcal{O}  \left(\sqrt{\frac{\lambda \lambda_{\max}(W)}{\mu}}\log\frac{1}{\varepsilon}\right) \ \ \text{communication rounds and}
\end{equation*}
\begin{equation*}
    \mathcal{\tilde{O}} \left(   \left(r\frac{\lambda \lambda_{\max}(W)}{\mu}+\frac{\sqrt{r} L}{ \mu} \right)\log \frac{1}{\varepsilon} \right) \ \ \text{local oracle calls on each node.}
\end{equation*}
\end{theorem}
One can find the parameter settings for \Cref{alg:sliding_opt_comm} in Appendix \ref{proofs}.
\subsection{Sliding for case  $\tfrac{L}{\mu}\sqrt{\chi} = \mathcal{O}\left(\sqrt{\tfrac{\lambda\lambda_{\max}(W)}{\mu} }\right)$}
\begin{algorithm}[h] 
	\caption{Sliding for large $\lambda$ (personalization parameter)}
	\label{alg:sliding_big_lambda}
	\hspace*{\algorithmicindent} {\bf Parameters:} stepsize $\eta$, $\alpha$\\
	\hspace*{\algorithmicindent} {\bf Initialization:} choose  $ x^0,y^0$, $x^0_m = x^0$, $y^0_m = y^0$  for all nodes
	\begin{algorithmic}[1]
		\For{$k=0,1,2,\ldots$}
		\Statex \textcolor{blue}{\textbf{Local updates}} for all clients: \State  $v^k_{x_m} = x_m^k - \eta \nabla_{x} f(x_m^k, y_m^k)$, \ \ \ \   $v^k_{y_m} = y_m^k + \eta \nabla_{y} f(x_m^k, y_m^k) $ \label{alg_line:big_lambda:2}
		\State Find $\hat U^k$ using method $\mathcal{M}$ with \textcolor{red}{\textbf{communications}}, such that $ \| \hat{U}^k - U^k \|^2 \leq \delta$, where $U^k$ is a solution of: \label{alg_line:big_lambda:3}
		\begin{equation}\label{inner_problem_big_lambda}
		\begin{split}
			\min_{U_x}   \left\{\frac{\lambda}{2} \|\sqrt{W}U_x\|^2  + \frac{1}{2\eta} \|U_x - V_x^k\|^2\right\} ~ \text{and} ~
		    \max_{U_y}  \left\{- \frac{\lambda }{2}\|\sqrt{W}U_y\|^2  - \frac{1}{2\eta}\|U_y - V_y^k\|^2 \right\}
		    \end{split}
		\end{equation}
        \Statex \textcolor{blue}{\textbf{Local updates}} for all nodes:
		\State  
		\label{alg_line:big_lambda:4}
		$
		  x_m^{k+1} =  \hat u^k_{x_m} + \eta \left(\nabla_{x} f(x_m^k, y_m^k)  - \nabla_{x} f(\hat u_{x_m}^k, \hat u_{y_m}^k)\right)
		$
		\State
		\label{alg_line:big_lambda:5}
		$y_m^{k+1} =  \hat u^k_{y_m} - \eta  \left(\nabla_{y} f(x_m^k, y_m^k) - \nabla_{y} f(\hat u_{x_m}^k, \hat u_{y_m}^k)\right)$	
		\EndFor
	\end{algorithmic}
\end{algorithm}
For this case we present Algorithm~\ref{alg:sliding_big_lambda}. This algorithm is the Tseng method \cite{doi:10.1137/S0363012998338806} with a resolvent/proximal operator calculation (\cref{alg_line:big_lambda:3}). Here, as in  \Cref{alg:sliding_opt_comm}, the proximal operator is computed inexactly.  Note that we need to communicate with other devices only when we solve the problem \eqref{inner_problem_big_lambda} and need to multiply by the matrix $W$. The problem \eqref{inner_problem_big_lambda} is divided into two minimization subproblems, by $X$, and by $Y$. Hence, the problem \eqref{inner_problem_big_lambda} is solved by Fast Gradient Descent.  Further,  we note that the algorithm's steps in lines \ref{alg_line:big_lambda:2}, \ref{alg_line:big_lambda:4}, and \ref{alg_line:big_lambda:5} are local and separable on each machine. The following theorem states the convergence rate of \Cref{alg:sliding_big_lambda} with Accelerated Gradient Descent.
\begin{theorem}\label{th:sliding_big_lambda}
Consider the problem \eqref{PF} under  Assumptions \ref{ass:smooth} - \ref{ass:sc}. Then, to find an $\varepsilon$-solution to the problem \eqref{PF},  \Cref{alg:sliding_big_lambda} with Accelerated Gradient Descent requires
\begin{equation*}
	\mathcal{O}\left(\frac{L}{\mu}\sqrt{\chi}\log\frac{L}{\mu}\log\frac{1}{\varepsilon}\right) ~~ \text{communication rounds and}
\end{equation*}
\begin{equation*}
    \mathcal{O}\left(\frac{L}{\mu}\log\frac{1}{\varepsilon}\right) ~~ \text{local oracle calls on each node.}
\end{equation*} 
\end{theorem}

As expected, the communication complexity of \Cref{alg:sliding_big_lambda} with Fast Gradient Descent is $\mathcal{O}\left(\tfrac{L}{\mu}\sqrt{\chi}\log{\tfrac{1}{\varepsilon}}\right)$ in the strongly convex -- strongly concave case. It is optimal when $L \chi  = \mathcal{O}\left( \lambda \lambda_{\max}(W)\right)$. The local gradient complexity is $\tilde{\mathcal{O}}\left(\tfrac{L}{\mu}\right)$, which is, up to log-factors identical to the lower bound for the local calls. One can find the parameter settings for \Cref{alg:sliding_big_lambda} in Appendix \ref{proofs}.

\begin{algorithm}[h]
	\caption{Randomized for Decentralized Min-Max (RDMM)}
	\label{alg_sum}
	\hspace*{\algorithmicindent} {\bf Parameters:} stepsize $\gamma$, probability $p$, probability $\rho$\\
	\hspace*{\algorithmicindent} {\bf Initialization:} choose  $ x^0,y^0$, $x^0_m = x^0$, $y^0_m = y^0$  for all $m$
	\begin{algorithmic}[1]
\For {$k=0,1, 2, \ldots$ }
\Statex \textcolor{blue}{\textbf{Local updates}} for all clients:
\State $\bar x_m^k = (1 - \rho) x_m^k + \rho u^k_{x_m}$, \ \  $\bar y_m^k = (1 - \rho) y_m^k + \rho u^k_{y_m}$
\Statex All devices \textcolor{red}{\textbf{communicate}} to locally compute: 
\State $\bar u_{x_m}^k = \lambda\sum_{i=1}^M w_{m,i} u_{x_m}^k$, \ \  $\bar u_{y_m}^k = \lambda\sum_{i=1}^M w_{m,i} u_{x_m}^k$
\Statex \textcolor{blue}{\textbf{Local updates}} for all clients:
\State $x_m^{k+\frac{1}{2}} =  \bar x_m^k - \eta (\nabla_x f(u^k_{x_m}, u^k_{y_m}) + \bar u_{x_m}^k)$, 
\State $y_m^{k+\frac{1}{2}} = \bar y_m^k - \eta ( - \nabla_y f(u^k_{x_m}, u^k_{y_m}) + \bar u_{y_m}^k) $
\Statex Generate $\xi^k =  \begin{cases}
 1,&  \text{with probability} ~~ 1 - p \\
0 ,& \text{with probability} ~~ p
\end{cases},$
\label{alg_sum:step5}
\Statex If $\xi^k = 0$ all devices \textcolor{red}{\textbf{communicate}} and compute:  \label{alg_sum:step6}
\State \ \ \ $g_{x_m}^k = \frac{\lambda}{p}\sum_{i=1}^M w_{m,i} \left(x^{k+\frac{1}{2}}_{m} - u_{x_m}^k\right)$, \ \ \ $g_{y_m}^k = \frac{\lambda}{p}\sum_{i=1}^M w_{m,i} \left(y^{k+\frac{1}{2}}_{m} - u_{y_m}^k\right)$
\Statex If $\xi^k = 1$ all devices make \textcolor{blue}{\textbf{local computations}}: \label{alg_sum:step9}
\Statex \ \ \  Generate an vector of indexes $\hat{\xi}^k_m$ according to distribution $Q$
\State \ \ \ $g_{x_m}^k = \frac{1}{1-p}\left(\nabla_x f_{\hat{\xi}_m^k}\left(x_m^{k+\frac{1}{2}}, y_m^{k+\frac{1}{2}}\right) - \nabla_x f_{\hat{\xi}_m^k}(u_{x_m}^{k}, u_{y_m}^{k})\right)$
\State \ \ \ $g_{y_m}^k = - \frac{1}{1-p}\left(\nabla_
y f_{\hat{\xi}_m^k}\left(x_m^{k+\frac{1}{2}}, y_m^{k+\frac{1}{2}}\right) - \nabla_y f_{\hat{\xi}_m^k}(u_{x_m}^{k}, u_{y_m}^{k})\right)$
\Statex \textcolor{blue}{\textbf{Local updates}} for all clients:
\State  $x_m^{k+1} =  \bar x_m^k - \eta \left( g_{x_m}^k +  \nabla_{x} f(u^k_{x_m}, u^k_{y_m})+ \bar u^k_{x_m}\right)$, 
\State $y_m^{k+1} =  \bar y_m^k - \eta  \left(g
_{y_m}^k -  \nabla_y f(u^k_{x_m}, u^k_{y_m}) +  \bar u^k_{y_m}\right)$

\Statex Generate $\delta^{k}=  \begin{cases}
1,&  \text{with probability} ~~ 1 - \rho \\
0 ,& \text{with probability} ~~ \rho
\end{cases},$
\State $u^{k+1}_{x_m} = \delta^{k} u^k_{x_m} + (1 - \delta^{k}) x_m^{k+1}$, \ \ \ $u^{k+1}_{y_m} = \delta^{k}  u^k_{y_m} + (1 - \delta^{k}) y_m^{k+1}$
\EndFor
\end{algorithmic}
\end{algorithm}

\subsection{Local method via Variance Reduction}

Our first two methods make several iterations between communications when $\lambda$ is small (or vice versa, for big $\lambda$ make some communications between one local iteration). The following method (\Cref{alg_sum}) is also sharpened on the alternation of local iterations and communications, but it makes them more evenly. Our method is similar to the randomized local methods (for example, as the method from \cite{hanzely2020federated}), but it uses not only importance sampling, but also implicit variance reduction technique \cite{alacaoglu2021stochastic}.

The following theorem states the convergence rate of \Cref{alg_sum}.

\begin{theorem}\label{theorem_randomized}
Let Algorithm \ref{alg_sum} be applied for solving \eqref{PF} under Assumptions \ref{ass:smooth} and \ref{ass:sc} on local functions $f_{m}(x_m, y_m) = \frac{1}{r}\sum_{i=1}^rf_{m, i}(x_m, y_m)$. Then, to find an $\varepsilon$-solution to \eqref{PF} \Cref{alg_sum} requires
\begin{equation*}
    \mathcal{O}  \left(\frac{\lambda \lambda_{\max}(W)}{\mu}\log\frac{1}{\varepsilon}\right) \ \ \text{communication rounds and}
\end{equation*}
\begin{equation*}
    \mathcal{\tilde{O}} \left(   \left(r+\frac{\sqrt{r(L^2 + \lambda^2 \lambda^2_{\max}(W))}}{ \mu} \right)\log \frac{1}{\varepsilon} \right) \ \ \text{local oracle calls on each node.}
\end{equation*}

\end{theorem}
One can find the parameter settings for \Cref{alg_sum} in Appendix \ref{proofs}. Below we will discuss and compare all these methods.

\section{Experiments}\label{exps}
We divided our experiments into two parts: 1) toy experiments on strongly convex -- strongly concave bilinear saddle point problems to verify the theoretical results and 2) adversarial training of neural networks to compare deterministic (\Cref{alg:sliding_opt_comm}) and stochastic  (\Cref{alg_sum}) approaches.

\subsection{Toy experiments}
We conduct our toy experiments on bilinear problem:
\begin{equation}
    \label{bilinear}
     \textstyle{f_m(x,y) =  x^\top A_m y + a^\top_m x + b^\top_m y + \tfrac{\beta}{2}  \|x\|^2 -  \tfrac{\beta }{2} \|y\|^2},
\end{equation}
where $A_m \in \R^{d \times d}$, $a_m, b_m \in \R^d$. We take $d=100$ and generate positive definite matrices $A_m$ and vectors $a_m, b_m$ randomly, such that $L = 5$. We take $M =16$ and $\beta = 0,1$. We use three topologies of network: complete graph, star and ring. In all experiments, we compare the algorithms in the rate of convergence the solution in terms of the number of communications (for Algorithm~\ref{alg:sliding_opt_comm} -- outer iterations, for Algorithm~\ref{alg:sliding_big_lambda} -- inner iterations).
 
\begin{figure}[h]
\begin{minipage}{1.\textwidth}
\begin{minipage}{0.35\textwidth}
  \centering
\includegraphics[width =  \textwidth ]{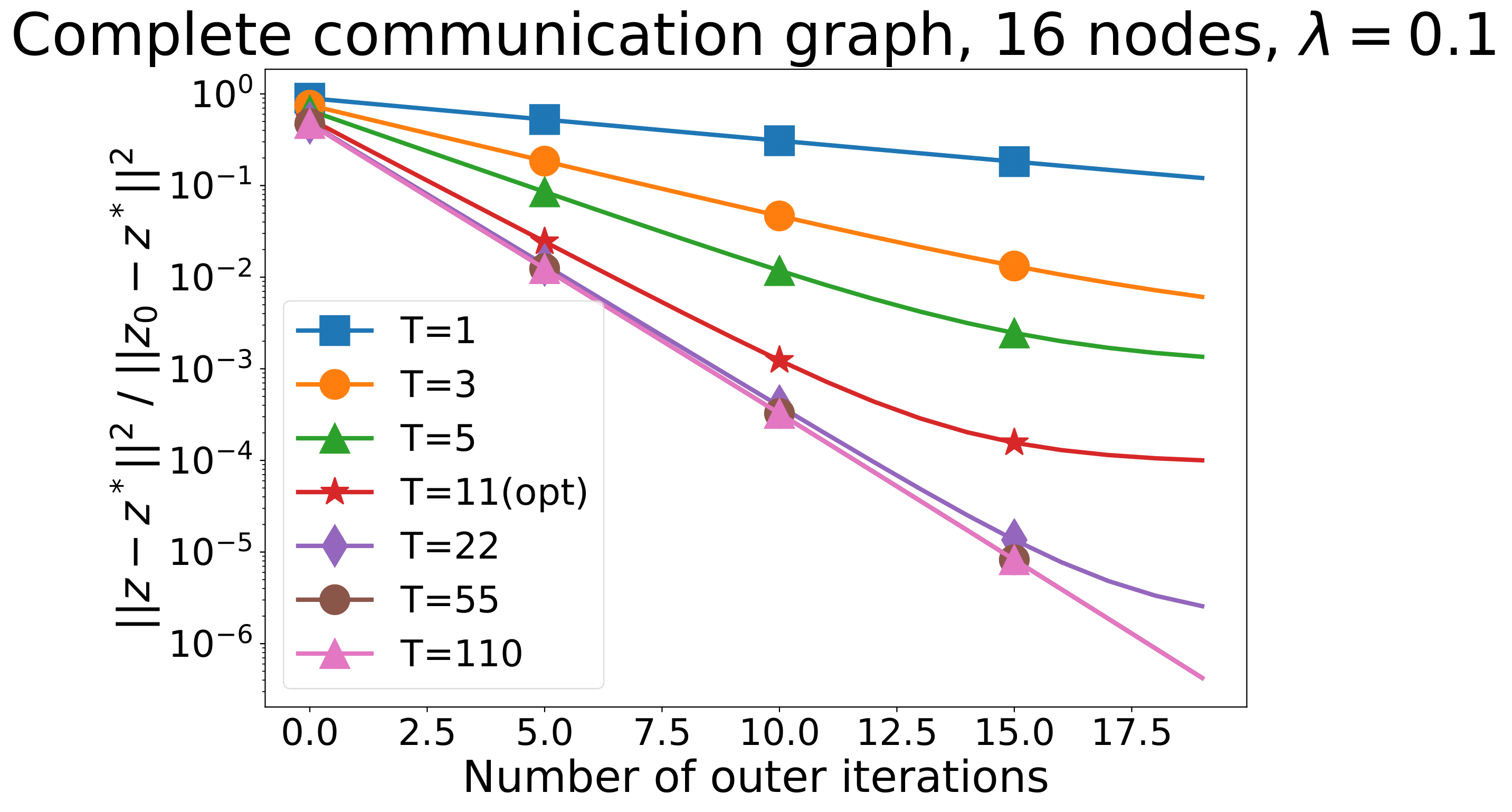}
\end{minipage}%
\begin{minipage}{0.32\textwidth}
  \centering
\includegraphics[width =  \textwidth ]{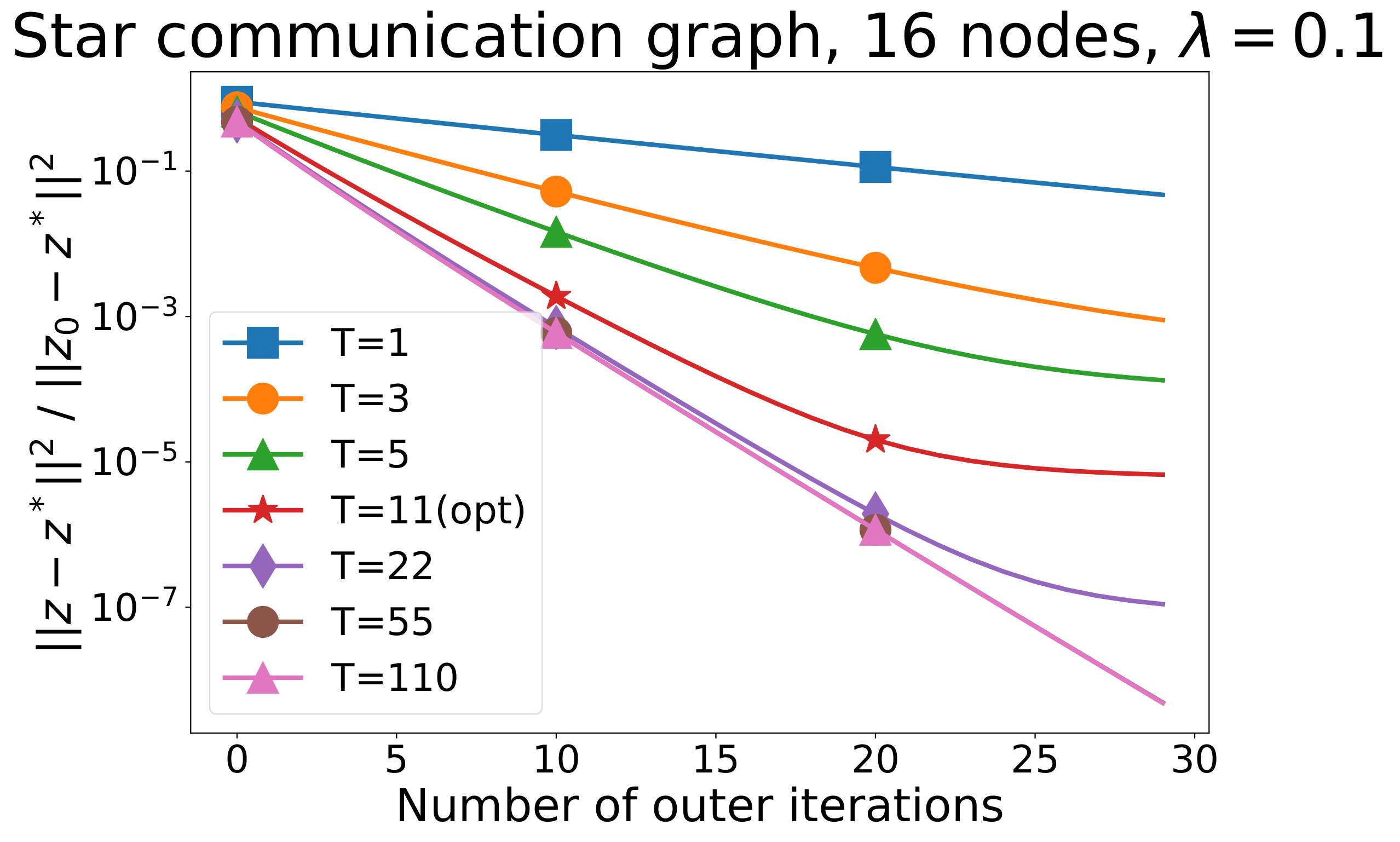}
\end{minipage}%
\begin{minipage}{0.32\textwidth}
  \centering
\includegraphics[width =  \textwidth ]{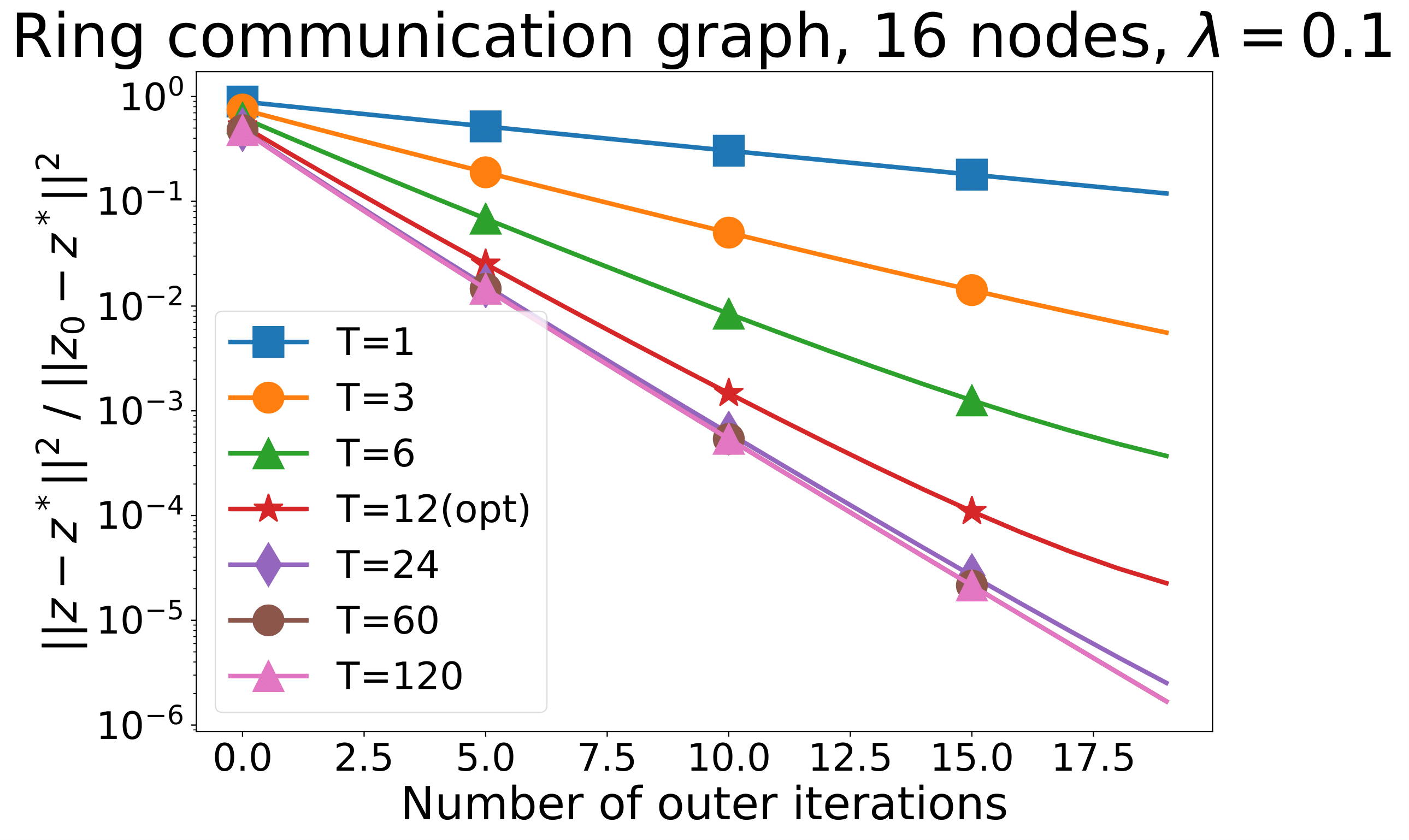}
\end{minipage}%
\\
\begin{minipage}{0.35\textwidth}
  \centering
\includegraphics[width = \textwidth ]{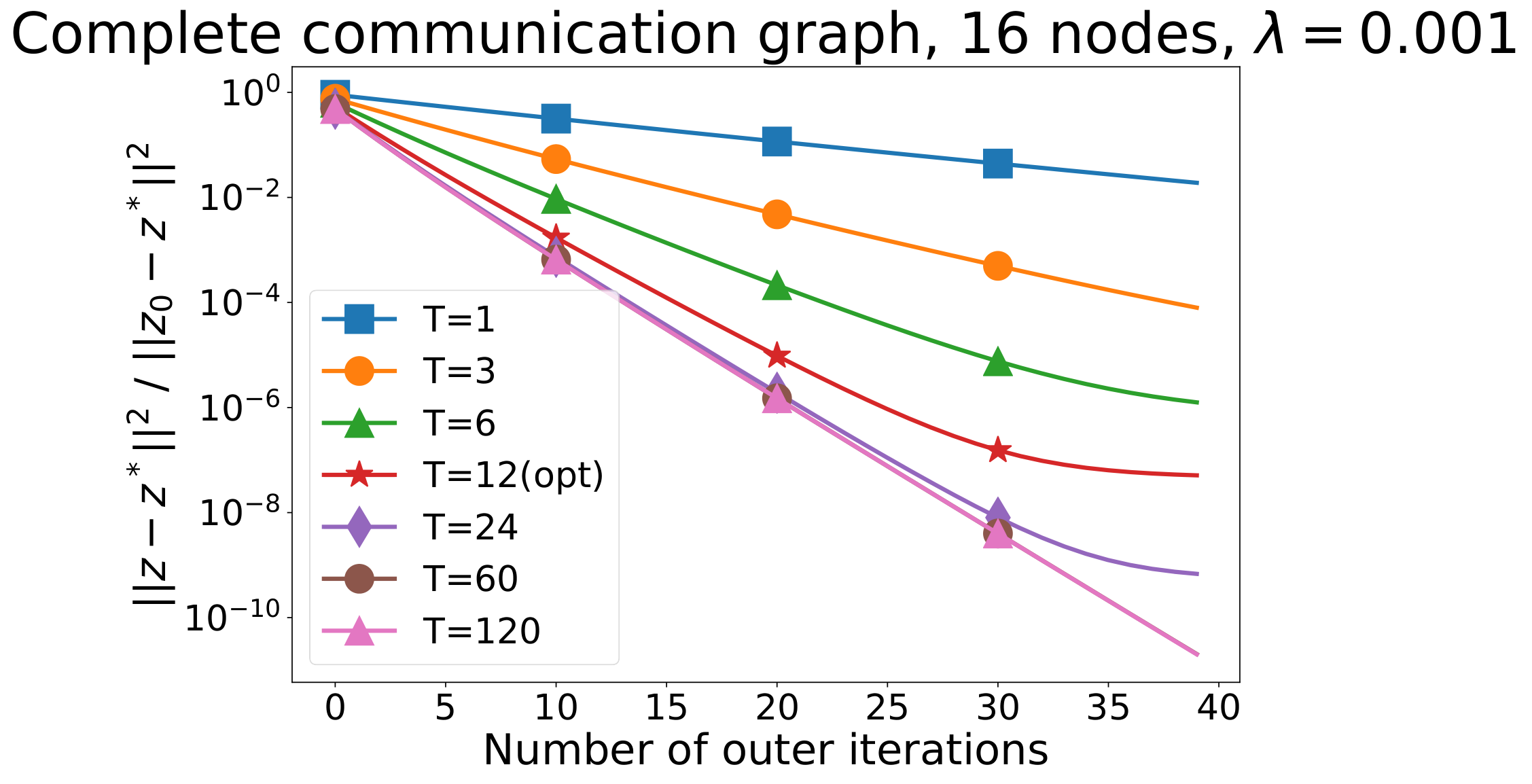}
\end{minipage}%
\begin{minipage}{0.32\textwidth}
  \centering
\includegraphics[width = \textwidth ]{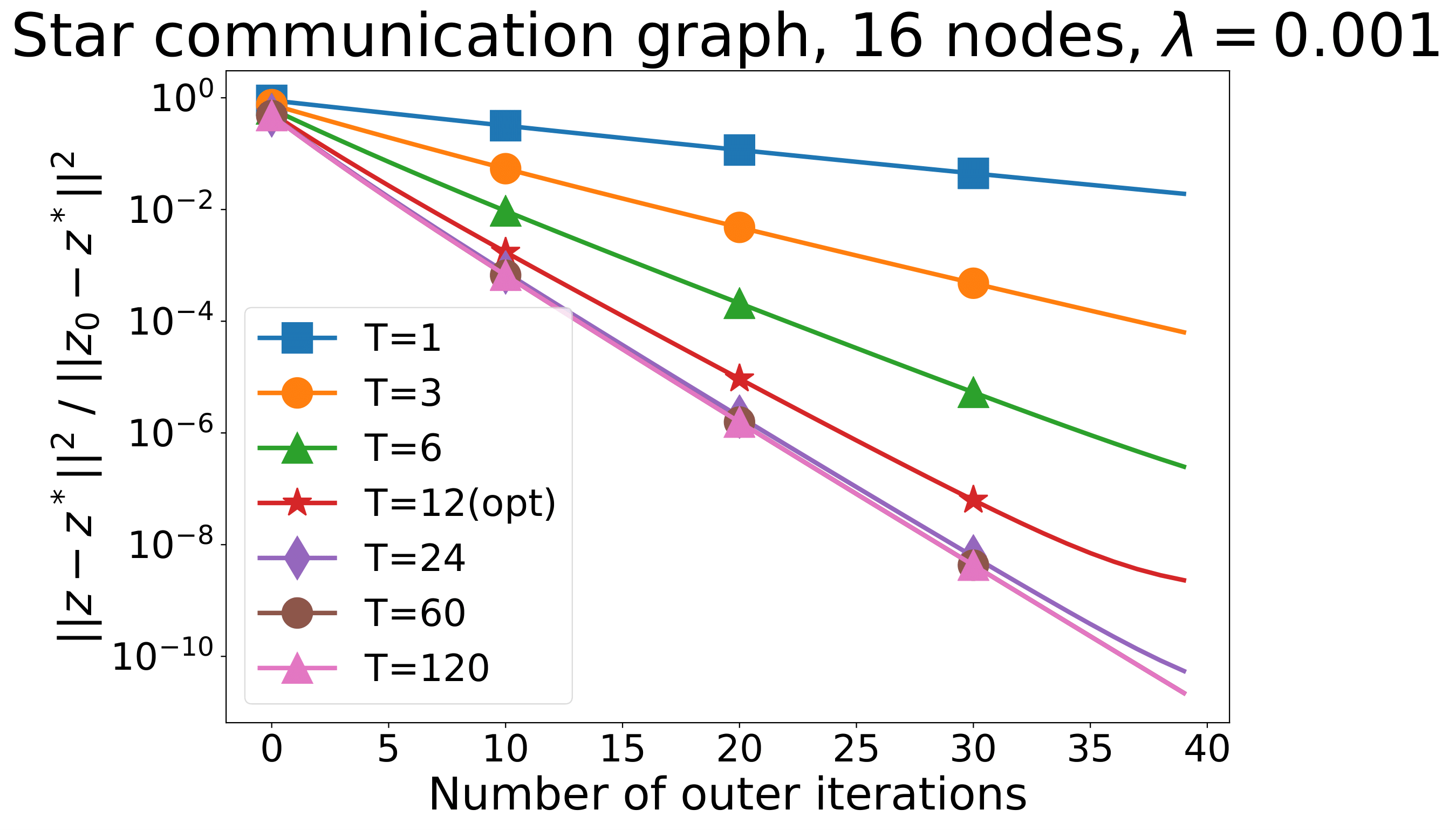}
\end{minipage}%
\begin{minipage}{0.32\textwidth}
  \centering
\includegraphics[width = \textwidth ]{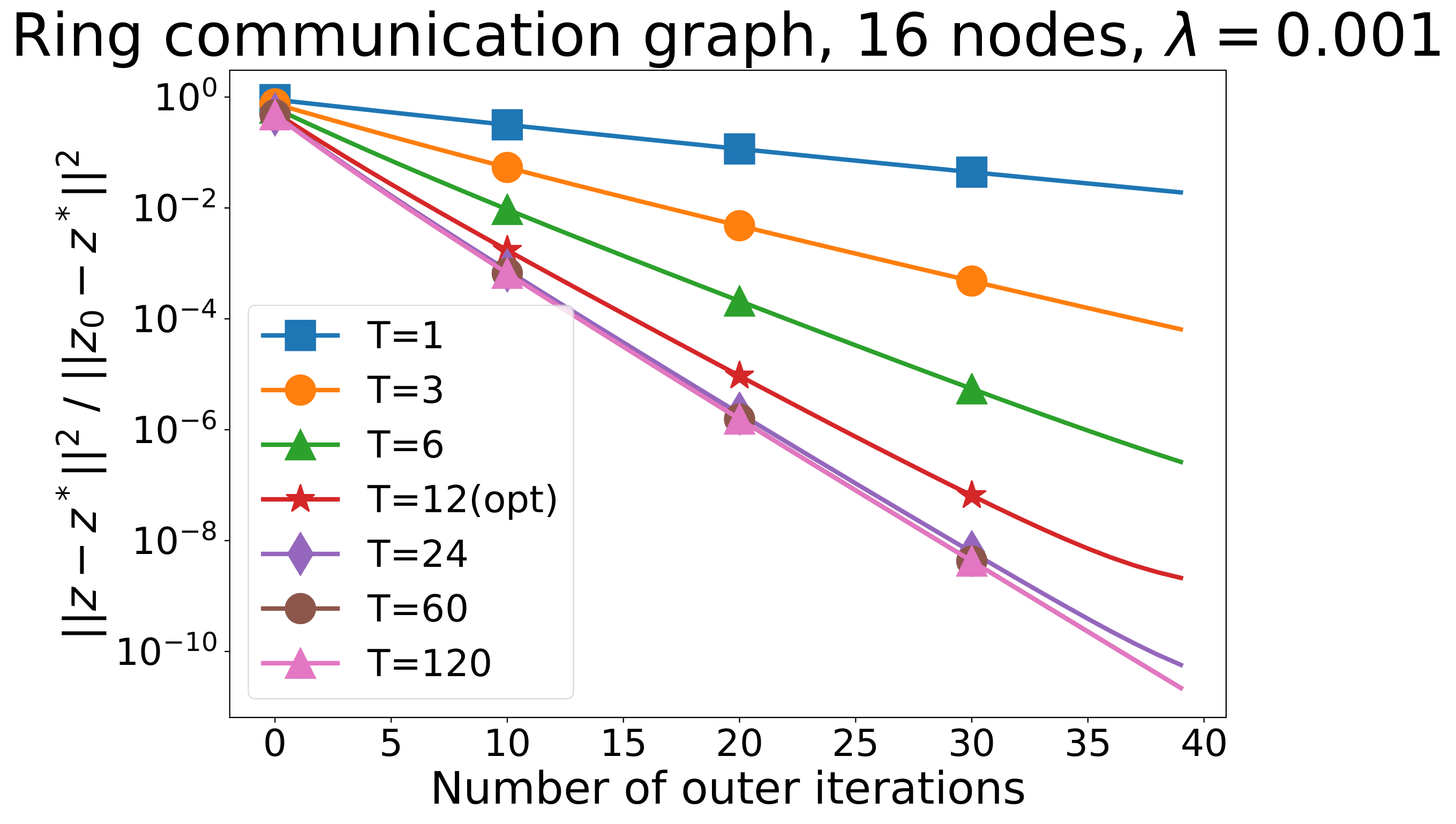}
\end{minipage}
\\
\begin{minipage}{0.35\textwidth}
  \centering
\scriptsize{(a) complete}
\end{minipage}%
\begin{minipage}{0.32\textwidth}
  \centering
\scriptsize{(b) star}
\end{minipage}%
\begin{minipage}{0.32\textwidth}
  \centering
\scriptsize{(c) ring}
\end{minipage}%
\captionof{figure}{Comparison of Algorithm \ref{alg:sliding_opt_comm} with different $T$ on different networks for \eqref{PF}+\eqref{bilinear} with $\lambda = 0,1$.}
\label{fig:toy1}
\end{minipage}

\begin{minipage}{1.\textwidth}
\begin{minipage}{0.35\textwidth}
  \centering
\includegraphics[width =  \textwidth]{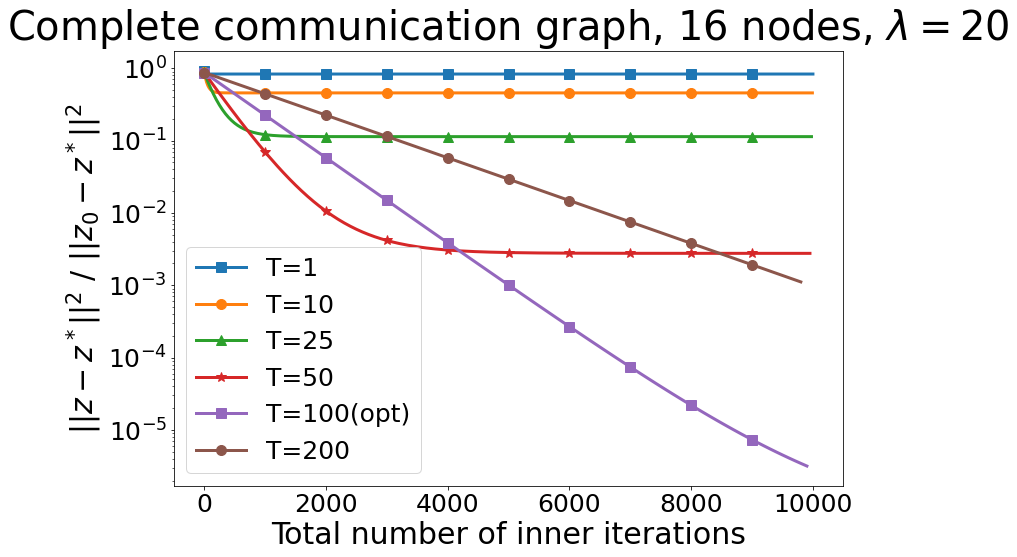}
\end{minipage}%
\begin{minipage}{0.32\textwidth}
  \centering
\includegraphics[width =  \textwidth ]{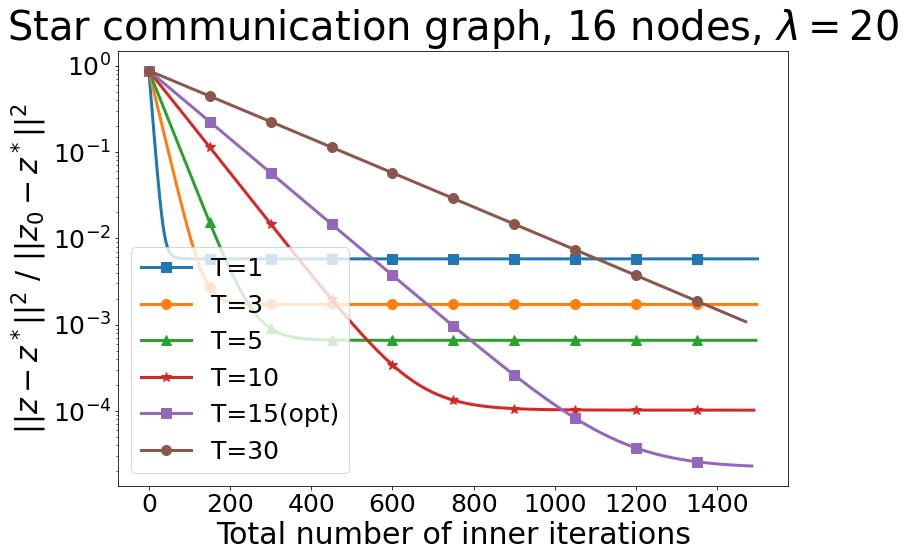}
\end{minipage}%
\begin{minipage}{0.32\textwidth}
  \centering
\includegraphics[width =  \textwidth ]{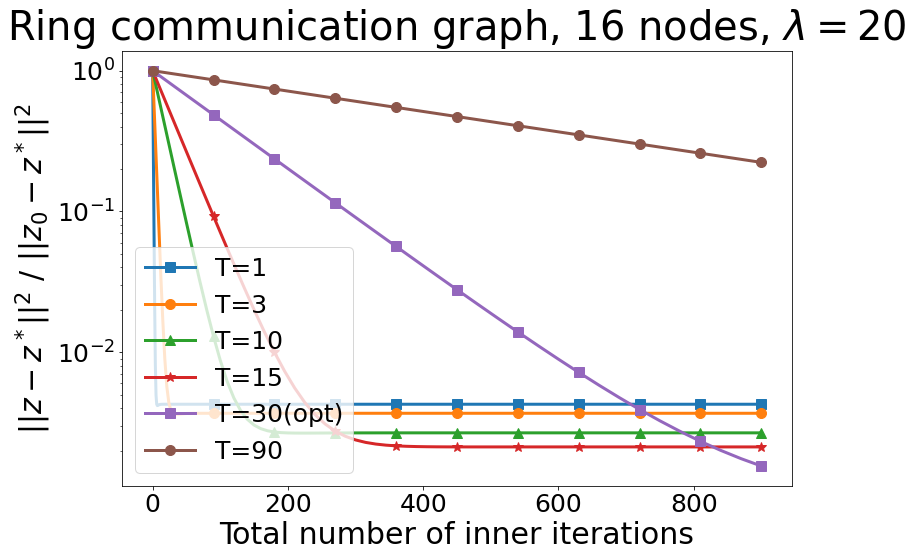}
\end{minipage}%
\\
\begin{minipage}{0.35\textwidth}
  \centering
\includegraphics[width = \textwidth]{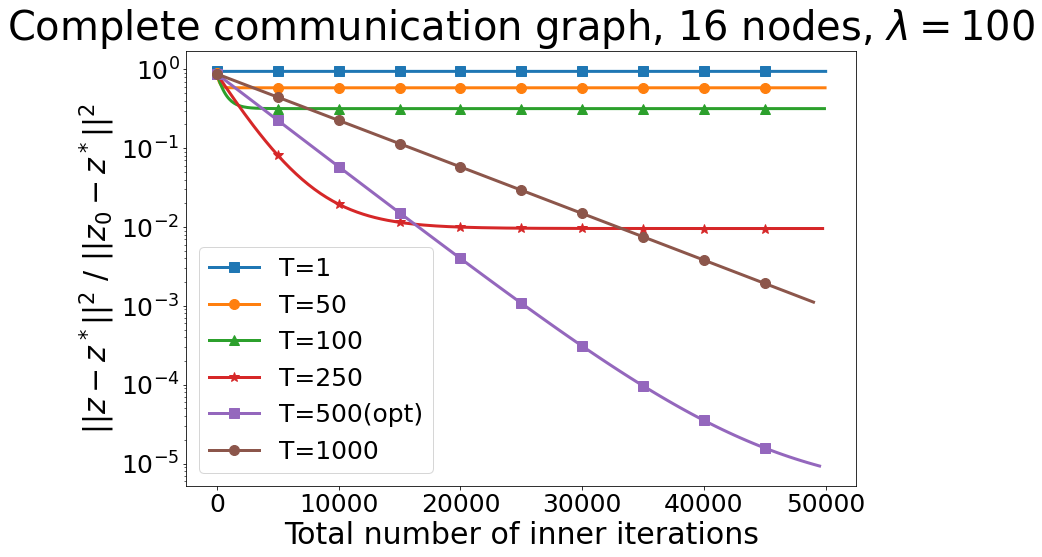}
\end{minipage}%
\begin{minipage}{0.32\textwidth}
  \centering
\includegraphics[width = \textwidth ]{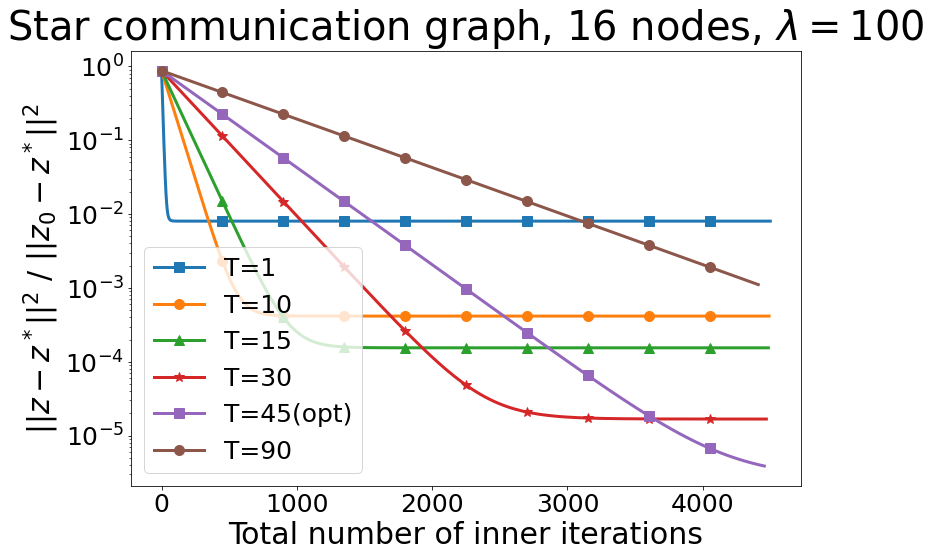}
\end{minipage}%
\begin{minipage}{0.32\textwidth}
  \centering
\includegraphics[width = \textwidth ]{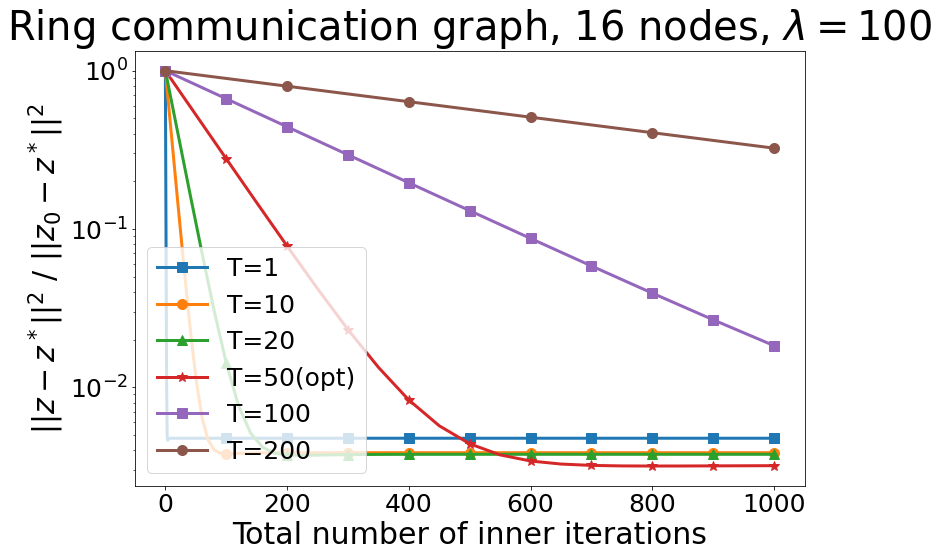}
\end{minipage}%
\\
\begin{minipage}{0.35\textwidth}
  \centering
\scriptsize{(a) complete}
\end{minipage}%
\begin{minipage}{0.32\textwidth}
  \centering
\scriptsize{(b) star}
\end{minipage}%
\begin{minipage}{0.32\textwidth}
  \centering
\scriptsize{(c) ring}
\end{minipage}%
\captionof{figure}{Comparison of Algorithm \ref{alg:sliding_big_lambda} with different $T$ on different networks for \eqref{PF}+\eqref{bilinear} with $\lambda = 20$.}
\label{fig:toy2}  
\end{minipage}
\end{figure}

$\bullet$ In the first experiment, we compare Algorithm \ref{alg:sliding_opt_comm} with $\eta = \frac{1}{3\sqrt{\lambda \lambda_{\max}(W)\beta}}$ (as in theory) and the different numbers of inner iterations $T$ (for the subproblem \eqref{sliding:eq:prox}). See results on Figure \ref{fig:toy1}.  

We see that from the point of view of the number of communications, the theoretically optimal number of innner steps $T_{\text{opt}}$ is almost optimal in practice. It is also seen that there is a certain limiting $T$ after which an increase in the number of inner iterations does not give a particular acceleration in terms of communications (outer iterations).

$\bullet$ In the second experiment, we compare Algorithm \ref{alg:sliding_big_lambda} with $\eta = \frac{1}{2L}$ (as in theory) and the different numbers of inner iterations $T$ (for the subproblem \eqref{inner_problem_big_lambda}). See results on Figure \ref{fig:toy2}.

We see that from the point of view of the number of communications, the large number innner steps $T> T_{\text{opt}}$ only slows down the convergence. On the contrary, a small number of inner iterations $T< T_{\text{opt}}$ accelerates, but degrades the accuracy of the solution. The optimal $T_{\text{opt}}$ gives a good balance of accuracy and rate.

\begin{figure}[h]
\begin{minipage}{0.34\textwidth}
  \centering
\includegraphics[width =  0.8\textwidth]{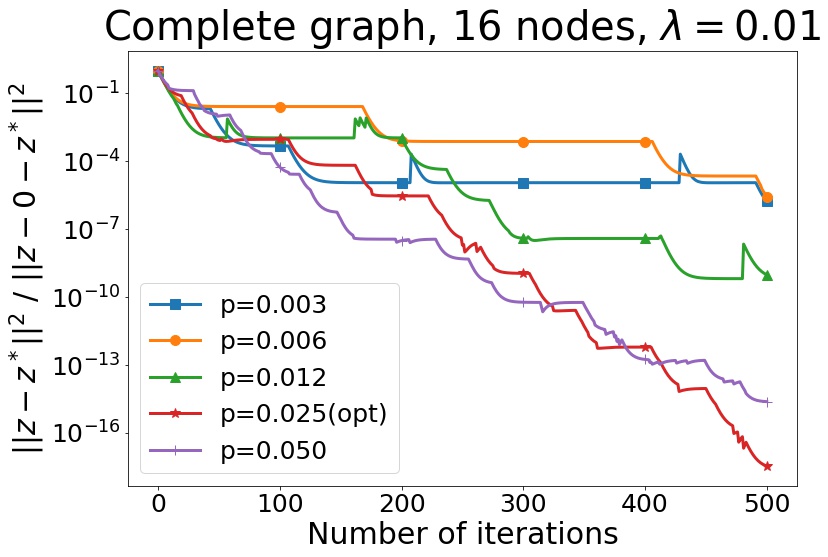}
\end{minipage}%
\begin{minipage}{0.30\textwidth}
  \centering
\includegraphics[width =  0.88\textwidth ]{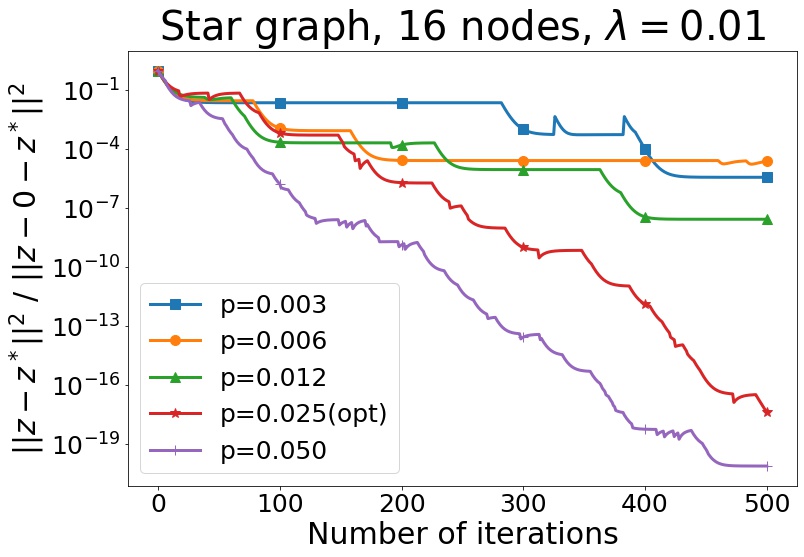}
\end{minipage}%
\begin{minipage}{0.30\textwidth}
  \centering
\includegraphics[width =  0.88\textwidth ]{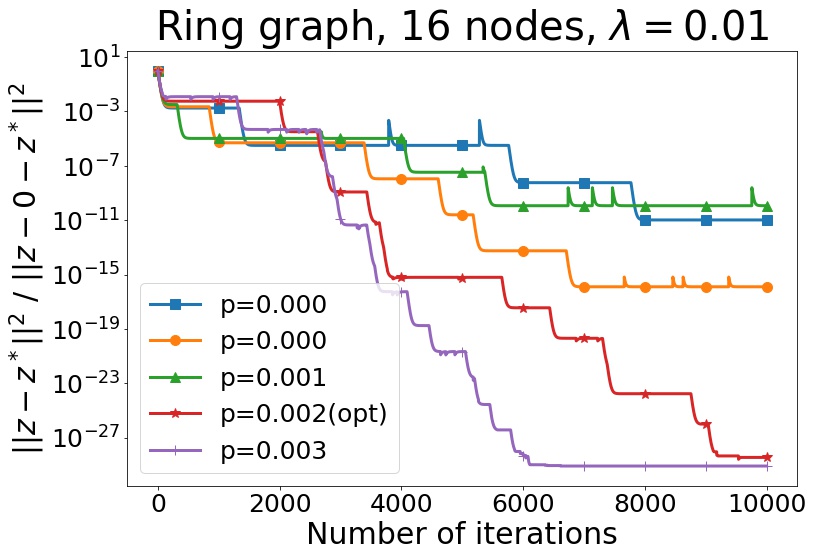}
\end{minipage}%
\\
\begin{minipage}{0.34\textwidth}
  \centering
\includegraphics[width =  0.8\textwidth]{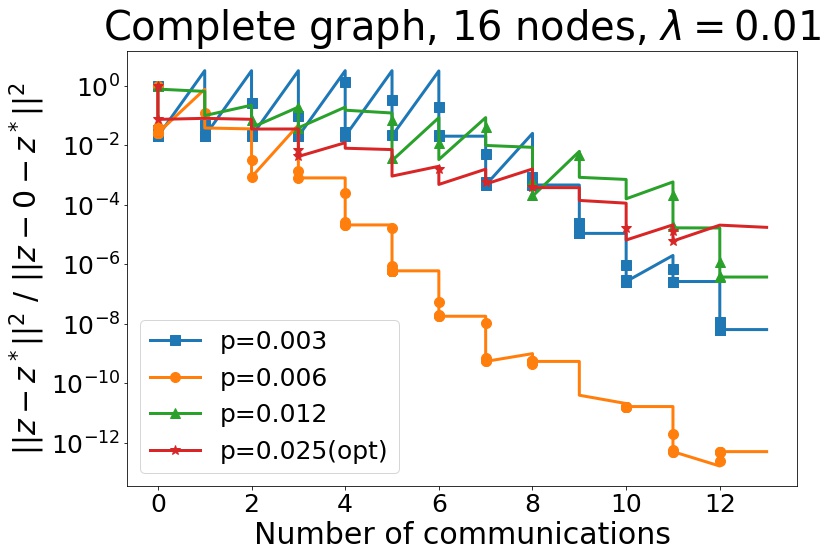}
\end{minipage}%
\begin{minipage}{0.30\textwidth}
  \centering
\includegraphics[width =  0.88\textwidth ]{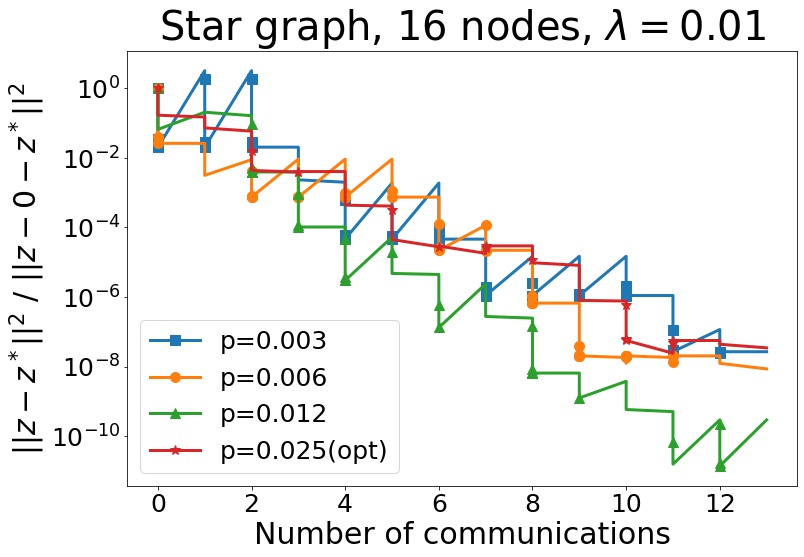}
\end{minipage}%
\begin{minipage}{0.30\textwidth}
  \centering
\includegraphics[width =  0.88\textwidth ]{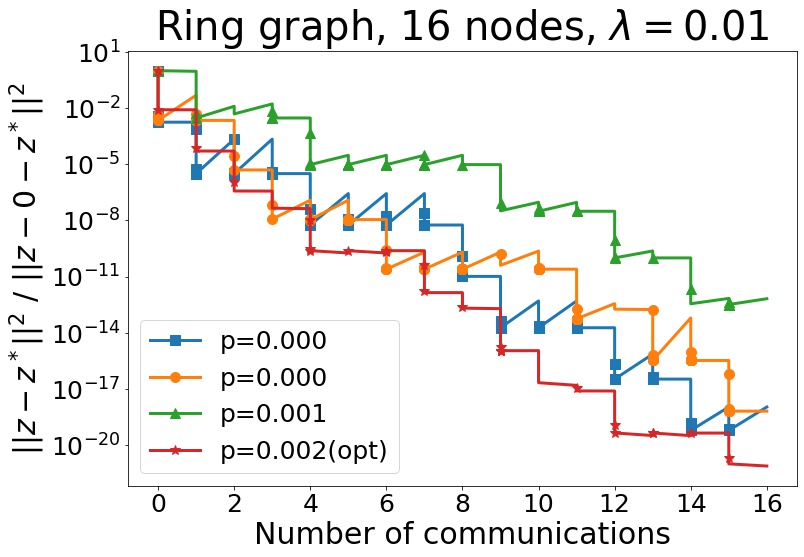}
\end{minipage}%
\\
\begin{minipage}{0.34\textwidth}
  \centering
(a) complete
\end{minipage}%
\begin{minipage}{0.30\textwidth}
  \centering
(b) star
\end{minipage}%
\begin{minipage}{0.30\textwidth}
  \centering
(c) ring
\end{minipage}%
\caption{Comparison of \Cref{alg_sum} with different $p = \rho$ on different networks for \eqref{PF}+\eqref{bilinear} with $\lambda = 0,01$.\\ {\bf Top:} in terms of all iterations, {\bf bottom:} in terms of communications.}
\label{fig:toy5}
\end{figure}

\begin{figure}[h]
\begin{minipage}{0.34\textwidth}
  \centering
\includegraphics[width =  0.8\textwidth]{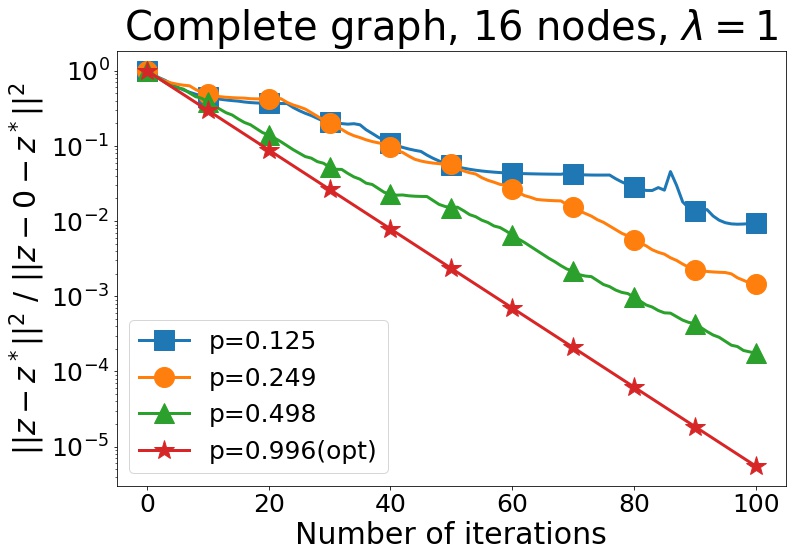}
\end{minipage}%
\begin{minipage}{0.30\textwidth}
  \centering
\includegraphics[width =  0.88\textwidth ]{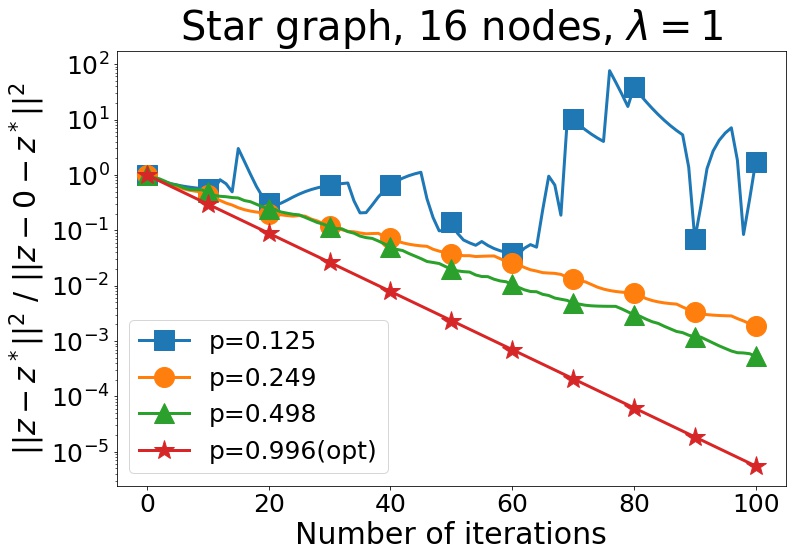}
\end{minipage}%
\begin{minipage}{0.30\textwidth}
  \centering
\includegraphics[width =  0.88\textwidth ]{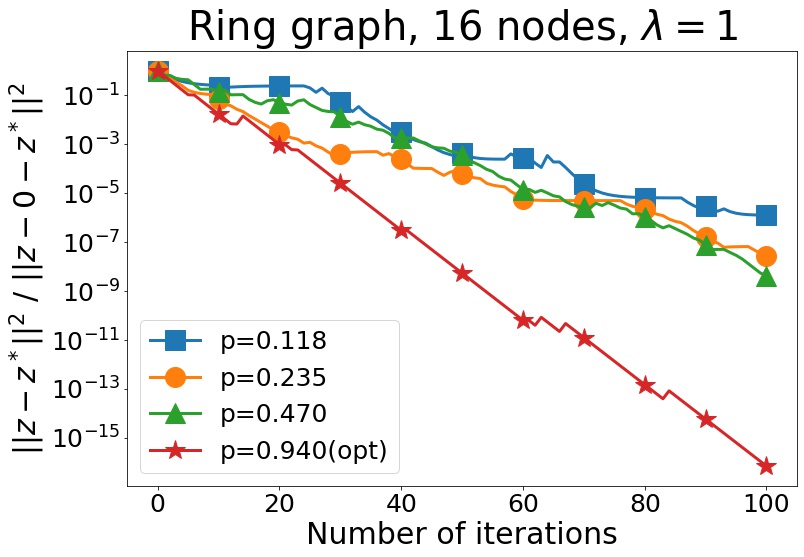}
\end{minipage}%
\\
\begin{minipage}{0.34\textwidth}
  \centering
\includegraphics[width =  0.8\textwidth]{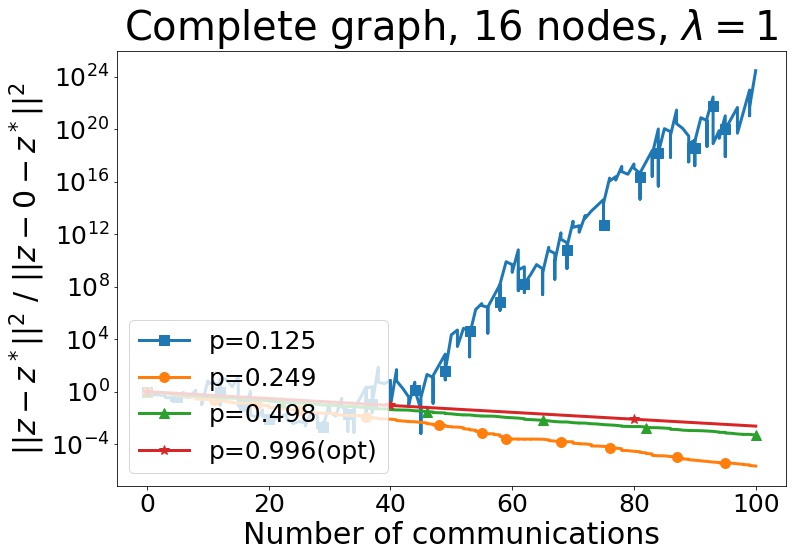}
\end{minipage}%
\begin{minipage}{0.30\textwidth}
  \centering
\includegraphics[width =  0.88\textwidth ]{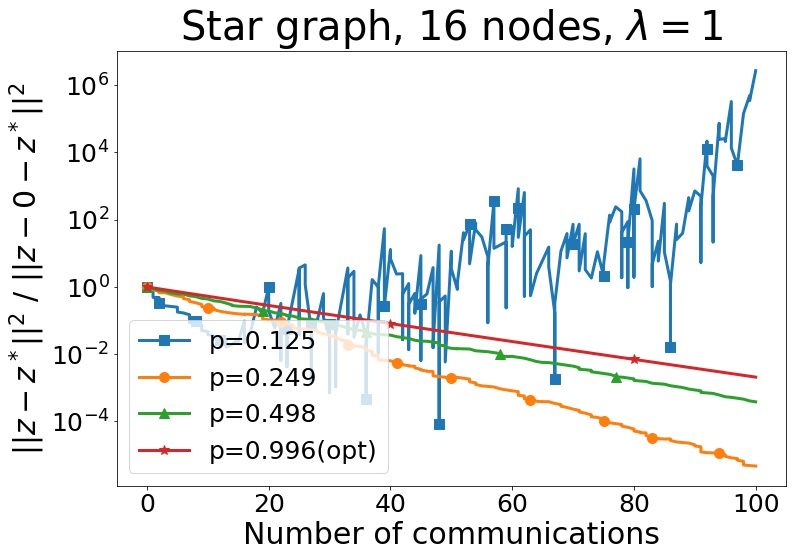}
\end{minipage}%
\begin{minipage}{0.30\textwidth}
  \centering
\includegraphics[width =  0.88\textwidth ]{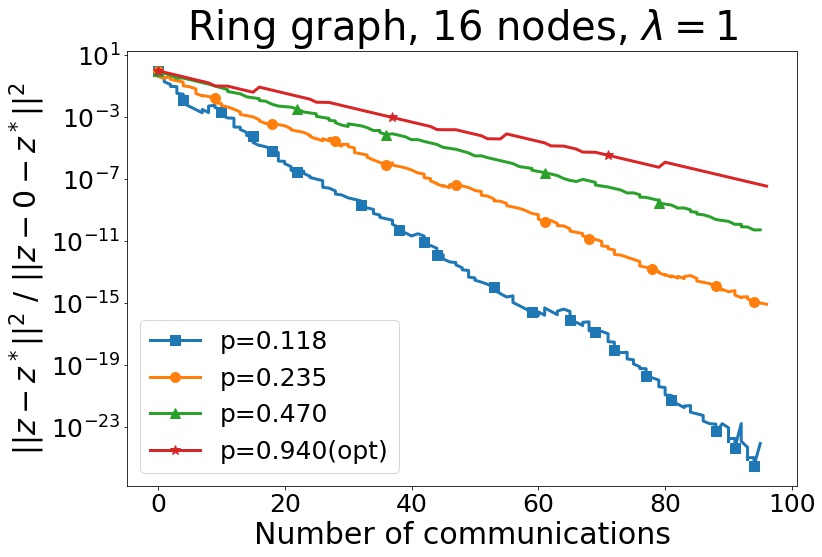}
\end{minipage}%
\\
\begin{minipage}{0.34\textwidth}
  \centering
(a) complete
\end{minipage}%
\begin{minipage}{0.30\textwidth}
  \centering
(b) star
\end{minipage}%
\begin{minipage}{0.30\textwidth}
  \centering
(c) ring
\end{minipage}%
\caption{Comparison of \Cref{alg_sum} with different $p = \rho$ on different networks for \eqref{PF}+\eqref{bilinear} with $\lambda = 1$. \\
{\bf Top:} in terms of all iterations, {\bf bottom:} in terms of communications.}
\label{fig:toy6}
\end{figure}

$\bullet$ In the third experiment, we compare \Cref{alg_sum} for problem with $r = 1$ with $\eta = \frac{\sqrt{p}}{2(L + \lambda \lambda_{\max}(W))}$ (as in theory) and the different probabilities $ p = \rho$. See results on Figures \ref{fig:toy5} and \ref{fig:toy6}.

In terms of the number of communications, the optimal value is $\rho = p = \frac{\lambda^2 \lambda^2_{\max}(W)}{\lambda^2 \lambda^2_{\max}(W) + L^2}$. 
But it can be seen on Figures \ref{fig:toy5} (bottom) and \ref{fig:toy6}(bottom), the probability (frequency of communications) can be reduced. But the optimal $\rho = p = \frac{\lambda^2 \lambda^2_{\max}(W)}{\lambda^2 \lambda^2_{\max}(W) + L^2}$ outperforms the smaller probabilities in terms of the total number of iterations.

\subsection{Intuition for Neural Networks} \label{intuition}

We would like to understand the intuition and adapt the proposed algorithms for federated training of neural networks. 
For simplicity, we will consider only the centralized case. In this situations,  the regularizer $\frac{\lambda}{2} \| \sqrt{W}X\|^2 = \frac{\lambda}{2} \sum_{m=1}^M \|x_m - \bar x \|^2$ and $\lambda_{\max}(W) = 1$. Additionally, let us define the smoothness constant of the neural network for particular dataset as follows: $L = \frac{1}{\eta_{\text{opt}}}$, where $\eta_{\text{opt}}$ is the learning rate that is recommended to be used 
by concrete algorithm 
for given dataset and neural network architecture
(for example, such information can be found in tutorials
or open-source codes). This definition comes from the standard results, that for smooth functions the stepsize $\sim \frac{1}{L}$. We do not say that this is a good definition of $L$, but we only need it for the intuition. We also mention that we are interested in the case when $\lambda \ll L$.

Now, let us start parsing and adapting the \Cref{alg:sliding_opt_comm} and \Cref{alg_sum}.

\textbf{Algorithm \ref{alg:sliding_opt_comm}.} Let us take $\alpha = 1$ and $\eta = \frac{1}{ \lambda \theta}$ in Algorithm \ref{alg:sliding_opt_comm}, with constant $\theta > 1$. 
Then, on \cref{alg1:line2} $v_{x_m}^k = x_m^k$, $v_{y_m}^k = y_m^k$ and on \cref{alg1:line:comm} $\bar v_{x_m}^k = x_m^k - \bar x^k$, $\bar v_{y_m}^k = y_m^k - \bar y^k$. Next, on \cref{alg1_line:subproblem}, local models are trained taking into account the regularizer:
\begin{align*}
&\min_{x_m}  \max_{y_m} \left\{\lambda\langle  x_m^k - \bar x^{k},x_m \rangle + \frac{\lambda\theta}{2} \|x_m- x_m^k \|^2 + f_m(x_m,y_m) - \lambda\langle  y^k_m - \bar y^{k},y_m \rangle
			  - \frac{\lambda \theta}{2} \|y_m - y_m^k \|^2\right\}.
\end{align*}
This problem is equal (in terms of output solutions $(x_m,y_m)$) to the following problem: 
\begin{align}\label{ifnn:subproblem}
&\min_{x_m}  \max_{y_m}  \left\{\frac{\lambda\theta}{2} \|x_m- \bar x_m^k \|^2 + f_m(x_m,y_m)
	- \frac{\lambda \theta}{2} \|y_m - \bar y_m^k \|^2\right\},
\end{align}
where 
\begin{equation}\label{trust}
    \bar x_m^k = \tfrac{1}{\theta}\bar x^k + (1 - \tfrac{1}{\theta})x_m^k, \ \ \  \bar y_m^k = \tfrac{1}{\theta}\bar y^k + (1 - \tfrac{1}{\theta})y_m^k.
\end{equation}
It is evident that the level of trust in the average model is dependent on $\theta$.
In this case, the duration of model training is unknown (number of iterations $T$), but one
can utilize the practical stopping criteria (see \cref{alg1_line:subproblem}). The only thing that we note is that we need to solve this problem with the learning rate $\eta_{\text{opt}}$. 
This is due to the fact that $L \geq \lambda$; therefore, we can assume that the "smoothness" constant of the problem \eqref{sliding:eq:prox} is equal to $L$ and we can take learning rate $\sim \frac{1}{L} = \eta_{\text{opt}}$.
\\
Under assumption that problem \eqref{ifnn:subproblem} is solved exactly we get $x_m^{k+1} = \hat x_m^{k+1}$ on \cref{alg1_line:update} and $u_{x_m}^{k+1} = x_m^{k+1}$ on \cref{alg1_line:acc}.
This is the whole essence of Algorithm \ref{alg:sliding_opt_comm}.
The only hyper-parameters that we can tune are $\theta$ and the number of iterations $T$. 
\\
\textbf{ \Cref{alg_sum}.} The  \Cref{alg_sum} with $r=1$ looks rather long,
let us for simplicity discuss the main idea in a centralized setting. We have
\begin{align*}
x^{k+1}_m = \begin{cases}
x^{k}_m - \frac{\eta }{1 - p}  \nabla_x f(x^{k}_m, y^{k}_m),&  \text{with prob.} ~~ 1 - p \\
x^{k}_m - \frac{\eta \lambda }{p} (x^{k}_m - \bar x^k) ,& \text{with prob.} ~~ p,
\end{cases}
\end{align*}
where
$p$ can be chosen according to importance sampling, in which case we have  $p = \frac{\lambda}{\lambda + L} \approx \frac{\lambda}{L}$. 
Further, let $\eta = \frac{\eta_{\text{opt}}}{q}$ with some $q \geq 1$. Then we get an update:
\begin{align*}
x^{k+1}_m = \begin{cases}
x^{k}_m - \frac{\eta_{\text{opt}}}{q}  \nabla_x f(x^{k}_m, y^{k}_m),&  \text{with prob.} ~~ 1 - p \\
x^{k}_m - \frac{1}{q} (x^{k}_m - \bar x^k) ,& \text{with prob.} ~~ p
\end{cases}
\end{align*}
Afterward, the algorithm makes a communication:
\begin{align}
\label{trust11}
   x^{k+1}_m = x^{k}_m - \tfrac{1}{q} (x^{k}_m - \bar x^k),
\end{align}
once in $\frac{1}{p} = \frac{1}{\lambda \eta_{\text{opt}}}$ iterations, otherwise make a local step: $x^{k+1}_m = x^{k}_m - \frac{\eta_{\text{opt}}}{q}  \nabla_x f(x^{k}_m, y^{k}_m)$.
\\
\textbf{Remark.} We wrote only about update of $x^{k}_m$, the update for $y$ is easy to get in a similar way.
\\
\textbf{Remark.} In order for the comparison of  \Cref{alg:sliding_opt_comm} and  \Cref{alg_sum} to be fair, it is necessary to balance the number of communications and local iterations for both algorithms, that is why we take $T = \frac{1}{p} = \frac{1}{\eta_{\text{opt} }\lambda} = \frac{10}{\lambda}$, where $T$ -- parameter of  \Cref{alg:sliding_opt_comm} and $p$ -- of  \Cref{alg_sum}. In the main paper take $\theta = q = 2$. This is because we aim to balance the degree of trust and reliance that each local model has on the average model (see \eqref{trust} and \eqref{trust11}).

\subsection{Neural networks with adversarial noise}
The goal of this experiment is to compare our new methods:  \Cref{alg:sliding_opt_comm} and  \Cref{alg_sum} for a neural network complemented with a robust loss:
%
$$
    f_m(x_m, y_m) := \frac{1}{N_m} \sum_{n=1}^{N_m} \ell(g(x_m, a_n + y_n), b_n)  + \frac{\beta_x}{2} \| x_m\|^2 - \frac{\beta_y}{2} \|y_m \|^2,
$$  
where $x_m$ are the weights of the $m$th model, $\{(a_n, b_n)\}_{n=1}^{N_m}$ are pairs of the training data on the $m$th node, $y_m$ is the so-called adversarial noise, which introduces a small effect of perturbation in the data, coefficients $\beta_x$ and $\beta_y$ are the regularization parameters. 
The inclusion of noise $y_m$ in the optimization process adds an interesting dimension to the standard loss function of a machine learning model. While the primary objective is still the minimization of the loss, the maximization of the noise term plays a crucial role in enhancing the model's robustness and stability.
This approach can be used for the simple regression problem or even for complex neural networks. By doing so, the trained model achieves good results against adversarial attacks (see e.g., \cite{nouiehed2019solving} and references therein). 
\begin{table*}[h]
\caption{Parameters for comparison of \Cref{alg:sliding_opt_comm} and \Cref{alg_sum}.}
\label{tab:parameters3}
\centering
\begin{threeparttable}
\begin{tabular}{cccc}
\toprule 
$\lambda$ & $T = 10/\lambda $ & $p = \lambda/10$ & Results\\ \hline
$1/4$& 40 (1 epoch)& $1/40$ &  Figure 1 (a)\\ \hline
$1/40$& 400 (10 epochs)& $1/400$ & Figure 1 (b)\\ \hline
$1/80$& 800 (20 epochs)& $1/800$  & Figure 1 (c)\\ \hline
\end{tabular}
\end{threeparttable} 
\end{table*}

\textbf{Data and model.} We consider the benchmark of image classification on the CIFAR-10 \cite{cifar10} dataset. It contains $50,000$ and $10,000$ images in the training and validation sets, respectively, equally distributed over $10$ classes. To emulate the distributed scenario, we partition the dataset into $N$ non-overlapping subsets in a heterogeneous manner. For each subset, we select a major class that forms $25\%$ of the data, while the rest of the data split is filled uniformly by other classes.
We parameterize each of the learners as a convolutional neural network, taking ResNet-18 \cite{he2015deep} architecture as the backbone. As a loss function, we use multi-class cross-entropy,  complemented with an adversarial noise.

\textbf{Setting.} To train ResNet18 in CIFAR-10, one can use stochastic gradient descent with momentum $0.9$, the learning rate of $0.1$ and a batch size of $128$ ($40$  batches = $1$ epoch). This is one of the default learning settings. Based on these settings, we build our settings using the intuition of algorithms (for details about tuning and intuition of our Algorithms, see Section \ref{intuition}). In order for the comparison of  \Cref{alg:sliding_opt_comm} and  \Cref{alg_sum} to be fair, it is necessary to balance two things: 1) the number of communications and local iterations for both algorithms, 2) the level how much each of the local models trusts the average model and relies on it. That is why we need carefully choose $T$ (the number of inner/local iterations in  \Cref{alg:sliding_opt_comm}) and $p$ (probability in  \Cref{alg_sum}). For more details how to choose $T$ and $p$ and how to tune level of reliance on the global model see Section \ref{intuition}.
Tables \ref{tab:parameters3} 
shows all the experimental setups that we consider.

\begin{figure}[h]
\begin{minipage}{0.33\textwidth}
  \centering
\includegraphics[width =  \textwidth ]{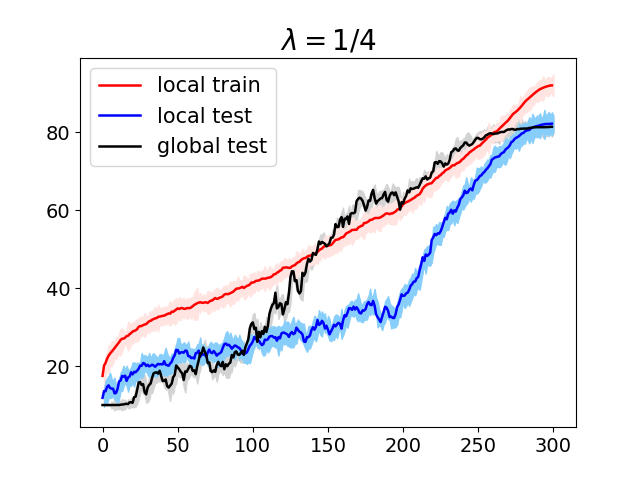}
\end{minipage}%
\begin{minipage}{0.33\textwidth}
  \centering
\includegraphics[width =  \textwidth ]{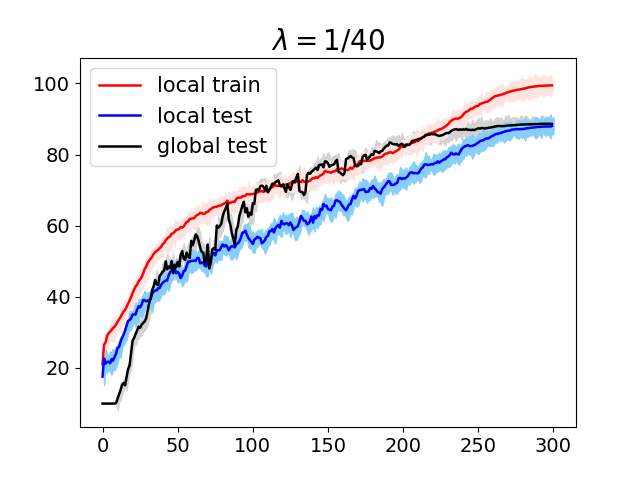}
\end{minipage}%
\begin{minipage}{0.33\textwidth}
  \centering
\includegraphics[width =  \textwidth ]{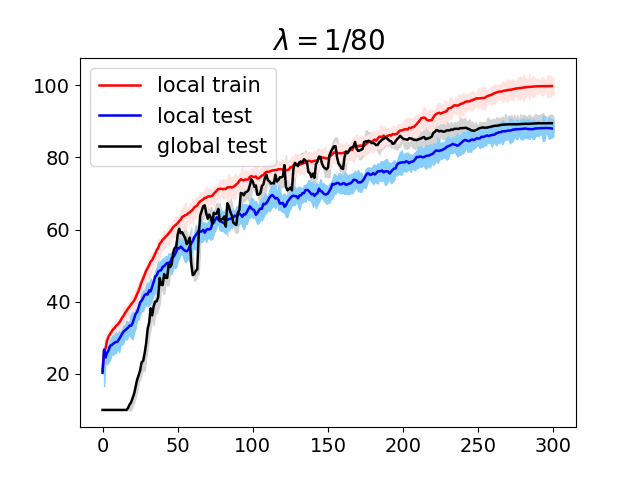}
\end{minipage}%
\\
\begin{minipage}{0.33\textwidth}
  \centering
\includegraphics[width =  \textwidth ]{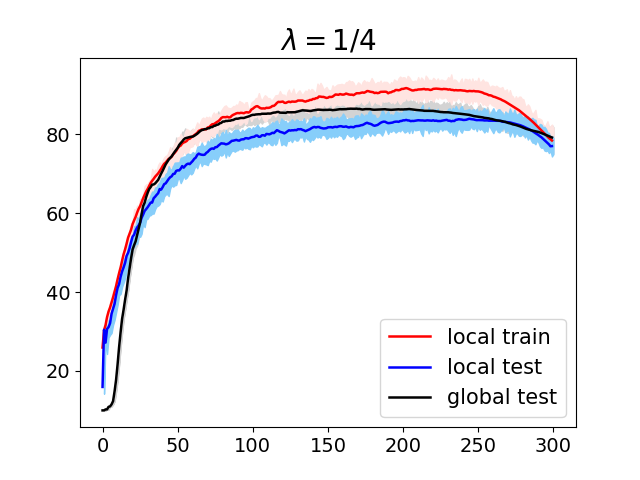}
\end{minipage}%
\begin{minipage}{0.33\textwidth}
  \centering
\includegraphics[width =  \textwidth ]{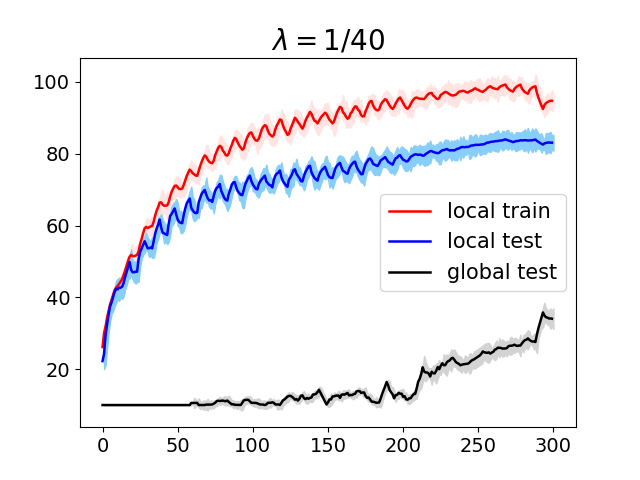}
\end{minipage}%
\begin{minipage}{0.33\textwidth}
  \centering
\includegraphics[width =  \textwidth ]{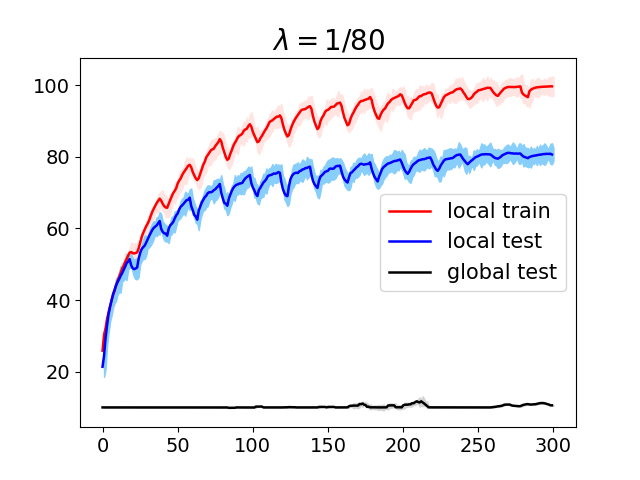}
\end{minipage}%
\\
\begin{minipage}{0.33\textwidth}
  \centering
\scriptsize{(a)}
\end{minipage}%
\begin{minipage}{0.33\textwidth}
  \centering
\scriptsize{(b)}
\end{minipage}%
\begin{minipage}{0.33\textwidth}
  \centering
\scriptsize{(c)}
\end{minipage}%
\captionof{figure}{Average accuracy in during process of learning with different average parameters $p$ and $T$. The first line presents the results of \Cref{alg:sliding_opt_comm}, the second -  \Cref{alg_sum}.
Red line -- accuracy of the local model on local train data,  blue line - accuracy of the local model on local test data, black line --  accuracy of the global model on global test data. The experiment was repeated 5 times, the deviations are reflected.}
\label{fig:1}
\end{figure}

\textbf{Results.} One can see the results of the experiment in Figure~\ref{fig:1}. Table~\ref{tab:parameters3} shows the correspondence of the various settings to the parameters.
\begin{minipage}{0.005\textwidth}
\end{minipage}

\textbf{Discussions.} We compare algorithms based on the balance of the local and global models, i.e. if the algorithm is able to train well both local and global models, then we find the FL balance by this algorithm. The results show that the Local SGD technique (\Cref{alg_sum}) outperformed the \Cref{alg:sliding_opt_comm} only with a fairly frequent device communication (Figure \ref{fig:1} (a)). In other cases (Figure \ref{fig:1} (b), (c)),  \Cref{alg_sum} was unable to train the global model, although it withstood the good quality of the local models. It turns out that the technique of  \Cref{alg:sliding_opt_comm} can be considered robust for Federated Learning, even in the case of neural networks.

\section{Conclusions and Future work}

In this paper, we present a novel formulation for the Personalized Federated Learning Saddle Point Problem \eqref{PF}. This formulation incorporates a penalty term that accounts for the specific structure of the network and is applicable to both centralized and decentralized network settings. Additionally, we provide the lower bounds both on the communication and the number of local oracle calls required to solve problem \eqref{PF}. Furthermore, we have developed the novel methods (\Cref{alg:sliding_opt_comm}, \Cref{alg:sliding_big_lambda}, \Cref{alg_sum}) for this problem that are optimal up to logarithmic factor in certain scenarios (see \Cref{tab:main}). These algorithms are based on sliding or variance reduction techniques. The theoretical analysis and experimental evidence corroborate our methods. Moreover, we have customized our approach for neural network training.

Possible interesting areas for further research are related to the practical features that arise in the federated learning setup, such as asynchronous transmissions and information compression to minimize communication costs, among other issues. It is worth considering the use of the variance reduction technique in accelerated sliding to develop an algorithm that is highly efficient in terms of communication and number of iterations. This is particularly relevant for cases in which the function at each node has a sum structure.

\subsection*{Acknowledgment}

This work was supported by a grant for research centers in the field of artificial intelligence, provided by the Analytical Center for the Government of the Russian Federation in accordance with the subsidy agreement (agreement identifier 000000D730324P540002) and the agreement with the Moscow Institute of Physics and Technology dated November 1, 2021 No. 70-2021-00138.

\bibliography{example_paper}
\newpage
\appendix
\section{Lower bounds} \label{app:lb}

\subsection{Notation}

In this section, we present a proof of lower bounds for the class of algorithms satisfying Assumption \ref{ass_lower_bounds}. Meanwhile, in order to simplify the obtaining of lower bounds and to make the problem a bit easier, we assume that communications are necessary only for the variable $y$. In more details, instead of \eqref{PF} we consider:
\begin{equation}
\label{main_problem_with_comm_by_y}
\min_{\x\in\mathbb{R}^{Md_x}}\max_{\y\in\mathbb{R}^{Md_y}}\left\{F(\x, \y) = \frac{1}{M}\sum^M_{m = 1} f_m(x_m, y_m) - \frac{\lambda}{2}\y^{\top}\mW\y\right\}, 
\end{equation}
where $\x^{\top} = (x^{\top}_1,\ldots,x^{\top}_M)$ and $\y^{\top} = (y^{\top}_1,\ldots,y^{\top}_M)$, $x_m \in \mathbb{R}^{d_x}$ and $y_m \in \mathbb{R}^{d_y}$ for any $m \in \{1,\ldots,M\}$, matrix $\mW = W\otimes I_{M}$, where $W$ is the gossip matrix (see  \Cref{def_gossip}). Moreover, $\mW$ satisfies the same properties of the gossip matrix:
\begin{enumerate}
    \item $\mW$ is symmetric and positive semi-defined matrix;
    \item $\mW_{i,j} \neq 0$ if and only if $i = j$ or $(i,j) \in \mathcal{E}$;
    \item $\text{ker}\mW$ is consensus space;
    \item $\lambda_{\max}(\mW) = \lambda_{\max}(W)$ and $\lambda^{+}_{\min}(\mW) = \lambda^{+}_{\min}(W)$, where $\lambda^{+}_{\min}$ is the smallest positive eigenvalue and $\lambda_{\max}$ is the largest eigenvalue;
\end{enumerate}
It is worth noting that for the simplicity of the proof, we work in vector space, unlike other sections where the proofs are carried out in matrix notation. We would also like to notation to the structure of the problem \eqref{main_problem_with_comm_by_y}. As mentioned, we consider a simplified formulation of the problem in comparison with the original one \eqref{PF}, but this only means that the lower bounds for the original problem \eqref{PF} will look either similar or more complicated 
than for a simple formulation \eqref{main_problem_with_comm_by_y}. 
Fortunately, the upper bounds for the problem \eqref{PF} coincide with the lower bounds for the problem \eqref{main_problem_with_comm_by_y}, which leads us to the following conclusion: the lower bounds are the same as for the problem \eqref{PF}.

\subsection{Proof of Theorem \ref{theorem_lower_bounds}}

As in many papers we give an example of a "bad" problem on which algorithms satisfying Assumption \ref{ass_lower_bounds} converge not better than the lower bounds provide. 

We start from the topology of the network. As the gossip matrix, we take the Laplacian of the linear chain. In more details, we choose ${W} = \tfrac1{2} U$, and
\begin{eqnarray*}
    U = 
    \begin{pmatrix}
        1 & -1 & & &   \\
        -1 & 2 & -1 &  &  \\
        & \ddots&\ddots & \ddots&  \\
        & &   -1 & 2 & -1\\
        & &    & -1 & 1\\
    \end{pmatrix}.
\end{eqnarray*}
One can check that this matrix satisfies Definition \ref{def_gossip} and $\chi$ is an upper bound for the condition number of the gossip matrix. We can also verify that $n - 1 > \sqrt{2 \chi} - 2 \geq \tfrac{1}{5} \sqrt{\chi}$ (since $\chi \geq 3$), $1 \leq \tfrac{2}{3} \lambda_{\max}(n)$ and $ \frac{4}{n^2}\leq \lambda^+_{\min}(n) \leq \tfrac{5}{n^2}$. For details see Appendix B of \cite{sadiev2022decentralized}. 

Next, we define the set of "bad" functions and their locations of the network. In particular, we take the dimension of the problem for both variables $d_x = d_y = d=2T$ with large enough $T$, where $T$ will be defined later. According to the location on the network we consider three types of the vertexes: the first type includes the 
leftmost vertex of the chain $\mathcal{V}_1=\left\{1\right\}$, the second type includes $\mathcal{V}_2=\left\{2, n - 1\right\}$,  the third type -- $\mathcal{V}_3=\left\{n\right\}$. Each type of node has its own functions: 
\begin{equation}
    \label{eq:bad}
    f_m(x, y) = 
    \begin{cases}
        \frac{\mu}{2}\|x\|^2 - \frac{\mu}{2}\|y\|^2 - a y(1) + x^{\top}A_1y, & \text{if } m \in \mathcal{V}_1 \\
        \frac{\mu}{2}\|x\|^2 - \frac{\mu}{2}\|y\|^2, & \text{if } m \in \mathcal{V}_2\\ 
        \frac{\mu}{2}\|x\|^2 -\frac{\mu}{2}\|y\|^2 + x^{\top}A_2y, & \text{if } m \in \mathcal{V}_3,
    \end{cases}
\end{equation}
with matrices $A_1$, $A_2$ are defined as follows
\begin{align*}
	A_1 = \begin{pmatrix}
	0 & 0 & 0& 0 & 0 &\dots &0  \\
	0 & 0 & -c & 0 & 0 &\ddots  &\vdots\\
	0 & 0 & c & 0 & 0 &\ddots  &\vdots \\
        0 & 0 & 0 & 0 & -c& \ddots  &\vdots \\
        0 & 0 & 0 & 0 & c &\ddots  &\vdots \\
	\vdots & \ddots & \ddots & \ddots & \ddots & \ddots& \vdots \\
	0 & \dots & \dots & \dots & \dots & \dots & b
	\end{pmatrix} \text{ and }
	A_2 =
	\begin{pmatrix}
	0 & -c & 0& 0 & 0 &\dots &0  \\
	0 & c & 0 & 0 & 0 &\ddots  &\vdots\\
	0 & 0 & 0 & -c & 0 &\ddots  &\vdots \\
        0 & 0 & 0 & c & 0 & \ddots  &\vdots \\
        0 & 0 & 0 & 0 & 0 &\ddots  &\vdots \\
	\vdots & \ddots & \ddots & \ddots & \ddots & \ddots& \vdots \\
	0 & \dots & \dots & \dots & \dots & \dots &c
	\end{pmatrix},
	\end{align*}
where $a, b, c$ will be defined later. Since $x_m$ of $\x$ are separate, it is to write the dual function $\Psi(\y)$ of our objective function $F(\x, \y)$:
\begin{equation}
    \label{eq:dual_problem}
    \Psi(\y) = \min_{\x}F(\x, \y) = \sum_{m=1}^M g_m(y_m)  - \frac{\lambda}{2}\y^{\top}\mW\y, 
\end{equation}
where 
\begin{equation*}
    g_m(y) = 
    \begin{cases}
        - \frac{\mu}{2}\|y\|^2 - a y(1) - \frac{1}{2 \mu} y^{\top} A_1^{\top} A_1y, & \text{if } m \in \mathcal{V}_1 \\
        - \frac{\mu}{2}\|y\|^2, & \text{if } m \in \mathcal{V}_2\\ 
        -\frac{\mu}{2}\|y\|^2 - \frac{1}{2 \mu} y^{\top} A_2^{\top} A_2 y, & \text{if } m \in \mathcal{V}_3,
    \end{cases}
\end{equation*}
To find the solution we can write down the optimality condition for  \eqref{eq:dual_problem}. For the first type of node, we denote the solution by $y^*$, for the third type node by $z^*$, and for the second type nodes by $u^*_2$, \ldots, $u^*_{n-1}$. First we write down for $y^*$:
\begin{equation}
    \label{x_y_3}
    \left( \frac{\mu}{\lambda} + \frac{1}{2} \right) y^*{(1)} + \frac{a}{\lambda} - \frac{1}{2} u^*_2{(1)} = 0,
\end{equation}
\begin{equation}
\label{x_y_2}
    \left(\frac{c^2}{\mu \lambda} + \frac{\mu}{\lambda} + \frac{1}{2}\right){y^*}(2t)- \frac{c^2}{\mu \lambda} {y^*(2t+1)} - \frac{1}{2}{u^*_2}(2t) = 0,\quad \text{for } 1\leq t \leq T-1,
\end{equation}
\begin{equation}
\label{x_y_1}
    \left(\frac{c^2}{\mu \lambda} + \frac{\mu}{\lambda} + \frac{1}{2}\right){y^* (2t+1)}- \frac{c^2}{\mu \lambda} {y^*(2t)} - \frac{1}{2}{u^*_2 (2t+1)} = 0,\quad \text{for } 1\leq t \leq T-1,
\end{equation}
\begin{equation}
    \label{x_y_4}
    \left( \frac{\mu}{\lambda} + \frac{b^2}{\mu \lambda} + \frac{1}{2} \right) y^*{(2T)} - \frac{1}{2} u^*_2{(2T)} = 0.
\end{equation}
Then for $z^*$:
\begin{equation}
\label{z_y_1}
    \left(\frac{c^2}{\mu \lambda} + \frac{\mu}{\lambda} + \frac{1}{2}\right){z^*}(2t-1)- \frac{c^2}{\mu \lambda} {z^*(2t)} - \frac{1}{2}{u^*_{n-1}(2t-1)} = 0,\quad \text{for } 1\leq t \leq T,
\end{equation}
\begin{equation}
\label{z_y_2}
    \left(\frac{c^2}{\mu \lambda} + \frac{\mu}{\lambda} + \frac{1}{2}\right){z^*(2t)}- \frac{c^2}{\mu \lambda} {z^*(2t-1)} - \frac{1}{2}{u^*_{n-1}(2t)} = 0,\quad \text{for } 1\leq t \leq T,
\end{equation}
Finally for $u^*_2$, \ldots, $u^*_{n-1}$:
\begin{equation}
    \label{yy_x}
    \left(1 + \frac{\mu}{\lambda}\right) u^*_2(t) - \frac{1}{2} u^*_3(t) - \frac{1}{2}y^*(t) = 0, \quad \text{for } 1\leq t \leq 2T,
\end{equation}
\begin{equation}
\label{yy}
    \left(1 + \frac{\mu}{\lambda}\right) u^*_{i}(t) - \frac{1}{2} u^*_{i+1}(t) - \frac{1}{2}u^*_{i-1}(t) = 0, \quad \text{for } 1\leq t \leq 2T,
\end{equation}
\begin{equation}
\label{yy_z}
    \left(1 + \frac{\mu}{\lambda}\right) u^*_{n-1}(t) - \frac{1}{2} u^*_{n-2}(t) - \frac{1}{2}z^*(t) = 0,\quad \text{for } 1\leq t \leq 2T.
\end{equation}
Note that these equations \eqref{x_y_3} -- \eqref{yy_z} correspond to equations (B.3) -- (B.11) from \cite{sadiev2022decentralized} with small changes: $ c \to \tfrac{c^2}{\mu \lambda}$ and $b \to \tfrac{b^2}{\mu \lambda}$. Hence, we can base our proof on the results from Appendix B of \cite{sadiev2022decentralized}. 

The next lemma gives an understanding of how the coordinates of the solution of \eqref{eq:dual_problem} are related to each other.

\begin{lemma}[Lemma 1 from \cite{sadiev2022decentralized}]\label{lem:1404}
The sequence $w_t$ satisfies the following recursion relation:
	\begin{equation*}
		w_{t+1}
		=
		Q
		w_{t} \quad \text{with} \quad
		Q = \begin{pmatrix}
	- \frac{B}{2 \tilde c} & ~~~~   &\frac{1}{\tilde c} \left(\tilde c+ \frac{\mu}{\lambda} +\frac{1}{2} - \frac{A}{2}\right)   \\
	-\frac{1}{\tilde c}\left(\tilde c+ \frac{\mu}{\lambda} +\frac{1}{2} - \frac{A}{2}\right) &~~~~&
	\frac{2}{B \tilde c}\left(\tilde c+ \frac{\mu}{\lambda} +\frac{1}{2} - \frac{A}{2}\right)^2 - \frac{2 \tilde c}{B}   \\
	\end{pmatrix},
	\end{equation*}
	where
	\begin{equation*}
		w_t =
		\begin{cases}
		\begin{pmatrix}
		{z^*(t)}\\
		{y^*(t)}
		\end{pmatrix} & \text{if } t \text{ is even} \\
		\begin{pmatrix}
		{y^*(t)}\\
		{z^*(t)}
		\end{pmatrix} & \text{if } t \text{ is odd}
		\end{cases}, \quad 
		A = \frac{\omega_2^{n-2} -  \omega_1^{n-2}}{\omega_2^{n-1} - \omega_1^{n-1}}, \quad B = \frac{\omega_2 -  \omega_1}{\omega_2^{n-1} - \omega_1^{n-1}}, \quad  \tilde c = \frac{c^2}{\mu \lambda}
	\end{equation*}
	with $\omega_1 = 1 + \tfrac{\mu}{\lambda} - \sqrt{\tfrac{2\mu}{\lambda} + \tfrac{\mu^2}{\lambda^2}}$ and $\omega_2 = 1 + \tfrac{\mu}{\lambda} + \sqrt{\tfrac{2\mu}{\lambda} + \tfrac{\mu^2}{\lambda^2}}$.
\end{lemma}
\begin{proof}
The proof is the same with the proof of Lemma 1 from \cite{sadiev2022decentralized}. 
\end{proof}

By choosing the parameters $a$ and $b$, we can get that $w_1$, $w_2$, \ldots, $w_{2T}$ are eigenvectors of the matrix $Q$, i.e. $w_{2T} = Q w_{2T-1} = \gamma w_{2T-1} = \gamma Q w_{2T-2} = \gamma^2 w_{2T-2}$ etc. This idea is implemented in the following lemma.

\begin{lemma}[Lemma 2 from \cite{sadiev2022decentralized}]\label{lem:2404}
For any $L$, $\mu$, and $\lambda$ ($L \geq 2\mu$, and $\lambda\lambda^+_{\min} \geq \mu$), there exists a choice of parameters $a$, $b$, $c$ such that $w_1$, $w_2$, \ldots $w_{2T}$ are eigenvectors of matrix $Q$ corresponding to the eigenvalue $\gamma \in (0;1)$, where
\begin{equation}
    \label{eq:gamma}
    \gamma \geq 1 - \max \left\{ 2\sqrt{\frac{\mu n^2 }{\lambda}}, 3\frac{\mu}{L - \mu}\right\}.
\end{equation}
Moreover, the problem \eqref{main_problem_with_comm_by_y} $+$ \eqref{eq:bad} with these parameters $a$, $b$, $c$ satisfies Assumptions \ref{ass:smooth} and \ref{ass:sc}. 
\end{lemma}
\begin{proof}
Following the proof of Lemma 2 from \cite{sadiev2022decentralized}, we give the values of $a$, $b$, $\tilde b$, $c$ and $\tilde c$:
\begin{equation}
\label{eq:c}
c = 
\begin{cases}
\sqrt{\lambda \mu}, \quad \text{for} \quad \mu + \sqrt{\lambda \mu} \leq L,\\
L - \mu , \quad \text{for} \quad \mu + \sqrt{\lambda \mu} > L,
\end{cases}
\quad \tilde c = \frac{c^2}{\mu \lambda}
\end{equation}
\begin{equation*}
\tilde b = \frac{B  \alpha }{2} - \frac{\mu}{\lambda} - \frac{1}{2} + \frac{A}{2}, \quad b = \sqrt{\mu \lambda \tilde b} \quad \text{and} \quad \text{any } a \neq 0,
\end{equation*}
where
\begin{equation*}
\resizebox{\linewidth}{!}{
$
\alpha = -\frac{1 - 2A + A^2 + B^2 + 4 \tilde c - 4A \tilde c + 4\frac{\mu}{\lambda} - 4A\frac{\mu}{\lambda} + 8\tilde c\frac{\mu}{\lambda} + 4\frac{\mu^2}{\lambda^2} + \sqrt{\left(-1 + 2A - A^2 + B^2 - 4\frac{\mu}{\lambda} + 4A\frac{\mu}{\lambda} - 4\frac{\mu^2}{\lambda^2}\right)\left(-1 + 2A - A^2 + B^2 - 8 \tilde c + 8Ac - 16c^2 - 4\frac{\mu}{\lambda} + 4A\frac{\mu}{\lambda} - 16c\frac{\mu}{\lambda} - 4\frac{\mu^2}{\lambda^2}\right)}}{2B\left(-1 + A - 2 \tilde c - 2\frac{\mu}{\lambda}\right)},
$
}
\end{equation*}
We verify that the problem \eqref{main_problem_with_comm_by_y} + \eqref{eq:bad} with these choice of $c$ and $b$ satisfies  Assumptions \ref{ass:smooth} and \ref{ass:sc}. Assumption \ref{ass:sc} is easy to check. To get  Assumption \ref{ass:smooth} we need that $c,b \leq L-\mu$. $c \leq L-\mu$ by definition \eqref{eq:c}. One can note that $\tilde c \in [0;1]$. In the proof of Lemma 2 from \cite{sadiev2022decentralized} the authors show that for $\tilde c \in [0;1]$ it holds that $\tilde b \in [0;\tilde c]$. Then, we get $b = \sqrt{\mu \lambda \tilde b} \in [0;c]$. It gives the verification of Assumption \ref{ass:smooth}.

Following the proof of Lemma 2 from \cite{sadiev2022decentralized}, we find the the smallest eigenvalue of $Q$
\begin{equation*}
\resizebox{\linewidth}{!}{
$
\gamma = \frac{4 - 8A + 4A^2 - 4B^2 + 16 \tilde c - 16A \tilde c + 16 \frac{\mu}{\lambda} - 16 A\frac{\mu}{\lambda} + 32 \tilde c\frac{\mu}{\lambda} + 16 \frac{\mu^2}{\lambda^2} - \sqrt{-256 B^2 \tilde c^2 + \left(-4 + 8A - 4A^2 + 4 B^2 - 16 \tilde c + 16Ac - 16\frac{\mu}{\lambda} + 16 A\frac{\mu}{\lambda} - 32c\frac{\mu}{\lambda} - 16\frac{\mu^2}{\lambda^2}\right)^2}}{16 B \tilde c}
$
}
\end{equation*}
and the corresponding eigenvector
\begin{equation*}
v = 
\begin{pmatrix}
\alpha \\
1
\end{pmatrix},
\end{equation*}
Using the same reasons as in \cite{sadiev2022decentralized}, with $\tilde b = \frac{B  \alpha }{2} - \frac{\mu}{\lambda} - \frac{1}{2} + \frac{A}{2}$, $w_{2T}$ is an eigenvector of $Q$. It means that $\gamma w_{2T} = Q w_{2T}$ or $w_{2T} =  \gamma Q^{-1} w_{2T}$. Lemma \ref{lem:1404} gives that $Q^{-1} w_{2T} = w_{2T-1}$. As the result, $w_{2T} = \gamma w_{2T-1}$, i.e. $w_{2T-1}$ is also an eigenvector of $Q$. Continuing further, we obtain that all vectors $w_{2T}$, \ldots, $w_1$ are eigenvectors of $Q$. The choice of parameter $a$ does not affect, it only determines the value of $\|w_1\|$.

It remains to show \eqref{eq:gamma}. We consider the three cases separately:

1) $\mu + \sqrt{\mu\lambda} \leq L$. In this case $\tilde c = 1$. We want to verify that $\gamma \in (0;1)$ and $\gamma \geq 1 - 2 \sqrt{\tfrac{\mu n^2}{\lambda}}$. This inequality need to be checked with the constraints: $x = \tfrac{\mu}{\lambda} > 0$, $x = \tfrac{\mu}{\lambda} \leq  \lambda^+_{\min} \leq \tfrac{5}{n^2}$ (since in the conditions of Lemma we assume that $\tfrac{\mu}{\lambda} \leq  \lambda^+_{\min}$ and above we estimated that $\lambda^+_{\min} \leq \tfrac{5}{n^2}$, when we construct the network).

2) $\mu + \sqrt{\mu \lambda} > L \geq \mu + \sqrt{\mu  \lambda \lambda_{\min}^+}$. In this case $\tilde c = \tfrac{(L - \mu)^2}{\lambda \mu} = \tfrac{\mu}{\lambda} \cdot \delta = x \delta$. We want to verify that $\gamma \in (0;1)$ and $\gamma \geq 1 - 2 \sqrt{\tfrac{\mu n^2}{\lambda}}$. This inequality need to be checked with the constraints: $\delta \geq 1$ (since in Theorem \ref{theorem_lower_bounds} we assume that $L \geq 2\mu$ and then $\delta = \left(\tfrac{L - \mu}{\mu}\right)^2 \geq 1$), $x = \tfrac{\mu}{\lambda} > 0$, $x = \tfrac{\mu}{\lambda} \leq  \lambda^+_{\min} \leq \tfrac{5}{n^2}$ and $\tfrac{1}{x} > \delta \geq  \tfrac{1}{x} \cdot \lambda^+_{\min} \geq \tfrac{1}{x} \cdot \tfrac{4}{n^2}$.

3) $\mu + \sqrt{\mu \lambda \lambda_{\min}^+}  > L$. In this case $\tilde c = \tfrac{(L - \mu)^2}{\lambda \mu} = \tfrac{\mu}{\lambda} \cdot \delta = x \delta$. We want to verify that $\gamma \in (0;1)$ and $\gamma \geq 1 - 3 \tfrac{\mu}{L - \mu} = 1 - 3 \sqrt{\tfrac{1}{\delta}}$. This inequality need to be checked with the constraints: $\delta \geq 1$, $x = \tfrac{1}{x} > 0$, $x = \tfrac{1}{x}\tfrac{1}{x} \leq  \lambda^+_{\min} \leq \tfrac{5}{n^2}$ and $\delta < \tfrac{1}{x} \cdot \lambda^+_{\min} \leq \tfrac{1}{x} \cdot \tfrac{5}{n^2}$.

These three cases are considered and proved in \cite{sadiev2022decentralized}. This fact finishes the proof of our lemma.
\end{proof}

The previous Lemmas show what the exact view of the solution. Now let us determine how quickly we can find it with any algorithm.

\begin{lemma} \label{lem_3404}
Let the problem \eqref{main_problem_with_comm_by_y} + \eqref{eq:bad} be solved by any method that satisfies Assumption \ref{ass_lower_bounds}. Then after $K$ iterations with $q$ communication rounds,  only the first $\left\lfloor \tfrac{q}{n-1} \right\rfloor$ coordinates of the   global output can be non-zero while  the rest of the $d-\left\lfloor \tfrac{q}{n-1} \right\rfloor$ coordinates are strictly equal to zero.
\end{lemma}

\begin{proof}
We starting with introducing some additional notation. Let
\begin{align*}
    E_{0} := \{ 0\}, \quad E_{k} := \text{span} \{ e_1, \ldots, e_k\},
\end{align*}
where $\{e_k\}$ is vectors of the orthogonal Euclidean basis. Note that, the zero initialization gives $\mathcal{M}_{m, 0} = (E_0, E_0)$.

Suppose that, for some $m$,  $\mathcal{M}_{m, t} = \{(E_k, E_k)\}$ at some given time $t$. We need to analyze how $\mathcal{M}_{m}$ can change by performing only local computations. 

We consider the case when 
$k$ odd. After one local update, we have the following: 

$\bullet$ For the device on $\mathcal{V}_1$, it holds
\begin{equation*}\begin{aligned}
        x \in \text{span} \big\{&e_1~,~ x'~,~A_{1} y'\big\} = E_k,\\
        y \in \text{span} \big\{&e_1~,~ y'~,~A_1^T x'\big\} = E_k, 
\end{aligned}\end{equation*}
for given $(x', y') \in \mathcal{M}_{m, t}$. 
Since $A_1$ has a block diagonal structure,   after any local computations, we have  $\mathcal{M}_{m} = E_k$ and $\mathcal{M}_{m} = E_k$. The situation does not change, no matter how many local computations one  does.

$\bullet$ For the device on $\mathcal{V}_3$, it holds
\begin{equation*}\begin{aligned}
        x \in \text{span} \big\{&x'~,~A_2 y'\big\} = E_{k+1},\\
        y  \in \text{span} \big\{&y'~,~A_2^T x'\big\} = E_{k+1}, 
\end{aligned}\end{equation*}
for given $(x', y') \in \mathcal{M}_{m, t}$. It means that, after some local computations, one can  have $\mathcal{M}_{m} = (E_{k+1}, E_{k+1})$. Therefore,  the machine form $\mathcal{V}_3$ can progress by one new non-zero coordinate, but no more. The case with even $k$ can be analyzed analogically.

This means that we constantly have to transfer progress from $\mathcal{V}_1$ to $\mathcal{V}_3$ and vice versa. Initially, all devices have all zero coordinates. Further, machines on $\mathcal{V}_1$ can get the first non-zero coordinate (but only the first, the second is not), and the rest of the devices are left with all zeros. Next, we send the progress with the first non-zero coordinate to $\mathcal{V}_3$. To do this, $(n-1)$ communication rounds are needed. By doing the same way,  we can make the second coordinate non-zero, and then transfer this progress to $\mathcal{V}_1$. Then the process continues in the same way. This  completes the proof.
\end{proof}

The results of Lemmas \ref{lem:1404} -- \ref{lem_3404} are the same with Lemmas 1--3 from \cite{sadiev2022decentralized}. The only difference in Lemma \ref{lem:2404} with $\gamma$: $\sqrt{\frac{\mu}{L-\mu}} \to \frac{\mu}{L-\mu}$. Hence, following \cite{sadiev2022decentralized}, we just need to put $T = \tfrac{1}{2}\left( \max \{ 1, \log_{\gamma} \tfrac{1}{2} \} + \left\lfloor \tfrac{q}{n-1} \right\rfloor \right)$ in the dimension of the problem $d = 2T$ and then one can obtain that any algorithm solving \eqref{eq:dual_problem} needs at least
\begin{align*}
    q = \Omega\left( \min\left\{\sqrt{\frac{\lambda \lambda_{\max}(W)}{\mu }}, \frac{(L - \mu)\chi}{\mu} \right\}  \log\frac{1}{\varepsilon}\right) \quad \text{communication rounds}
\end{align*}
to find $\varepsilon$-solution.

\newpage
\section{Proofs of convergence theorems}\label{proofs}
Let us introduce some notation which we use in this section. We use matrix notation in our proofs, as in the main body of the paper. It is easy to check that the matrix notation in this case is not differ greatly from the standard vector notation; it is enough just to move from the Euclidean norm to the Frobenius norm and from the scalar product to the trace for $X = [x_1, \ldots, x_M]^T$ and $Y = [y_1, \ldots, y_M]^T$:
\begin{align*}
    \|X \|^2 = \| x_1\|^2 + \ldots + \| x_M\|^2, \quad \langle X,  Y \rangle = \text{Tr} [X^TY] = \langle x_1, y_1 \rangle + \ldots + \langle x_M, y_M \rangle,
\end{align*}
where $x_1, \ldots, x_M$ and $y_1, \ldots, y_M$ -- vectors from which matrices $X$ and $Y$ are composed according to Section \ref{sec:formul}. Additionally, we use matrix of the solution $X^* = [x^*, \ldots, x^*]^T$ and $Y^* = [y^*, \ldots, y^*]^T$.
Now we introduce the Bregman divergence for convex function $F(z)$ as
\begin{align}\label{def:BD_Psi}
    \mathrm{D}_{F}(z_2, z_1) = F(z_2) - F(z_1) - \langle \nabla F (z_1), z_2 - z_1\rangle.
\end{align}
Using convexity of $F(x)$ we get the following property of Bregman divergence for convex function $\mathrm{D}_{F}(z_2, z_1) \geq 0$.

We also use the next representation for brevity and simplicity. Composite saddle point problems (in particular, \eqref{PF}) can be considered in the form of variational inequalities \cite{nemirovski2004prox, juditsky2011solving}:
\begin{align}\label{pr:PF2}
     \text{Find} ~~ z^* \in \R^{M\times (d_x + d_y)} ~~ \text{such that} ~~  \langle B(z^*), z - z^* \rangle + \Psi(z) - \Psi(z^*)  \geq 0,~~ \forall z \in \R^{M\times (d_x + d_y)}, 
\end{align}
where $z:= [X, Y]$, $X = [x_1,\ldots,x_M]^T \in \R^{M\times d_x}$, $Y = [y_1,\ldots,y_M]^T \in \R^{M\times d_y}$ and $B: \R^{M\times (d_x + d_y)} \to \R^{M\times (d_x + d_y)}$. 

Before proceeding to the proofs, it remains to introduce the following assumptions.
\begin{assumption}\label{ass:var_ineq:lipschitz}
The operator $B: \R^{M\times (d_x + d_y)}$ is $L_B$-Lipschitz continuous, i.e. for all $z_1, z_2 \in \R^{M\times (d_x + d_y)}$ we have
$
  \|B(z_1) - B(z_2)\| \leq L_B \|z_1-z_2\|.
$

For case $B(z) = \frac{1}{r}\sum_{i=1}^rB_i(z)$ we assume that operator $B_i(z)$ is $L_{B}$-Lipschitz continuous.  
\end{assumption}

\begin{assumption}\label{ass:var_ineq:str_monotone}
The operator $B: \R^{M\times (d_x + d_y)} \to \R^{M\times (d_x + d_y)}$ is $\mu_B$-strongly monotone, i.e. for all $z_1, z_2 \in \R^{M\times (d_x + d_y)}$ we have
$
     \left\langle B(z_1) - B(z_2), z_1 - z_2 \right\rangle \geq \mu_B\|z_1-z_2\|^2.
$
\end{assumption}

\begin{assumption}\label{ass:var_ineq:smooth}
The function $\Psi(z): \R^{M\times (d_x + d_y)} \to \R$ is convex and $L_{\Psi}$-smooth.
\end{assumption}

For the problem \eqref{PF}, $B(z) = [\nabla_X F(X, Y), -\nabla_Y F(X,Y)]$, where $F(X, Y) = \sum_{m=1}^M f_m(x_m, y_m)$, $\nabla_X F(X, Y) = [\nabla_x f_1(x_1, y_1), \ldots, \nabla_x f_m(x_m, y_m)]^T$, $\nabla_Y F(X, Y) = [\nabla_y f_1(x_1, y_1), \ldots, \nabla_y f_m(x_m, y_m)]^T$ and $\Psi(z) = \frac{\lambda}{2}\langle z, W z \rangle = \frac{\lambda}{2}\langle X, WX \rangle + \frac{\lambda}{2}\langle Y, WY \rangle$. Using \Cref{ass:smooth} and \Cref{ass:sc} it is easy to notice that $L_B = L$ and $\mu_B = \mu$. Also, $L_{\Psi} = \lambda \lambda_{\max}(W)$, where $\lambda_{\max}(W)$ is the maximum eigenvalue of $W$ and $\Psi(z)$ is convex function due to $W$ is positive semi-definite by \Cref{def_gossip}. 

For the case when $f_m(x_m, y_m) = \frac{1}{r}\sum_{i=1}^r f_{m, i}(x_m, y_m)$ we say that $B(z) = \frac{1}{r}\sum_{i=1}^r B_i(z)$, where $B_i(z) = [\nabla_X F_i(X, Y), -\nabla_Y F_i(X,Y)]$, $F_i(X, Y) = \sum_{m=1}^M f_{m, i}(x_m, y_m)$, $\nabla_X F_i(X, Y) = [\nabla_x f_{1, i}(x_1, y_1), \ldots, \nabla_x f_{m, i}(x_m, y_m)]^T$, $\nabla_Y F_i(X, Y) = [\nabla_y f_{1, i}(x_1, y_1), \ldots, \nabla_y f_{m, i}(x_m, y_m)]^T$. It is easy to notice that $L_{B} = L$

\subsection{Proof of \Cref{th:sliding_opt_comm}}
\Cref{alg:sliding_opt_comm_v2} is adapted version of \Cref{alg:sliding_opt_comm} in introduced notation.
\begin{algorithm}[h]
	\caption{\Cref{alg:sliding_opt_comm} in new notation}
	\label{alg:sliding_opt_comm_v2}
	\hspace*{\algorithmicindent} {\bf Parameters:} stepsize $\alpha$,  $\eta$\\
	\hspace*{\algorithmicindent} {\bf Initialization:} choose  $ z^0$, $v^0 = u^0 = z^0$  for all nodes $m \in (1, M)$
	\begin{algorithmic}[1]
		\For{$k=0,1,2,\ldots$}
		\State $v^k = \alpha z^k + (1 - \alpha) u^k $ \label{ae:line:2}
		\State 
		Find an approximate solution $\hat z^{k+1}$ of $ G^k(z) = 0$, where
		\begin{equation}\label{sliding:eq:prox}
			 G^k(z) =  \nabla \Psi(v^{k}) + \frac{z - z^k}{\eta} + B(z)
		\end{equation} 
    \State$z^{k+1} = z^k - \eta(\nabla \Psi( v^{k}) + B(\hat z^{k+1}))$ \label{ae:line:4}
		\State $u^{k+1} =  v^k + \alpha(\hat z^{k+1} - z^k)$ 
		\EndFor
	\end{algorithmic}
\end{algorithm}

\begin{lemma}\label{lemma:scl_prod}
	 Under Assumptions \ref{ass:var_ineq:lipschitz}-\ref{ass:var_ineq:smooth}, the following inequality holds for \Cref{alg:sliding_opt_comm_v2}.
	\begin{align*}
 -2\left \langle 
            \nabla \Psi (v^{k}) + B (\hat z^{k+1}),
            \hat z^{k+1} - z^*\right \rangle \leq&-\frac{2}{\alpha}\left \langle 
            \nabla \Psi(v^{k})  - \nabla \Psi( z^*),
            u^{k+1} - v^k\right \rangle 
            \\&+ \frac{2(1 - \alpha)}{\alpha}(\mathrm{D}_{\Psi}(u^k, z^*) - \mathrm{D}_{\Psi}(v^k, z^*)) 
            \\&-2 \mathrm{D}_{\Psi}(v^{k}; z^*) - \mu_B\|\hat z^{k+1} - z^*\|^2.
\end{align*}
\end{lemma}
\begin{proof}
Using the first-order necessary condition for the problem \eqref{PF} $ 
              \nabla \Psi(z^*) + B(z^*)  =  0 $ and $\mu_B$-strong monotoncity $B(z)$,  we get
\begin{align*}
    -2\left \langle 
            \nabla \Psi( v^{k}) \right.&+\left. B (\hat z^{k+1}),
            \hat z^{k+1} - z^*\right \rangle 
            \\ = & -2\left \langle 
            \nabla \Psi(v^{k}) - \nabla \Psi(z^*),
            \hat z^{k+1} - z^*\right \rangle  -2\left \langle 
             B (\hat z^{k+1}) - B (z^*),
            \hat z^{k+1} - z^*\right \rangle
        \\\leq & -2\left \langle 
            \nabla \Psi(v^{k}) - \nabla \Psi(z^*),
            \hat z^{k+1} - z^*\right \rangle  - \mu_B\|\hat z^{k+1} - z^*\|^2
            \\= & -2\left \langle 
            \nabla \Psi(v^{k}) - \nabla \Psi(z^*),
            \hat z^{k+1} - z^k\right \rangle -2\left \langle 
            \nabla \Psi(v^{k}) - \nabla \Psi(z^*),
             z^{k} - v^k\right \rangle
            \\&-2\left \langle 
            \nabla \Psi(v^{k}) - \nabla \Psi(z^*),
             v^{k} - z^*\right \rangle - \mu_B\|\hat z^{k+1} - z^*\|^2.
\end{align*}
Using convexity of $\Psi (z)$ and the definition of the Bregman divergence \eqref{def:BD_Psi} for the function $\Psi (z)$, we get
\begin{align*}
 -2\left \langle 
            \nabla \Psi( v^{k}) \right.&+\left. B (\hat z^{k+1}),
            \hat z^{k+1} - z^*\right \rangle 
            \\\leq & -2\left \langle 
            \nabla \Psi(v^{k}) - \nabla \Psi(z^*),
            \hat z^{k+1} - z^k\right \rangle -2\left \langle 
            \nabla \Psi(v^{k}) - \nabla \Psi(z^*),
             z^{k} - v^k\right \rangle 
            \\&-2\left ( \Psi(v^{k}) -  \Psi(z^*)\right) + 2\left \langle 
            \nabla \Psi(z^*),
             v^{k} - z^*\right \rangle - \mu_B\|\hat z^{k+1} - z^*\|^2
             \\= &-2\left \langle 
            \nabla \Psi(v^{k}) - \nabla \Psi(z^*),
            \hat z^{k+1} - z^k\right \rangle -2\left \langle 
            \nabla \Psi(v^{k}) - \nabla \Psi(z^*),
             z^{k} - v^k\right \rangle 
            \\&- 2 \mathrm{D}_{\Psi}(v^{k};z^*) - \mu_B\|\hat z^{k+1} - z^*\|^2.
\end{align*}
Now, we use \cref{ae:line:2} of \Cref{alg:sliding_opt_comm_v2} and get

\begin{align*}
-2\left \langle 
            \nabla \Psi(v^{k}) + B (\hat z^{k+1}),
            \hat z^{k+1} - z^*\right \rangle \leq&-\frac{2}{\alpha}\left \langle 
            \nabla \Psi(v^{k}) -  \nabla \Psi(z^*),
            u^{k+1} - v^k\right \rangle
            \\&+ \frac{2(1 - \alpha)}{\alpha}\left \langle 
            \nabla \Psi(v^{k}
            ) - \nabla \Psi(z^*),
            u^{k} - v^k\right \rangle
            \\&-2 \mathrm{D}_{\Psi}(v^{k}; z^*) - \mu_B\|\hat z^{k+1} - z^*\|^2.
\end{align*}
Using definition and property of the Bregman divergence for the convex function $\Psi (z):$  \eqref{def:BD_Psi} and $\mathrm{D}_{\Psi}(z_2, z_1) \geq 0$, we get
\begin{align*}
  -2\left \langle 
            \nabla \Psi(v^{k}) + B (\hat z^{k+1}),
            \hat z^{k+1} - z^*\right \rangle \leq & -\frac{2}{\alpha}\left \langle 
            \nabla \Psi(v^{k}) - \nabla \Psi(z^*),
            u^{k+1} - v^k\right \rangle 
            \\& + \frac{2(1 - \alpha)}{\alpha}(\mathrm{D}_{\Psi}(u^k, z^*) - \mathrm{D}_{\Psi}(v^k, z^*)) 
            \\&-2 \mathrm{D}_{\Psi}(v^{k}; z^*)
            - \mu_B\|\hat z^{k+1} - z^*\|^2 .
\end{align*}
This completes the proof of Lemma.
\end{proof}

\begin{lemma}\label{lemma:lyapunov_func}
	Consider  \Cref{alg:sliding_opt_comm_v2} for the problem \eqref{pr:PF2} under Assumptions \ref{ass:var_ineq:lipschitz}-\ref{ass:var_ineq:smooth}, with the following tuning: 
\begin{equation}\label{ae:choice}
    \alpha = \min\left\{1,\sqrt{\frac{\mu_B}{L_{\Psi}}}\right\}, \quad  \eta = \min \left\{\frac{1}{3\mu_B},\frac{1}{3 L_{\Psi} \alpha}\right\},
	\end{equation}
and let $\hat z^{k+1}$   in \cref{ae:line:2} satisfy    
	\begin{equation}\label{aux:grad_app}
	\| G^k (\hat z^{k+1})\|^2  \leq \frac{1}{6\eta^2}\|\hat z^{k+1} - z^k\|^2.
	\end{equation}
Then, the following inequality holds:
	\begin{equation}\label{ae:rec}
		\mathrm{\Phi}^{k+1} \leq \left(1 - \frac{\alpha}{3}\right)\mathrm{\Phi}^k,
	\end{equation}
	where  
	\begin{equation}\label{ae:Phi}
		\mathrm{\Phi}^k = \frac{1}{\eta}\|z^k - z^*\|^2 + \frac{2}{\alpha}\mathrm{D}_{\Psi}(u^k, z^*).
	\end{equation}
\end{lemma}
\begin{proof}
Using \cref{ae:line:4} of \Cref{alg:sliding_opt_comm_v2}, we get
	\begin{align*}
		\frac{1}{\eta}\left\|
            z^{k+1} - z^*\right\|^2 
		=&
		\frac{1}{\eta}\left\|
            z^{k} - z^*\right\|^2
		+ \frac{2}{\eta} \left \langle
            z^{k+1} - z^k, 
            z^{k} - z^*\right \rangle
		+\frac{1}{\eta}\left\|
            z^{k+1} - z^k\right\|^2
		\\=&
		\frac{1}{\eta}\left\|
            z^{k} - z^*\right\|^2
		-2\left \langle 
            \nabla \Psi(v^{k}) + B(\hat z^{k+1}),
            z^{k} - z^*\right \rangle
		+\frac{1}{\eta}\left\|
            z^{k+1} - z^k\right\|^2
            \\=&
		\frac{1}{\eta}\left\|
            z^{k} - z^*\right\|^2
		+2\eta\left \langle 
            \nabla \Psi (v^{k}) + B(\hat z^{k+1}), 
              \frac{\hat z^{k+1} - z^{k}}{\eta}
            \right \rangle
        \\&-2\left \langle 
            \nabla \Psi( v^k) + B(\hat z^{k+1}),
             \hat z^{k+1} - z^*\right \rangle
		+\frac{1}{\eta}\left\|
            z^{k+1} - z^k\right\|^2.
    \end{align*}
Since $2\langle a, b\rangle = \|a+b\|^2 - \|a\|^2 - \|b\|^2$, we get
    \begin{align*}
        \frac{1}{\eta}\left\|
            z^{k+1} - z^*
            \right\|^2 
		=&
		\frac{1}{\eta}\left\|
            z^{k} - z^*
            \right\|^2 + \eta\left \| 
            \nabla \Psi( v^{k}) + B(\hat z^{k+1}) + \frac{\hat z^{k+1} - z^{k}}{\eta} \right\|^2
		\\&-\eta\left \| \nabla \Psi (v^{k}) +B(\hat z^{k+1})\right\|^2 -\frac{1}{\eta}\left\|
              \hat z^{k+1} - z^{k}\right \|^2
        \\&-2\left \langle 
           \nabla \Psi(v^{k}) +B(\hat z^{k+1}), 
             \hat z^{k+1} - z^*\right \rangle
		+\frac{1}{\eta}\left\|
            z^{k+1} - z^k\right\|^2.
    \end{align*}
Using \cref{ae:line:4} of \Cref{alg:sliding_opt_comm_v2}, we get
    \begin{align*}
        \frac{1}{\eta}\left\|
            z^{k+1} - z^*
            \right\|^2 
		=&
		\frac{1}{\eta}\left\|
            z^{k} - z^*\right\|^2 + + \eta\left \| 
            \nabla \Psi(v^{k}) + B(\hat z^{k+1}) + \frac{\hat z^{k+1} - z^{k}}{\eta} \right\|^2
		\\&-\frac{1}{\eta}\left \| 
            z^{k+1} - z^k \right\|^2 -\frac{1}{\eta}\left\|
              \hat z^{k+1} - z^{k}\right \|^2
        \\&-2\left \langle 
            \nabla \Psi(v^{k}) +B(\hat z^{k+1}), 
             \hat z^{k+1} - z^*\right \rangle
		+\frac{1}{\eta}\left\|
            z^{k+1} - z^k\right\|^2
            \\ = &
		\frac{1}{\eta}\left\|
            z^{k} - z^*\right\|^2 + 2 \eta\left \| 
             G^k (\hat z^{k+1})  
            \right\|^2
		 -\frac{1}{\eta}\left\|
              \hat z^{k+1} - z^{k}\right \|^2
        \\&-2\left \langle 
            \nabla \Psi(v^{k}) +B(\hat z^{k+1}), 
             \hat z^{k+1} - z^*\right \rangle.
    \end{align*} 
Using \Cref{lemma:scl_prod} and \cref{ae:line:4} of \Cref{alg:sliding_opt_comm_v2} , we get
\begin{align*}
    \frac{1}{\eta}\left\|
            z^{k+1} - z^*\right\|^2 
		\leq&\frac{1}{\eta}\left\|
            z^{k} - z^*\right\|^2 + 2 \eta\left \| 
             G^k (\hat z^{k+1})  
            \right\|^2-\frac{2}{3\eta}\left\|
              \hat z^{k+1} - z^{k}\right \|^2
            \\&-\frac{2}{\alpha}\left( \left\langle 
            \nabla \Psi(v^{k}) - \nabla \Psi(z^*) ; 
             u^{k+1} - v^k \right\rangle + \frac{1}{6 \alpha\eta}\|u^{k+1} 
            - v^k \|^2\right)  
            \\&+ \frac{2(1 - \alpha)}{\alpha}(\mathrm{D}_{\Psi}(u^k, z^*) - \mathrm{D}_{\Psi}(v^k, z^*)) -2 \mathrm{D}_{\Psi}(v^{k}; z^*)
            - \mu_B\|\hat z^{k+1} - z^*\|^2.
\end{align*}
Since $\eta \leq \frac{1}{3L_{\Psi}\alpha} $ (by \eqref{ae:choice}), then
\begin{align*}
    \frac{1}{\eta}\left\|
            z^{k+1} - z^*
           \right\|^2 
		\leq&\frac{1}{\eta}\left\|
            z^{k} - z^*\right\|^2 + 2 \eta\left \| 
             G^k (\hat z^{k+1})  
            \right\|^2
		 -\frac{2}{3\eta}\left\|
              \hat z^{k+1} - z^{k}\right \|^2
            \\&-\frac{2}{\alpha}\left( \left\langle 
            \nabla \Psi(v^{k}) -  \nabla \Psi(z^*) , 
             u^{k+1} - v^k \right\rangle + \frac{L_{\Psi}}{2}\|u^{k+1} 
            - v^k \|^2\right)
            \\&+ \frac{2(1 - \alpha)}{\alpha}(\mathrm{D}_{\Psi}(u^k, z^*) - \mathrm{D}_{\Psi}(v^k, z^*)) -2 \mathrm{D}_{\Psi}(v^{k}, z^*)
             - \mu_B\|\hat z^{k+1} - z^*\|^2 .
\end{align*}
 $L_{\Psi}$-smoothness of $\Psi(z)$ 
 gives
\begin{align*}
    \frac{1}{\eta}\left\|
            z^{k+1} - z^*\right\|^2 
		\leq&\frac{1}{\eta}\left\|
            z^{k} - z^*\right\|^2 + 2 \eta\left \| 
             G^k (\hat z^{k+1})  
            \right\|^2
		 -\frac{2}{3\eta}\left\|
              \hat z^{k+1} - z^{k}\right \|^2
            \\&-\frac{2}{\alpha}(\mathrm{D}_{\Psi}(u^{k+1}, z^*) - \mathrm{D}_{\Psi} (v^k, z^*)) 
            + \frac{2(1 - \alpha)}{\alpha}(\mathrm{D}_{\Psi}(u^k, z^*) - \mathrm{D}_{\Psi}(v^k, z^*)) 
            \\&-2 \mathrm{D}_{\Psi} (v^{k}, z^*)
             - \mu_B\|\hat z^{k+1} - z^*\|^2
        \\=&\frac{1}{\eta}\left\|
            z^{k} - z^* \right\|^2 + 2\left(  \eta\| 
             G^k (\hat z^{k+1})\|^2 -\frac{1}{6\eta}\|\hat z^{k+1} - z^k\|^2 \right)
            -\frac{2}{\alpha}\mathrm{D}_{\Psi}(u^{k+1}, z^*) 
            \\&+ \frac{2(1 - \alpha)}{\alpha}\mathrm{D}_{\Psi}(u^k, z^*)  - \mu_B\left(\|\hat z^{k+1} - z^*\|^2 + \frac{1}{3\mu_B\eta}\|\hat z^{k+1} - z^k\|^2\right).
\end{align*}
Since $\eta \leq \frac{1}{3\mu_B}$ (by \eqref{ae:choice}) and using inequality $-\|a - b\|^2 \geq - 2\|a\|^2 - 2\|b\|^2$, we get
\begin{align*}
    \frac{1}{\eta}\left\|
            z^{k+1} - z^*\right\|^2 
		\leq&\frac{1}{\eta}\left\|
            z^{k} - z^*\right\|^2 + 2\left(  \eta\| 
             G^k (\hat z^{k+1})\|^2   - \frac{1}{6\eta}\|\hat z^{k+1} - z^k\|^2 \right) 
             \\&- \frac{2}{\alpha}\mathrm{D}_{\Psi}(u^{k+1}, z^*)  + \frac{2(1 - \alpha)}{\alpha}\mathrm{D}_{\Psi}(u^k, z^*) 
             - \mu_B\| z^{k} - z^*\|^2.
\end{align*}
Since \eqref{aux:grad_app}, we get 
\begin{align*}
    \frac{1}{\eta} \|z^{k+1} - z^*\|^2 + 
		\frac{2}{\alpha}\mathrm{D}_{\Psi}(u^{k+1}, z^*)  &\leq \frac{1}{\eta}\left(1 - \mu_B\eta\right)\|z^{k} - z^*\|^2 + \frac{2(1 - \alpha)}{\alpha}\mathrm{D}_{\Psi}(u^k, z^*) 
 \\&\leq(1 - \alpha)\left[\frac{1}{\eta}\|z^{k} - z^*\|^2 
 + \frac{2}{\alpha}\mathrm{D}_{\Psi}(u^k, z^*)\right].
\end{align*}
In the last inequality we use that $\alpha > \frac{\alpha}{3}$, $\eta \mu_B \geq \frac{\alpha}{3}$. If $L_{\Psi} \leq \mu_B$, then $\alpha = 1$, $\eta \mu_B = \frac{1}{3} = \frac{\alpha}{3}$. If $L_{\Psi} > \mu_B$, then $\alpha = \sqrt{\frac{\mu_B}{L_{\Psi}}}$, $\eta \mu_B = \sqrt{\frac{\mu_B}{3 L_{\Psi}}} \geq \frac{\alpha}{3}$.

By \eqref{ae:Phi} definition of $\mathrm{\Phi}^k$, we get
\begin{align*}
    \mathrm{\Phi}^{k+1} \leq \left(1 - \frac{\alpha}{3}\right) \mathrm{\Phi}^k.
\end{align*}
\end{proof}

\begin{lemma} \label{lem2}
	Assume that for the problem \eqref{sliding:eq:prox} we use Extra Step method with starting points $z^k$ and number of iterations:
	\begin{equation}
	\label{T1}
		T = \mathcal{O}  \left(\left(1 + \eta L_B\right)\log(1 + \eta L_B) \right).
	\end{equation}
	Then for an output $\hat{z}^T$ it holds that 
	\begin{equation*}
		\| G^k (\hat z^{T})\|^2  \leq \frac{1}{6\eta^2}\|\hat z^{T} - z^k\|^2.
	\end{equation*}
\end{lemma}
\begin{proof} Due to \eqref{sliding:eq:prox} is $\left(\frac{1}{\eta} +  L_B\right)$-Lipschitz, we get the following inequality
\begin{align*}
    \|G^k (z)\|^2 \leq\left(L_B + \frac{1}{\eta}\right)^2\| z - z^{k+1}_*\|^2,
\end{align*}
where $z^{k+1}_*$ is the solution to the problem \eqref{sliding:eq:prox}. It means that $\frac{\frac{1}{6\eta^2}\|\hat z^{k+1} - z^k\|^2}{\left(L_B + \frac{1}{\eta}\right)^2}$-solution to \eqref{sliding:eq:prox} satisfies condition \eqref{aux:grad_app}. To get this solution, the Extra Step method needs the following number of iterations
\begin{equation*}
		T = \mathcal{O}  \left(\left(1 + \eta L_B\right)\log(\eta L_B + 1) \right).
\end{equation*}
It follows from the convergence estimates for the Extra Step method and the fact that the problem \eqref{sliding:eq:prox} is $\frac{1}{\eta}$-strongly monotone, as well as $\left(\frac{1}{\eta} +  L_B\right)$-Lipschitz.
\end{proof}

\begin{theorem}\label{th:sliding_opt_comm:var_ineq}
Let Algorithm \ref{alg:sliding_opt_comm_v2} be applied for solving \eqref{pr:PF2} under Assumptions \ref{ass:var_ineq:lipschitz}-\ref{ass:var_ineq:smooth}, with $\sqrt{\frac{L_{\Psi}}{\mu_B}} \leq \frac{L_B}{\mu_B}$. Then, to find an $\varepsilon$-solution to the problem \eqref{pr:PF2},  \Cref{alg:sliding_opt_comm_v2} requires
\begin{equation*}
	\mathcal{O}\left(\max\left\{1, \sqrt{\frac{L_{\Psi}}{\mu_B}} \right\}\log\frac{1}{\varepsilon}\right) ~~ \text{oracle calls of} \ \ \nabla \Psi \ \ \text{and}
\end{equation*} 
\begin{equation*}
    \mathcal{O}\left(\frac{L_B}{\mu_B}\log \frac{L_B}{\mu_B}\log\frac{1}{\varepsilon}\right) ~~ \text{oracle calls of} \ B.
\end{equation*} 
\end{theorem}

\begin{proof}
Using the property of the Bregman divergence ($\mathrm{D}_{\Psi}(z_1, z_2) \geq 0 $) and running the recursion \eqref{ae:rec}, we get point $z^k$ after $k$ iterations such that
\begin{equation*}
    \frac{1}{\eta}\|z^k - z^*\|^2  \leq \mathrm{\Phi}^k \leq C \left(1 - \frac{\alpha}{3}\right)^k,
\end{equation*}
where $C$ is defined as
\begin{equation*}
    C = \frac{1}{\eta}\|z^0 - z^*\|^2 + \frac{2}{\alpha}\mathrm{D}_{\Psi}(u^0, z^*).
\end{equation*}
It means, that after $K = \frac{3}{\alpha} \log \frac{1}{\varepsilon}$ iterations of \Cref{alg:sliding_opt_comm_v2} we get a point $z^K$ satisfies the following inequality
\begin{equation*}
    \|z^K - z^*\|^2 \leq \varepsilon.
\end{equation*}

Also, \Cref{alg:sliding_opt_comm_v2} needs to find point $\hat z^{k+1}$ satisfies \eqref{aux:grad_app} on each iteration. By \Cref{lem2}, this point can be find after $T = \mathcal{O}  \left(\left(1 + \eta L_B\right)\log(1 + \eta L_B) \right)$ iterations of the Extra Step method. Moreover, Extra Step is not requires to call oracle $\nabla \Psi(z)$. It means that \Cref{alg:sliding_opt_comm_v2} calls oracle $\nabla \Psi(z)$ only $K$ times with
\begin{equation*}
    K = \frac{3}{\alpha} \log \frac{1}{\varepsilon} = \mathcal{O}\left(\max\left\{1, \sqrt{\frac{L_{\Psi}}{\mu_B}} \right\}\log\frac{1}{\varepsilon}\right)
\end{equation*} and oracle $B(z)$ only $K \times T$ times, where
\begin{align*}
    K \times T &=  \mathcal{O}\left(\frac{3}{\alpha}(1 + \eta L_B)\log(1 + \eta L_B)\log \frac{1}{\varepsilon}\right)
    \\& = \mathcal{O}\left(\frac{3}{\alpha}\left(1 + \min\left\{\frac{1}{3\mu_B}, \frac{1}{3L_{\Psi}\alpha}\right\} L_B\right)\log\left(1 + \min\left\{\frac{1}{3\mu_B}, \frac{1}{3L_{\Psi}\alpha}\right\} L_B\right)\log \frac{1}{\varepsilon}\right)
    \\& = \mathcal{O}\left( \frac{L_B}{\mu_B}\log\frac{L_B}{\mu_B}\log \frac{1}{\varepsilon}\right)
\end{align*}
after $K$ iterations of \Cref{alg:sliding_opt_comm_v2}.
\end{proof}

To get \textbf{\Cref{th:sliding_opt_comm}}, we apply \Cref{alg:sliding_opt_comm_v2}  for solving the problem \eqref{PF} with $B(z) = [\nabla_X F(X, Y)^T, -\nabla_Y F(X,Y)^T]^T$, $\Psi(z) = \frac{\lambda}{2}\langle z, Wz \rangle = \frac{\lambda}{2}\langle X, WX \rangle + \frac{\lambda}{2}\langle Y, WY \rangle$. As mentioned above, for this case $L_B = L$, $\mu_B = \mu$ and $L_{\Psi} = \lambda \lambda_{\max}(W)$.

\subsection{Proof of \Cref{theorem_sum_structure}}

\begin{lemma}\label{lem2_sum}
	Assume that for solving the problem \eqref{sliding:eq:prox}, where $B(z) = \frac{1}{r}\sum_{i=1}^r B_i(z)$ we use Randomized Extra Step Method from \cite{alacaoglu2021stochastic} with a starting points $z^k$ and number of iterations:
\begin{equation}
	\label{T1_sum}
		T = \mathcal{O}  \left(\left(r+ \sqrt{r}\left(1 + \eta L_B\right)\log\left(1 + \eta L_B\right)\right) \right).
	\end{equation}
	Then, for an output $\hat z^T$ it holds that 
	\begin{equation*}
		\mathbb{E}\sqn{G^k(\hat z^T)} \leq \mathbb{E}\sqn{\hat z^T - z^k}.
	\end{equation*}
\end{lemma}
\begin{proof}
Proof of this lemma is similar to the proof of \Cref{lem2} and follows from the convergence estimates for the Randomized Extra Step method from \cite{alacaoglu2021stochastic} and the fact that the problem \eqref{sliding:eq:prox} is $\frac{1}{\eta}$-strongly monotone, as well as $\left(\frac{1}{\eta} +  L_B\right)$-Lipschitz.
\end{proof}

\begin{theorem}\label{th:sliding_opt_comm:var_ineq_sum}
Let Algorithm \ref{alg:sliding_opt_comm_v2} be applied for solving \eqref{pr:PF2} with $B(z) =  \frac{1}{r}\sum_{j=1}^rB_j(z)$ under Assumptions \ref{ass:var_ineq:lipschitz}-\ref{ass:var_ineq:smooth}, with $\sqrt{\frac{L_{\Psi}}{\mu_B}} \leq \frac{L_B}{\mu_B}$. Then, to find an $\varepsilon$-solution to the problem \eqref{pr:PF2},  \Cref{alg:sliding_opt_comm_v2} requires
\begin{equation*}
	\mathcal{O}\left(\max\left\{1, \sqrt{\frac{L_{\Psi}}{\mu_B}} \right\}\log\frac{1}{\varepsilon}\right) ~~ \text{oracle calls of} \ \ \nabla \Psi \ \ \text{and}
\end{equation*} 
\begin{equation*}
    \mathcal{O}\left(\left(r\sqrt{\frac{L_{\Psi}}{\mu_B}} + \sqrt{r}\frac{L_B}{\mu_B}\right)\log \frac{L_B}{\mu_B}\log\frac{1}{\varepsilon}\right) ~~ \text{oracle calls of} \ B_j.
\end{equation*} 
\end{theorem}

\begin{proof}
Proof of this theorem is similar to the proof of \Cref{th:sliding_opt_comm:var_ineq}. As mentioned in \Cref{th:sliding_opt_comm:var_ineq} after $K = \frac{3}{\alpha} \log \frac{1}{\varepsilon}$ iterations of \Cref{alg:sliding_opt_comm_v2} we get a point $z^K$ satisfies the following inequality
\begin{equation*}
    \mathbb{E}\|z^K - z^*\|^2 \leq \varepsilon.
\end{equation*}

Also, \Cref{alg:sliding_opt_comm_v2} needs to find point $\hat z^{k+1}$ satisfies \eqref{aux:grad_app} on each iteration. By \Cref{lem2}, this point can be find after $T = \mathcal{O}  \left(\left(r + \sqrt{r}\eta L_B\right)\log(1 + \eta L_B) \right)$ iterations of Randomized Extra Step method. Moreover, this method is not requires to call oracle $\nabla \Psi(z)$. It means that \Cref{alg:sliding_opt_comm_v2} calls oracle $\nabla \Psi(z)$ only $K$ times with
\begin{equation*}
    K = \frac{3}{\alpha} \log \frac{1}{\varepsilon} = \mathcal{O}\left(\max\left\{1, \sqrt{\frac{L_{\Psi}}{\mu_B}} \right\}\log\frac{1}{\varepsilon}\right)
\end{equation*} and oracle $B_j(z)$ only $K \times(r + T)$ times, where
\begin{align*}
    K \times (r + T) &=  \mathcal{O}\left(\frac{3}{\alpha}\left(r + r + \sqrt{r}\left(1 + \eta L_B\right)\right)\log(1 + \eta L_B)\log \frac{1}{\varepsilon}\right)
    \\& = \mathcal{O}\left(\frac{3}{\alpha}\left(r + \sqrt{r}\min\left\{\frac{1}{3\mu_B}, \frac{1}{3L_{\Psi}\alpha}\right\} L_B\right)\log\left(1 + \min\left\{\frac{1}{3\mu_B}, \frac{1}{3L_{\Psi}\alpha}\right\} L_B\right)\log \frac{1}{\varepsilon}\right)
    \\& = \mathcal{O}\left(\left( r\sqrt{\frac{L_{\Psi}}{\mu_B}} + \sqrt{r}\frac{L_B}{\mu_B}\right)\log\frac{L_B}{\mu_B}\log \frac{1}{\varepsilon}\right),
\end{align*}
after $K$ iterations of \Cref{alg:sliding_opt_comm_v2} with Randomized Extra Step method.
\end{proof}

To get \textbf{\Cref{theorem_sum_structure}}, we apply \Cref{alg:sliding_opt_comm_v2}  for solving the problem \eqref{PF}, where $f_m(x_m, y_m) = \frac{1}{r}\sum_{j=1}^rf_{m,j}(x_m, y_m)$ with $B_j(z) = [\nabla_X F_j(X, Y)^T, -\nabla_Y F_j(X,Y)^T]^T$, $\Psi(z) = \frac{\lambda}{2}\langle z, W z \rangle = \frac{\lambda}{2}\langle X, WX \rangle + \frac{\lambda}{2}\langle Y, WY \rangle$. As mentioned above, for this case $L_B = L$, $\mu_B = \mu$ and $L_{\Psi} = \lambda \lambda_{\max}(W)$.
\subsection{Proof of \Cref{th:sliding_big_lambda}}

\Cref{alg:sliding_vi} is adapted version of \Cref{alg:sliding_big_lambda} in introduced notation.
\begin{algorithm}[t] 
	\caption{\Cref{alg:sliding_big_lambda} in new notation}
	\label{alg:sliding_vi}
	\hspace*{\algorithmicindent} {\bf Parameters:} stepsize $\eta$\\
	\hspace*{\algorithmicindent} {\bf Initialization:} choose  $ x^0,y^0$, $x^0_m = x^0$, $y^0_m = y^0$  for all $m$, form matrix $z_0$
	\begin{algorithmic}[1]
		\For{$k=0,1,2,\ldots$}
		\State $v^k = z^k - \eta B(z^k)$
		\label{alg:sliding_vi:step2}
		\State 
		Find an approximate solution $\hat u^{k}$ of $ G^k(z) = 0$, where
		\begin{equation}\label{inner_problem_sliding_vi}
		\begin{split}
			 G^k(z) = \nabla \Psi(z) + \frac{z - v^k}{\eta} 
		\end{split}
		\end{equation}
		\State  
		\label{alg:sliding:step4}
		$
		  z^{k+1} = \hat{u}^{k} + \eta \left( B(z^{k})  -  B(\hat{u}^{k})\right)
		$
		\EndFor
	\end{algorithmic}
\end{algorithm}


\begin{lemma} \label{lem1}
For Algorithm \ref{alg:sliding_vi} it holds:
	\begin{align*}
    \sqn{z^{k+1} - z^*} &\leq (1 - \eta \mu_B)\sqn{z^k - z^*} + \left(2 + \frac{4\eta L_B^2}{\mu_B} +\frac{4}{\eta \mu_B} + 4\eta^2 L_B^2\right)\sqn{u^k - \hat u^k}  
	\\&\hspace{0.4cm} -(1 - 3\eta \mu_B - 4\eta^2 L_B^2) \sqn{z^k - \hat u^k}.
	\end{align*}
\end{lemma}
\begin{proof} Let us use the additional notation $u^{k}$ for the solution of the problem \eqref{inner_problem_sliding_vi} on iteration $k$ for short.
	\begin{align*}
		\left\|z^{k+1} - z^*\right\|^2 &=
		\sqn{z^k - z^*} + 2\<z^{k+1} - z^k, z^k - z^*> + \sqn{z^{k+1} - z^k} 
		\\&=
		\sqn{z^k - z^*} + 2\<z^{k + 1} - z^k,z^k - u^k> + 2\<z^{k+1} - z^k, u^k - z^*>  
		\\&\quad 
		+ \sqn{z^{k + 1} - z^k}
		\\&=
		\sqn{z^k - z^*}  + \sqn{z^{k+1} - u^k} - \sqn{z^k - u^{k}} + 2\<z^{k+1} - z^k, u^k - z^*> 
		\\&= 
		\sqn{z^k - z^*}  
		+ \sqn{z^{k+1} - u^k} - \sqn{z^k - u^k}
        \\&\quad
        + 2\< \hat{u}^{k} + \eta \left(B (z^k) - B( \hat{u}^{k})\right) - z^k, u^k - z^*>
		\\&= 
		\sqn{z^k - z^*} + \sqn{z^{k+1} - u^k} - \sqn{z^k -  u^k} 
        \\&\hspace{0.4cm}+ 2\< \hat{u}^k + \eta B(z^k)  - z^k, u^k - z^*> 
		 + 2\< - \eta B(\hat{u}^k), u^k - z^*> .
	\end{align*}
With expressions for $v^k$ from Algorithm \ref{alg:sliding_vi}, we have
	\begin{align*}
\sqn{z^{k+1} - z^*} &\leq
	\sqn{z^k - z^*} + \sqn{z^{k+1} -  u^k} - \sqn{z^k - u^k}
	\\&\hspace{0.4cm} + 2\< \hat u^k - v^k, u^k - z^*>  + 2\< - \eta B(\hat u^k), u^k - z^*>  
	\\&= 
	\sqn{z^k - z^*} + \sqn{z^{k+1} - u^k} - \sqn{z^k - u^k} 
	\\&\hspace{0.4cm} + 2\< \hat u^k - u^k, u^k - z^*> + 2\< u^k - v^k ,  u^k - z^*>   + 2\< -\eta B(\hat u^k), u^k - z^*> .
	\end{align*}
According to the optimal condition for $u^k$,
we get
    \begin{align}\label{eq:sliding_vi}
\sqn{z^{k+1} - z^*} &\leq
	\sqn{z^k - z^*} + \sqn{z^{k+1} - u^k} - \sqn{z^k - u^k}
    \notag\\&\hspace{0.4cm} + 2\< \hat u^k - u^k , u^k - z^*> + 2\< -\eta \nabla \Psi(u^k), u^k - z^*>  + 2\< - \eta B(\hat{u}^k),  u^k - z^*> 
	\notag\\&=
	\sqn{z^k - z^*} + \sqn{z^{k+1} - u^k} - \sqn{z^k - u^k}
    \notag\\&\hspace{0.4cm} + 2\< \hat u^k - u^k , u^k - z^*> + 2\< -\eta  \nabla \Psi(u^k), u^k - z^*> 
    \notag\\&\hspace{0.4cm} + 2\< - \eta B(u^k), u^k - z^*> + 2\< - \eta \left(B (\hat u^k) - B(u^k)\right), u^k - z^*> .
	\end{align}
With property of the solution $z^*$: 
$$\<B(z^*) + \nabla \Psi(z^*), z^* - z > \leq 0, \forall z \in \mathbb{R}^{M(d_x + d_y)}$$
Using $\mu_B$-strong monotonicity  of $B(z)$ and convexity of $\Psi(z)$, we obtain
    \begin{align*}
\sqn{z^{k+1} - z^*} &\leq
	\sqn{z^k - z^*} + \sqn{z^{k+1} - u^k} - \sqn{z^k - u^k}  
	\\&\hspace{0.4cm} + 2\< \hat u^k - u^k, u^k - z^*> - 2\< \eta (\nabla \Psi(u^k) + B (u^k) -  \nabla \Psi (z^*) - B(z^*)), u^k - z^*> 
    \\&\hspace{0.4cm}- 2\< \eta \left(B(\hat u^k) - B( u^k)\right),  u^k - z^*>  
	\\&\leq
	\sqn{z^k - z^*} + \sqn{z^{k+1} - u^k} - \sqn{z^k - u^k} - 2 \eta \mu_B \sqn{ u^k - z^*} 
	\\&\quad
	- 2\< \eta \left(B(\hat u^k) - B(u^k)\right) - (\hat u^k - u^k), u^k - z^*> .
	\end{align*}
By Young's inequality, we have
	\begin{align*}
\sqn{z^{k+1} - z^*} &\leq
	\sqn{z^k - z^*} + \sqn{z^{k+1} - u^k} - \sqn{z^k - u^k}  - 2 \eta \mu_B \sqn{ u^k - z^*}  
	\\&\quad + \frac{2}{\eta \mu_B}\sqn{\eta \left(B(\hat u^k) - B( u^k)\right)  - (\hat u^k - u^k)}  + \frac{\eta \mu_B}{2}\sqn{ u^k - z^*} 
	\\&\leq 
	\sqn{z^k - z^*} + \sqn{\hat u^k + \eta \left(B(z^k) - B(\hat u^k)\right) -  u^k} - \sqn{z^k -  u^k} - \frac{3 \eta \mu_B}{2} \sqn{  u^k - z^*} 
    \\&\hspace{0.4cm} + \frac{4}{\eta \mu_B}\sqn{\eta \left(B (\hat u^k) - B( u^k)\right)}  + \frac{4}{\eta \mu_B}\sqn{\hat u^k - u^k}  
	\\&\leq 
	\sqn{z^k - z^*} + \left(2 + \frac{4}{\eta \mu_B}\right)\sqn{\hat u^k - u^k}  + 2\sqn{\eta  \left(B(z^k) - B(\hat u^k)\right)} - \sqn{z^k - u^k} 
 \\&\hspace{0.4cm} - \frac{3 \eta \mu_B}{2} \sqn{u^k - z^*}  + \frac{4}{\eta \mu_B}\sqn{\eta  \left(B(\hat u^k) - B(u^k)\right)}.
	\end{align*}
Then we use $L_B$-Lipschitzness of $B(z)$
    \begin{align*}
\sqn{z^{k+1} - z^*} &\leq
	\sqn{z^k - z^*} + \frac{4\eta L_B^2}{\mu_B}\sqn{ \hat u^k - u^k} + 2\eta^2 L_B^2\sqn{z^k - \hat u^k}  + \left(2 + \frac{4}{\eta \mu_B}\right)\sqn{\hat u^k - u^k}   
	\\&\hspace{0.4cm} - \sqn{z^k - u^k} - \frac{3 \eta \mu_B}{2} \sqn{ u^k - z^*} 
	\\&\leq
	\sqn{z^k - z^*}  + \left(2 + \frac{4\eta L_B^2}{\mu_B} +\frac{4}{\eta \mu_B} + 4\eta^2 L_B^2\right)\sqn{\hat u^k - u^k} 
	\\&\hspace{0.4cm}-(1 - 4\eta^2 L_B^2) \sqn{z^k - u^k} - \frac{3 \eta \mu_B}{2} \sqn{ u^k - z^*} .
	\end{align*}
By inequality $\sqn{a+b} \geq \frac{2}{3}\sqn{a} - 2\sqn{b}$, we have
	\begin{align*}
    \sqn{z^{k+1} - z^*}
    &\leq (1 - \eta \mu_B)\sqn{z^k - z^*}  + \left(2 + \frac{4\eta L_B^2}{\mu_B} +\frac{4}{\eta \mu_B} + 4\eta^2L_B^2\right)\sqn{\hat u^k -  u^k} 
    \\&\hspace{0.4cm}-(1 - 3\eta \mu_B - 4\eta^2 L_B^2) \sqn{z^k - u^k}.
	\end{align*}

\end{proof}

\begin{theorem}
Assume that the problem \eqref{inner_problem_sliding_vi} be solved with relative precision $\tilde \delta$ (i.e. $ \| \hat u^k - u^k \|^2 \leq \tilde \delta \| z^k - u^k \|^2$):
\begin{align}
\label{e1}
    \tilde \delta = \frac{1}{2\left(2 + \frac{4 \eta L_B}{\mu_B} + \frac{4}{\eta \mu_B} + 4\eta^2L_B\right)}.
\end{align} Additionally, stepsize $\eta$ is
\begin{align}
\label{gamma1}
    \eta = \min\left\{ \frac{1}{12 \mu_B}; \frac{1}{16L_B}\right\}.
\end{align}
Then, Algorithm \ref{alg:sliding_vi} converges linearly to the solution $z^*$ and it holds that $\sqn{z^{k+1} - z^*} \leq \varepsilon$ after 
\begin{align}
\label{K1}
    K = \mathcal{O} \left(\frac{1}{\eta \mu_B} \log \frac{\sqn{z^{0} - z^*}}{\varepsilon} \right)\quad \text{iterations}.
\end{align}
\end{theorem}

\begin{proof} The result from \Cref{lem1} gives
\begin{align*}
    \sqn{z^{k+1} - z^*} &\leq
	(1 - \eta \mu_B)\sqn{z^k - z^*}  + \left(2 + \frac{4\eta L_B^2}{\mu_B} +\frac{4}{\eta \mu_B} + 4\eta^2 L_B^2\right)\tilde \delta  \sqn{z^k - u^k}
	\\&\hspace{0.4cm}-(1 - 3\eta \mu_B - 4\eta^2 L_B^2) \sqn{z^k - u^k}.
\end{align*}
With the choice $\tilde \delta$ from \eqref{e1}, we obtain
\begin{align*}
	\sqn{z^{k+1} - z^*}&\leq (1 - \eta \mu_B)\sqn{z^k - z^*} -\left(\frac{1}{2} - 3\eta \mu_B - 4\eta^2 L_B^2\right) \sqn{z^k - u^k}.
\end{align*}
The proof is completed by choosing $\eta$ from \eqref{gamma1}.
\end{proof}

We apply \Cref{alg:sliding_vi}  for solving the problem \eqref{PF} with $B(z) = [\nabla_X F(X, Y)^T, -\nabla_Y F(X,Y)^T]^T$, $\Psi(z) = \frac{\lambda}{2}\langle z, Wz \rangle = \frac{\lambda}{2}\langle X, WX \rangle + \frac{\lambda}{2}\langle Y, WY \rangle$. The inner problem \eqref{inner_problem_sliding_vi} in Algorithm \ref{alg:sliding_vi} is equal to the inner problem \eqref{inner_problem_big_lambda} in Algorithm \ref{alg:sliding_big_lambda}.

\begin{lemma} \label{lem_fgm}
	Assume that for the problem \eqref{inner_problem_big_lambda} we use fast gradient method with starting points $X^k, Y^k$ and number of iterations:
	\begin{equation}
	\label{T1_big_lambda}
		T = \mathcal{O}  \left(\left(1 + \sqrt{\frac{\lambda \lambda_{\max}(W) + \frac{1}{\eta}}{\lambda\lambda_{\min}^+(W) + \frac{1}{\eta}}}\right)\log\frac{1}{\tilde \delta} \right).
	\end{equation}
	Then for an output $u^k$ it holds that 
	\begin{equation*}
		\sqn{u^k - \hat{u}^k} \leq \tilde \delta\sqn{z^k - \hat u^k}.
	\end{equation*}
\end{lemma}
\textbf{Proof:} We solve \eqref{inner_problem_big_lambda} on the line 3 of Algorithm \ref{alg:sliding_big_lambda}. This saddle point problem is divided into 2 minimization problems. Then they can be solved by Fast Gradient Method \cite{nesterov2018lectures}. The complexity of solving any of these two problems is $ \mathcal{\tilde O}(1)$ on the $\textbf{Ker }W$ and is $ \mathcal{\tilde O}\left(1 + \sqrt{\frac{\lambda \lambda_{\max}(W) + \frac{1}{\eta}}{\lambda\lambda_{\min}^+(W) + \frac{1}{\eta}}}\right)$ on the $\left(\textbf{Ker }W\right) ^{\perp}$. It follows from the convergence estimates for Fast Gradient Method and the fact that problems \eqref{inner_problem_big_lambda} are $\frac{1}{\eta}$-strongly-convex-strongly-concave, as well as $\frac{1}{\eta}$-smooth on the  $\textbf{Ker }W$ and $\lambda\lambda_{\min}^+(W) + \frac{1}{\eta}$-strongly-convex-strongly-concave, as well as $\lambda \lambda_{\max}(W) + \frac{1}{\eta}$-smooth on the $\left(\textbf{Ker }W\right) ^{\perp}$.
\EndProof

To get prove of \textbf{Theorem \ref{th:sliding_big_lambda}}. We note that \eqref{K1} also corresponds to calls of $f_m(x_m, y_m)$ gradients which in turn corresponds to the number of local computation on each node. Substitution of \eqref{gamma1} in \eqref{K1} gives that
\begin{align*}
    K = \mathcal{O} \left( \left( 1 + \frac{L}{\mu} \right) \log \frac{\sqn{X^{0} - X^*} + \sqn{Y^{0} - Y^*}}{\varepsilon} \right).
\end{align*}
It is also easy to estimate the total number of communication rounds:
\begin{align*}
    K \times T  &= 
    \mathcal{O} \left( \frac{1}{\eta \mu} \left( 1 + \sqrt{\frac{\lambda \lambda_{\max}(W)+ \frac{1}{\eta}}{\lambda\lambda_{\min}^+(W) + \frac{1}{\eta}}}\right)\log\frac{1}{\tilde \delta} \log \frac{\sqn{z^{0} - z^*} }{\varepsilon} \right) 
    \\&\leq 
    \mathcal{O} \left( \frac{1}{\eta \mu} \left(1 +  \sqrt{1 + \frac{\lambda \lambda_{\max}(W) }{\lambda\lambda_{\min}^+(W) + \frac{1}{\eta}}}\right)\log\frac{1}{\tilde \delta} \log \frac{\sqn{z^{0} - z^*}}{\varepsilon} \right)
    \\&\leq \mathcal{O} \left( \frac{1}{\eta \mu} \left(1 +  \sqrt{1 + \frac{\lambda \lambda_{\max}(W) }{\max\left\{\lambda\lambda_{\min}^+(W), \frac{1}{\eta}\right\}}}\right)\log\frac{1}{\tilde \delta} \log \frac{\sqn{z^{0} - z^*}}{\varepsilon} \right)
    \\&= \mathcal{O} \left( \frac{1}{\eta \mu} \left(1 +  \sqrt{\min\left\{\chi, \eta \lambda \lambda_{\max}(W))\right\}}\right)\log\frac{1}{\tilde \delta} \log \frac{\sqn{z^{0} - z^*}}{\varepsilon} \right)
    \\&\leq\mathcal{O} \left( \frac{1}{\eta \mu} \left(1 +  \sqrt{\min\left\{\chi, \frac{\lambda \lambda_{\max}(W)}{L}\right\}}\right)\log\frac{1}{\tilde \delta} \log \frac{\sqn{z^{0} - z^*}}{\varepsilon} \right) 
    \\&= \mathcal{O} \left( \frac{1}{\eta \mu}  \sqrt{\min\left\{\chi, \frac{\lambda \lambda_{\max}(W)}{L}\right\}}\log\frac{1}{\tilde \delta} \log \frac{\sqn{z^{0} - z^*}}{\varepsilon} \right)
    \\&= \mathcal{O} \left(  \left(1+ \frac{L}{\mu}\right)\sqrt{\min\left\{\chi, \frac{\lambda \lambda_{\max}(W)}{L}\right\}}\log\frac{1}{\tilde \delta} \log \frac{\sqn{z^{0} - z^*}}{\varepsilon} \right) 
    \\&= \mathcal{O} \left(   \frac{L}{\mu}\sqrt{\min\left\{\chi, \frac{\lambda \lambda_{\max}(W)}{L}\right\}}\log\frac{1}{\tilde \delta} \log \frac{\sqn{z^{0} - z^*}}{\varepsilon} \right)
    \\&= \mathcal{O} \left( \min\left\{\frac{L}{\mu}\sqrt{\chi}, \frac{\sqrt{\lambda\lambda_{\max}(W)L}}{\mu}\right\} \log\frac{1}{\tilde \delta} \log \frac{\sqn{X^{0} - X^*} + \sqn{Y^{0} - Y^*}}{\varepsilon} \right).
\end{align*}
Here we use that $\eta = \min\left\{ \frac{1}{12 \mu_B}; \frac{1}{16L_B}\right\} = \min\left\{ \frac{1}{12 \mu}; \frac{1}{16L}\right\}$.


\subsection{Proof of \Cref{theorem_randomized}} \label{proof_t5}

Now we move from Slidings (Algorithms \ref{alg:sliding_opt_comm} and \ref{alg:sliding_big_lambda}) to the proof of Algorithm \ref{alg_sum}.

\begin{algorithm}[h]
	\caption{\Cref{alg_sum} in new notation}
	\label{alg_sum_vi}
	\hspace*{\algorithmicindent} {\bf parameters:} stepsize $\eta$, probability $p$, probability $\rho$\\
	\hspace*{\algorithmicindent} {\bf initialization:} choose  $ z^0 $, $z^0_m = z^0$ for all $m$
	\begin{algorithmic}[1]
\For {$k=0,1, 2, \ldots$ }
\State $\bar z^k = (1 - \rho) z^k + \rho u^k$

\State $z^{k+1/2} = \bar z^k - \eta  (B(u^k) + \nabla \Psi (u^k))$, 

\Statex Generate $\xi^k =  \begin{cases}
 1,&  \text{with probability} ~~ 1 - p \\
0 ,& \text{with probability} ~~ p
\end{cases},$
\label{alg_sum_vi:step5}
\Statex \ \ If $\xi^k = 0$: \label{alg3_sum_vi:step6}
\State \ \ \ $G^k = \frac{1}{p}\left(\nabla \Psi\left(z^{k+1/2}\right) - \nabla \Psi(u^k)\right)$
\Statex \ \ If $\xi^k = 1$: \label{alg_sum_vi:step9}
\Statex \ \ \  Generate an vector of indexes $\hat{\xi}_k$ according to $Q$
\State \ \ $G^k = \frac{1}{1-p}\left(B_{\hat{\xi}_k}(z^{k+1/2}) - B_{\hat{\xi}_k}(u^{k})\right)$

\State  $z^{k+1} = \bar z^k - \eta \left( G^k + B(u^k)+ \nabla \Psi( u^k)\right)$ 

\Statex Generate $\xi^{k+1/2}=  \begin{cases}
1,&  \text{with prob.} ~~ 1 - \rho \\
0 ,& \text{with prob.} ~~ \rho
\end{cases},$
\State $u^{k+1} = \xi^{k+1/2} u^k + (1 - \xi^{k+1/2}) z^{k+1}$
\EndFor
\end{algorithmic}
\end{algorithm}

Let us consider our problem as a finite sum problem with $r+1$ terms: 
\begin{align*}
    \text{Find} \ z \ \text{such that}: B(z) + \nabla \Psi (z) = 0, \ \ \text{where} \ \ B(z) = \frac{1}{r}\sum_{i=1}^r B_j(z).
\end{align*}
For such a problem, one can use the results of the convergence of the variance reduction method \cite{alacaoglu2021stochastic} on which our method is based.

Let us use the additional notation $G (z) = \frac{\xi^k}{1- p}  B_{\hat{\xi}_k}(z) + \frac{1 - \xi^k}{p} \nabla \Psi(z)$ for short.

In the strongly-monotone case (Section 4.3 from \cite{alacaoglu2021stochastic}), the estimates on the number of iterations is 
\begin{align*}
    \mathcal{ O} \left(\left(\frac{1}{\rho} + \frac{ L_{\text{eff}} }{\sqrt{\rho}\mu}\right)\log\frac{1}{\varepsilon}\right), 
\end{align*}
where $L_{\text{eff}}$ can be computed as follows:
\begin{align*}
\EE\left[ \sqn{G(z_1) - G(z_2)} \right] 
&= (1-p)\mathbb{E}\left[ \frac{1}{(1-p)^2}\sqn{B_{\hat{\xi}_k}(z_1) - B_{\hat{\xi}_k}(z_2)} \right]
+ p\left[ \frac{1}{p^2}\sqn{\nabla \Psi( z_1) -  \nabla \Psi (z_2)} \right]\\
&= \frac{1}{1-p}\left[\sum_{i=1}^r p_i \cdot \sqn{B_i(z_1) - B_i(z_2)}\right] + \frac{1}{p} \sqn{\nabla \Psi(z_1) -  \nabla \Psi(z_2)}.
\end{align*}
Choose $p_i = \frac{1}{r}$:
\begin{align*}
  \EE\left[ \sqn{G(z_1) - G(z_2)} \right]  
&\leq \frac{1}{1-p}\sum_{i=1}^r \frac{L_B^2}{r} \sqn{z_1 -  z_2} + \frac{1}{p} L_{\Psi}^2  \sqn{z_1 -  z_2} 
\\&=\left( \frac{L_B^2}{1-p} + \frac{ L_{\Psi}^2}{p} \right) \sqn{z_1 -  z_2}  =L^2_{\text{eff}} \sqn{z_1 -  z_2}.
\end{align*}

Note that optimal complexities in \Cref{alg_sum_vi} for local computations and communications 
are achieved on \textbf{different sets of $p$ and $\rho$}. Let us get them separately.

$\bullet$ The local stochastic gradient complexity of a single iteration of \Cref{alg_sum_vi} is $0$ if $\xi^k = 0$ and $\xi^{k + \frac{1}{2}} = 1$, and $1$ if $\xi^k = 1$ and $\xi^{k+\frac{1}{2}} = 1$, and $r+1$ if $\xi^k = 1$ and $\xi^{k+\frac{1}{2}} = 0$, and $r$ if $\xi^k = 0$ and $\xi^{k+ \frac{1}{2}} = 0$. Thus, the
total expected local stochastic gradient complexity is bounded by
\begin{align*}
     \mathcal{ O} \left(\left(1 - p + r\rho \right)\left(\frac{1}{\rho} + \frac{ L_{\text{eff}} }{\sqrt{\rho}\mu}\right)\log\frac{1}{\varepsilon}\right), 
\end{align*}
For $\rho = \frac{1}{r}$, $p = \frac{L_{\Psi}}{\bar{L} + L_{\Psi}}$ the
total expected local stochastic gradient complexity of  \Cref{alg_sum_vi} becomes
\begin{align*}
    \mathcal{ O} \left(\left(1 - p + r\rho \right)\left(\frac{1}{\rho} + \frac{ L_{\text{eff}} }{\sqrt{\rho}\mu_B}\right)\log\frac{1}{\varepsilon}\right) \leq \mathcal{ O} \left(2\left(r + \frac{ \sqrt{r}L_{\text{eff}} }{\mu_B}\right)\log\frac{1}{\varepsilon}\right) = \mathcal{ O} \left(\left(r + \frac{ \sqrt{r}\bar{L} }{\mu_B}\right)\log\frac{1}{\varepsilon}\right), 
\end{align*}
where $\bar{L} = \sqrt{L_B + L_{\Psi}}$.

$\bullet$ The total communication complexity of \Cref{alg_sum_vi} is the sum of communication complexity coming from the full gradient computation (if statement that includes $\xi^{k + \frac{1}{2}}$) and the rest (if statement that includes $\xi^k$). The former requires a communication if $\xi^{k + \frac{1}{2}} = 0$,   the latter if $\xi^k$ is equal to $0$. The expected total communication $\mathcal{ O}\left(\rho + p\right)$ per iteration.  Thus, the
total communication complexity is bounded by
\begin{align*}
    \mathcal{ O} \left((p + \rho)\left(\frac{1}{\rho} + \frac{ L_{\text{eff}} }{\sqrt{\rho}\mu_B} \right)\log\frac{1}{\varepsilon}\right).
\end{align*}

For $\rho = p$, $p = \frac{L_{\Psi}^2}{\bar{L}^2 + L_{\Psi}^2}$ the total communication complexity of \Cref{alg_sum_vi} becomes
\begin{align*}
    \mathcal{ O} \left(\left(\rho +p\right)\left(\frac{1}{\rho}+\frac{ L_{\text{eff}}}{\sqrt{\rho}\mu_B}\right)\log{\frac{1}{\varepsilon}}\right)
    &= \mathcal{ O} \left(\left(1+\frac{ \sqrt{\rho}L_{\text{eff}}}{\mu_B}\right)\log{\frac{1}{\varepsilon}}\right) 
    \\&= \mathcal{ O} \left(\frac{L_{\Psi}}{\sqrt{L_B^2 + L_{\Psi}^2}}\frac{\sqrt{L_B^2 + L_{\Psi}^2} }{\mu_B}\log{\frac{1}{\varepsilon}}\right)
     \\&= \mathcal{ O} \left(\frac{L_{\Psi}}{\mu_B}\log{\frac{1}{\varepsilon}}\right). 
\end{align*}
To get proof of \textbf{\Cref{theorem_randomized}} we use that we apply \Cref{alg_sum_vi}  for solving the problem \eqref{PF} with $B_i(z) = [\nabla_X F_i(X, Y)^T, -\nabla_Y F_i(X,Y)^T]^T$, $\Psi(z) = \frac{\lambda}{2}\langle z, Wz \rangle = \frac{\lambda}{2}\langle X, WX \rangle + \frac{\lambda}{2}\langle Y, WY \rangle$. Also, for the problem \eqref{PF} $L_B = L$, $\mu_B = \mu$ and $L_{\Psi} = \lambda \lambda_{\max}(W)$.

\newpage
\section{Comparison with results on general distributed SPPs}\label{sec:comp}
In this section we compare the lower and upper bounds for decentralized personalized federated SPP with lower bounds and upper for \textbf{general} decentralized distributed SPP from literature. We will focus on the following works: \cite{liu2019decentralized_nc_nc}, \cite{Mukherjee2020:decentralizedminmax},\cite{beznosikov2020decentralized}, \cite{rogozin2021decentralized}. The comparison results are presented in Table \ref{table_sum}.  There are other papers in the literature, but for some reason they are not relevant to us, we will talk about them at the end of this section.

 In works \cite{liu2019decentralized_nc_nc}, \cite{Mukherjee2020:decentralizedminmax}, \cite{beznosikov2020decentralized} the authors consider the following Decentralized SPP:
\begin{equation}\label{DSPP}
    \min_{x \in \mathcal{X}}\max_{y \in \mathcal{Y}} \frac{1}{M}\sum_{m=1}^{M}f_m(x, y),
\end{equation}


There is no personalization in this problem, here we just solve a global distributed problem.

The upper bounds for this problem in strongly-convex-strongly-concave case presents in \cite{Mukherjee2020:decentralizedminmax}. These bounds are not optimal. For non-convex-non-concave case the upper bounds presents in \cite{liu2019decentralized_nc_nc}. The lower and near-optimal upper bounds for all cases of this problem presents in \cite{beznosikov2020decentralized}. In Table \ref{table_sum} we compare these results with lower and upper bounds for the problem \eqref{PF}. 
It can be seen that our estimates can be better, because of personalized setting. Those, we not only solve the problems of personalization in federated saddle point problems, but also benefit from it in communications.

In \cite{rogozin2021decentralized} the authors consider Decentralized SPP with local and global variables:
\begin{equation}\label{DSPP_local_var}
    \min_{x \in \mathcal{X},\\ p \in \mathcal{P}}\max_{y \in \mathcal{Y}, \\ r \in \mathcal{R}} \frac{1}{M}\sum_{m=1}^{M}f_m(x_m, p, y_m, r),
\end{equation}
where $x = (x_1^T, \dots, x_M^T)$, $y = (y_1^T, \dots, y_M^T)$ and $\mathcal{X} = \mathcal{X}_1 \times \dots \times \mathcal{X}_M$, $\mathcal{Y} = \mathcal{Y}_1 \times \dots \times \mathcal{Y}_M$. Variables $x_m, p, y_m, r$
have dimensions $d_x$, $d_p$, $d_y$, $d_r$, respectively. For each node, individual variables $x_m, y_m$ are restricted to sets $\mathcal{X}_m$, $\mathcal{Y}_m$ respectively. Sets $\mathcal{X}_m$, $\mathcal{Y}_m$, $m = 1, \dots , M$, $\mathcal{P}$, $\mathcal{R}$ are convex compacts. Each $f_m(\cdot, \cdot, y_m, r)$ is convex on $\mathcal{X}_m \times \mathcal{P}$ for every fixed $y_m \in \mathcal{Y}_m$, $r \in \mathcal{R}$. Each $f_m(x_m, p, \cdot, \cdot)$ is concave on $\mathcal{Y}_m \times \mathcal{R}$ for every fixed $x_m \in \mathcal{X}_m$, $p \in \mathcal{P}$.  Each node $m$ stores a local copy $p_m$, $r_m$ of the global variables $p$ and $r$, and consensus constraints $p_1 = \dots = p_M$, $r_1 = \dots = r_M$ are imposed.

 Unlike the problem \eqref{DSPP}, the problem \eqref{DSPP_local_var} considers local and global variables. It means that we can learn global and local models. Due to this \cite{rogozin2021decentralized} is closer to our formulation \eqref{PF}. In Table \ref{table_sum} one can find lower and upper bounds for \cite{rogozin2021decentralized} and note that they are very close to bounds for \eqref{DSPP} and worse that ours.

Also, decentralized min-max problem were considered in the works \cite{srivastava_distributed}, \cite{distributed_GANs} . However, in paper \cite{srivastava_distributed} the authors consider the problem $\min_{x \in X}\max_{i \in V}f_i(x)$ where $V$ is set of nodes in $W$, which is not convex (and also discrete). Moreover, this problem considered without smoothness assumption on $f_i$.  In paper \cite{distributed_GANs} the authors consider distributed stochastic saddle-point problem under the assumption that data is homogeneous which 
contradicts federated learning problem formulation. Therefore, we did not compare our results with the results from these papers. 

\textbf{Conclusion.} 
By personalized formulation we killing two birds with one stone: 1) can correctly mix global and local models for each user, 2) and also can significantly reduce the number of communications (especially when $\lambda \lambda_{\max}(W) \leq L \sqrt{\chi}$). The other works, which consider Distributed Saddle-Point Problem do not have these options. Despite the authors of \cite{rogozin2021decentralized} consider local and global variables, that seems like the personalization problem \eqref{PF}, they can not reduce the number of communications.

\renewcommand{\arraystretch}{2}
\begin{table*}[h]
    \centering
    \small
    \label{tab:comparison4}
\resizebox{0.94\textwidth}{!}{
\begin{threeparttable}
    \begin{tabular}{|c|c|c|c|c|}
    \cline{2-5}
    \multicolumn{1}{c|}{}
      &\textbf{\quad\quad\quad\quad Reference \quad\quad\quad\quad} &\textbf{Problem}& \textbf{Communication complexity} & \textbf{Local complexity} \\
    \hline
      \multirow{4}{*}{\rotatebox[origin=c]{90}{\textbf{Upper}}}
     & Mukherjee and Chakraborty \cite{Mukherjee2020:decentralizedminmax}  
     &\eqref{DSPP}
     &$\mathcal{O} \left( \chi^{\frac{4}{3}} \frac{L^{\frac{4}{3}}}{\mu^{\frac{4}{3}}} \log \frac{1}{\varepsilon} \right)$ &  $\mathcal{O} \left( \chi^{\frac{4}{3}} \frac{L^{\frac{4}{3}}}{\mu^{\frac{4}{3}}} \log \frac{1}{\varepsilon} \right)$
    \\ \cline{3-5}
    & Beznosikov et al. \cite{beznosikov2020decentralized}&\eqref{DSPP} & $\mathcal{O} \left( \sqrt{\chi} \frac{L}{\mu} \log^2 \frac{1}{\varepsilon} \right)$ &  $\mathcal{O} \left(  \frac{L}{\mu} \log \frac{L+\mu}{\mu} \log \frac{1}{\varepsilon}\right)$
    \\ \cline{3-5}
    & Rogozin et al. \cite{rogozin2021decentralized} &\eqref{DSPP_local_var}&$\mathcal{O} \left( \sqrt{\chi} \frac{L}{\mu} \log \frac{1}{\varepsilon} \right)$ &  $\mathcal{O} \left(  \sqrt{\chi} \frac{L}{\mu}\log \frac{1}{\varepsilon}\right)$ 
    \\ \cline{3-5}
    & \cellcolor{bgcolor2}{This paper}&\cellcolor{bgcolor2}{\eqref{PF}} &\cellcolor{bgcolor2} {$\mathcal{O} \left( \textcolor{black}{\min}\left[\sqrt{\chi}\frac{L}{\mu}; \textcolor{black}{\sqrt{\frac{\lambda \lambda_{\max}(W)}{\mu}}} \right]\log^2 \frac{1}{\varepsilon} \right)$}  & \cellcolor{bgcolor2} {$\mathcal{O} \left( \frac{L}{\mu} \log^2 \frac{1}{\varepsilon} \right)$} 
     \\ \cline{1-5} 
      \multirow{3}{*}{\rotatebox[origin=c]{90}{\textbf{Lower}}}  
    &  Beznosikov et al. \cite{beznosikov2020decentralized}  &\eqref{DSPP}& $\Omega \left( \sqrt{\chi} \frac{L}{\mu} \log \frac{1}{\varepsilon} \right)$ &  $\Omega \left(  \frac{L}{\mu} \log \frac{L+\mu}{\mu} \log \frac{1}{\varepsilon}\right)$
    \\ \cline{3-5}
     & Rogozin et al. \cite{rogozin2021decentralized}  &\eqref{DSPP_local_var} & $\Omega \left( \sqrt{\chi} \frac{L}{\mu} \log \frac{1}{\varepsilon} \right)$ &  $\Omega \left( \frac{L}{\mu}\log \frac{1}{\varepsilon}\right)$ 
    \\ \cline{3-5}
    & \cellcolor{bgcolor2}{This paper}&\cellcolor{bgcolor2}{\eqref{PF}} & \cellcolor{bgcolor2}{$\Omega \left( \textcolor{black}{\min}\left[\sqrt{\chi}\frac{L}{\mu}; \textcolor{black}{\sqrt{\frac{\lambda \lambda_{\max}(W)}{\mu}}} \right]\log \frac{1}{\varepsilon} \right)$}  &  \cellcolor{bgcolor2}{$\Omega \left( \frac{L}{\mu} \log \frac{1}{\varepsilon} \right)$} 
    \\\hline
    \end{tabular}   
    \end{threeparttable}
    }
    \caption{ Summary of lower bounds for \textbf{decentralized} personalized federated SPP and lower bounds for \textbf{decentralized} distributed SPP for finding an $\varepsilon$-solution in the deterministic gradient setup. 
    Convergence is measured by the distance to the solution. 
    {\em Notation:} $\mu$ = strong-convexity constant of $f$, $L$ = constant of $L$-smoothness of $f$,  
    $\lambda_{\max}(W)$ = maximum eigenvalue of $W$, $\lambda_{\min}(W)$ = minimum eigenvalue of $W$, $\chi = \lambda_{\max}(W) / \lambda_{\min}(W)$. }
    \label{table_sum}
\end{table*}

\newpage 
\section{Motivating examples} \label{sec:a_mot}
Most recently, significant attention of the community was devoted to personalized saddle problems in machine learning,e.g., \textbf{Personalized Search Generative Adversarial Networks (PSGANs)} \cite{PSGAN}.
The minimax game in PSGAN can be described as: given a query
posted by a user, the generator tries to produce a (negative) document that looks like fitting the user’s intent and to fool the discriminator; while the discriminator tries to draw a clear distinction between the relevant documents and the negative document samples generated by the generator. Formally, the problem are written as a personalized min-max problem :
\begin{align*}
    \min_{\theta} \max_{\psi} \sum_{q \in Q}\left(\mathbb{E}_{d \sim p_{true}(d|q,U,r)}\left[\ln D_{\psi}(d | q, U)\right] + \mathbb{E}_{d \sim p_{\theta}(d|q, U, r)}\left[\ln(1 - D_{\psi}(d|q,U)\right]\right),
\end{align*}
where $Q$ is a set of queries and each query $q \in Q$ is issued by a user $u$. $U$ is used to represent all historical search behaviours (search sessions) of $u$ before the current query $q$. The sessions in U splited into two parts according to their timing: the past
sessions $L_u$ and the current session $S_M$ , i.e., $U = L_u \cup \{S_M\}$, $L_u = \{S_1, \dots ,S_i
, \dots ,S_{M-1}\}$, where $M$ is the number of sessions associated with $u$. Each session $S_i$ is comprised
of a sequence of queries and each query includes a query string and a list of documents returned by the search engine, $q_j^i$ is the $j$-th query
in $i$-th session, and $D^i_j$
is the search results. $S_M$ includes the queries issued before $q$ in the same session. $d$ is
denoted a document in the results of query $q$ issued by user $u$, whose
historical search data is denoted by $U$. $p_{true}(d|q, U,r)$ is defined as the underlying true distribution
of relevance $r$ which is the personalized relevance preference of
user $u$ over document $d$ with respect to query $q$ and $u$’s historical
search data $U$. $\theta$ are parameters in generator, $\psi$ are parameters in discriminator, $p_{\theta}$ is a generated distribution,  $D_{\psi}(d|q, U) = \frac{\exp{f_{\psi}(d,q,U)}}{1 + \exp{f_{\psi}(d,q,U)}}$, $f_{\psi}$ is a function of discriminator. 

Another example of application personalise federated learning for saddle-point problem is \textbf{Robust models with adversarial noise}, considering that all objective functions in \eqref{PF} have the form \cite{madry2017towards}:
\begin{equation}
\label{eq:robust}
    f_m(x_m, y_m) := \tfrac{1}{N_m} \sum_{n=1}^{N_m} \ell(g(x_m, a_n + y_n), b_n)  + \tfrac{\beta_x}{2} \| x_m\|^2 - \tfrac{\beta_y}{2} \|y_m \|^2, \ \ \ \varphi(X) = \frac{\lambda}{2}\left\|\sqrt{W}X\right\|_F^2,
\end{equation}
where $x_m$ are the weights of the $m$th model, $\{(a_n, b_n)\}_{n=1}^{N_m}$ are pairs of the training data on the $m$th node, $y_m$ is the so-called adversarial noise, which introduces a small effect of perturbation in the data, coefficients $\beta_x$ and $\beta_y$ are the regularization parameters, $X = [x_1, \dots, x_{N_m}]$. 

Reformulation \eqref{eq:robust} is universal -- it is just adding robustness to any model training. Those, if we have some personalized minimization problem \cite{smith2017federated,hanzely2020federated,hanzely2020lower} and want to make the process more robust, we will get a personalized saddle point problem.


Personalized min-max problems are applied for \textbf{Lagrangian multipliers method}. Let us consider personalized minimization problem with constraints:
\begin{align*}
    \min_{g_i(x_j) \leq 0, i = \overline{1,n}}\sum_{m=1}^M f_m(x_m) + \frac{\lambda}{2}\|\sqrt{W}X\|_F^2,
\end{align*}
where $X = [x_1^T, \dots, x_M^T]$, $X \in \mathcal{X}$.
We can reformulate this problem with Lagrangian multipliers \cite{srivastava_distributed}, \cite{Nunec_distributed}:
\begin{align*}
    \min_{\mathcal{X}}\max_{\mu_i \geq 0}\sum_{m=1}^M f_m(x_m) + \frac{\lambda}{2}\|\sqrt{W}X\|_F^2 + \sum_{i=1}^n \mu_ig_i(x_j) = \min_{\mathcal{X}}\max_{\mu_i \geq 0}\sum_{m=1}^M F_m(x_m, \mu_m) + \frac{\lambda}{2}\|\sqrt{W}X\|_F^2 ,
\end{align*}
where $F_m(x_m, \mu_m) = f_m(x_m) + \sum_{j=m}\mu_ig_i(x_j)$ and solve this problem without constraints (or with simple constraints) instead of initial problem with constraints like $g_i(x_j) \leq 0$. This is personalized distributed min-max problem.

Lagrange multipliers are the most famous classical example of how a minimization problem can turn into a saddle point problem. But there are other popular examples of saddle point reformulation of the minimization problem in \textbf{non-smooth optimization via smooth reformulations} \cite{nesterov2005smooth,nemirovski2004prox}, in \textbf{supervised learning} (with non-separable loss \cite{Thorsten}, with non-separable regularizer\cite{bach2011optimization}),  in \textbf{unsupervised learning} \cite{NIPS2004_64036755}, in \textbf{reinforcement learning} \cite{Omidshafiei2017:rl,Jin2020:mdp}. It turns out that if we had a personalized minimization problem, and then for some reason (for example, to simplify the process of the solution) rewrote it in the form of a saddle point problem, then we already began to have a personalized saddle point problem.

 Finally, there are also examples personalized min-max problems in \textbf{Matrix Games Theory}. Let us consider standard example game: thief and policeman. This is a toy example to illustrate personalization. Let the city be a square of $n \times n$ small
squares. In each square there is a house and a police box. Let the values of $w_i$ houses also be known.

$\bullet$ Every night, the thief chooses which house to rob, and the policeman
chooses the booth in which he will be on duty;

$\bullet$ The probability of catching a thief if he or she robs a house in square $i$, and a policeman
is on duty in square $j$ is equal to:
$\exp(-\alpha  \text{dist}(i, j))$.
I.e. decreases with increasing distance between squares.

The thief wants to maximize his expected profit:
$w_i (1 - \exp(- \alpha \text{dist}(i, j)))$, policeman - minimize it. Then, this problem can be formulated as:
\begin{equation}\label{matrix_game}
\min_{\Delta^{d_x}}\max_{\Delta^{d_y}}x^T A y,
\end{equation}
where $A_{ij} = w_i (1 - \exp(- \alpha \text{dist}(i, j)))$. This game can be complicated: suppose we have country with $M$ cities, each city has thieve and policemen. Thieves and policemen can also move between cities. Then, the problem \eqref{matrix_game} can be formulated as \eqref{PF}. A key factor in personalization is how much thieves and policemen can move between cities. If all thieves/policemen sits only in theirs city, then the policeman/thief needs to worry only his local opponent and build a strategy for him. If thieves/policemen move freely between cities and the next time they can appear anywhere, then all the police/thieves need a common global strategy.

\end{document}